\documentclass[10pt]{wlscirep}
\usepackage[utf8]{inputenc}
\usepackage[T1]{fontenc}
\usepackage{bm}
\usepackage[justification=justified]{caption}
\usepackage{multirow}
\usepackage{makecell}
\usepackage[colorlinks=true, allcolors=blue]{hyperref}
\usepackage{marvosym}
\usepackage{xurl}

\DeclareMathAlphabet{\mathcal}{OMS}{cmsy}{m}{n}

\newcommand{\exttabref}[1]{Extended Data Table~\ref{#1}}

\title{Spatial Transcriptomics-Guided Alignment Enhances Molecular Profiling in Pathology Foundation Model}
\author[1]{Fengtao Zhou}
\author[1]{Yingxue Xu}
\author[2, 3, 4, 5]{Zhengyu Zhang}
\author[1]{Yihui Wang}
\author[1]{Zhengrui Guo}
\author[1]{Ling Liang}
\author[1]{Jiabo Ma}
\author[1]{Cheng Jin}
\author[1]{Ziyi Liu}
\author[1]{Huajun Zhou}
\author[1]{Hongyi Wang}
\author[6, 7, 8]{Du Cai}
\author[9]{Chenglong Zhao}
\author[1]{Xi Wang}
\author[10]{Can Yang}
\author[11]{Yu Wang}
\author[12]{Wenbin Li}
\author[6, 7, 8]{Feng Gao}
\author[13]{Zhe Wang}
\author[14]{Zhenhui Li}
\author[15]{Xiuming Zhang}
\author[2, 3, 4, 5]{Li Liang}
\author[1, 16, 17, 18, 19, \Letter]{Hao Chen}
\affil[1]{Department of Computer Science and Engineering, The Hong Kong University of Science and Technology, Hong Kong SAR, China}
\affil[2]{Department of Pathology, Nanfang Hospital, Southern Medical University, Guangzhou, China}
\affil[3]{Department of Pathology, School of Basic Medical Sciences, Southern Medical University, Guangzhou, China}
\affil[4]{Guangdong Province Key Laboratory of Molecular Tumor Pathology, Guangzhou, China}
\affil[5]{Jinfeng Laboratory, Chongqing, China}
\affil[6]{Department of General Surgery (Colorectal Surgery), The Sixth Affiliated Hospital, Sun Yat-sen University, Guangzhou, China}
\affil[7]{Guangdong Provincial Key Laboratory of Colorectal and Pelvic Floor Diseases, The Sixth Affiliated Hospital, Sun Yat-sen University, Guangzhou, China}
\affil[8]{Biomedical Innovation Center, The Sixth Affiliated Hospital, Sun Yat-sen University, Guangzhou, China}
\affil[9]{Department of Pathology, Shandong Provincial Qianfoshan Hospital, Jinan, China}
\affil[10]{Department of Mathematics, The Hong Kong University of Science and Technology, Hong Kong SAR, China}
\affil[11]{Department of Pathology, Zhujiang Hospital, Southern Medical University, Guangzhou, China}
\affil[12]{Department of Pathology, Cancer Hospital Chinese Academy of Medical Sciences, Beijing, China}
\affil[13]{State Key Laboratory of Holistic Integrative Management of Gastrointestinal Cancers, Department of Pathology, School of Basic Medicine and Xijing Hospital, Fourth Military Medical University, Xian, China}
\affil[14]{Department of Radiology, The Third Affiliated Hospital of Kunming Medical University, Yunnan Cancer Hospital, Kunming, China}
\affil[15]{Department of Pathology, The First Affiliated Hospital, School of Medicine, Zhejiang University, Hangzhou, China}
\affil[16]{Department of Chemical and Biological Engineering, Hong Kong University of Science and Technology, Hong Kong SAR, China}
\affil[17]{Division of Life Science, Hong Kong University of Science and Technology, Hong Kong SAR, China}
\affil[18]{HKUST Shenzhen-Hong Kong Collaborative Innovation Research Institute, Futian, Shenzhen, China}
\affil[19]{State Key Laboratory of Nervous System Disorders, The Hong Kong University of Science and Technology, Hong Kong SAR, China}

\affil[\Letter]{%
	Corresponding Authors \protect\par \vspace{-1.0em}
	\textbf{Lead Contact: Hao Chen (jhc@ust.hk)}
}

\begin{abstract}
	Comprehensive molecular profiling is essential for modern precision oncology but remains hindered by prohibitive costs, specimen exhaustion, and protracted turnaround times. While pathology foundation models (PFMs) have demonstrated potential for inferring molecular phenotypes from routine hematoxylin and eosin (H\&E) whole-slide images (WSIs), current architectures primarily rely on vision-centric self-supervised learning or vision-language alignment, lacking the spatially resolved molecular supervision required to connect subtle morphological features with underlying genomic alterations. Spatial transcriptomics (ST) emerges as a transformative technology that enables transcriptomic quantification within intact tissue sections, thereby preserving the precise spatial link between histology and molecular profiles. In this study, we present a Spatial Transcriptomics-guided Alignment framework for Molecular Profiling (\textbf{STAMP}), which endows PFMs with intrinsic molecular awareness. To support this paradigm, we curated HumanST-1k, a human ST dataset spanning diverse anatomical organs and sequencing platforms. This atlas yields 1.8 million pairs of H\&E patches and corresponding transcriptomic profiles, providing a corpus that links histological structures with their molecular states. To mitigate the technical noise inherent to raw transcriptomics, STAMP applies a pathway-informed alignment strategy that aggregates transcriptomic data into biologically functional pathways, which are subsequently integrated into PFMs via parameter-efficient fine-tuning. This alignment enriches the representation space of PFMs and unlocks their capacity to resolve sub-visual molecular signatures. The clinical utility of these augmented representations was validated through a multi-tier evaluation framework. At the microenvironmental level, STAMP facilitates the prediction of spatial gene expression and recognition of tumor function domains. Expanding to clinical scenarios, we evaluated STAMP across 18 clinical markers, spanning 5 tumor types, 13 centers, 67 cohorts, 18,636 patients, and 37,229 slides. STAMP consistently enhances the predictive performance of PFMs on diagnostic immunohistochemical biomarkers, actionable driver mutations, immunotherapy response indicators, and molecular prognostic signatures. Beyond retrospective internal and external evaluations, we confirm its real-world diagnostic value by demonstrating clinical utility in prospective observational cohorts. By transforming time-consuming and labor-intensive molecular assays into rapid H\&E-based computational phenotyping, STAMP conserves resources for comprehensive personalized profiling while prioritizing critical testing to accelerate therapeutic interventions.
\end{abstract}

\begin{document}

\flushbottom
\maketitle

\section*{Introduction}
Precision oncology has fundamentally transformed cancer management, shifting the paradigm from conventional histopathological evaluation toward molecular-driven therapeutic strategies~\cite{mateo2022delivering}. Comprehensive molecular profiling is now essential for accurate patient stratification and targeted treatment selection~\cite{akhoundova2022clinical, brlek2025advances}. However, standard clinical assays for acquiring these molecular phenotypes face severe logistical and economic bottlenecks. Techniques such as next-generation sequencing (NGS) and immunohistochemistry (IHC) are labor-intensive, cost-prohibitive, and inherently tissue-destructive, frequently leading to the exhaustion of precious diagnostic biopsy specimens~\cite{bera2019artificial, niazi2019digital}. Furthermore, prolonged turnaround times can delay critical therapeutic interventions, particularly for patients with narrow therapeutic windows. Consequently, there is a compelling imperative to develop computational methods capable of inferring complex molecular alterations directly from routine, cost-effective hematoxylin and eosin (H\&E)-stained whole-slide images (WSIs)~\cite{kather2019deep, kather2020pan, schmauch2020deep}. This computational ambition is grounded in the central biological premise that genomic and transcriptomic alterations drive phenotypic evolution, leaving discernible, albeit sub-visual, morphological signatures within the tissue architecture~\cite{saltz2018spatial, barkley2022cancer, jerby2018cancer}.

Recent advances in representation learning have enabled the development of large-scale pathology foundation models (PFMs) pretrained on extensive histopathology datasets~\cite{chen2024towards, xu2024whole, vorontsov2024foundation, wang2024pathology, ma2025generalizable}. The prevailing paradigm primarily relies on vision-centric self-supervised learning (SSL), particularly DINOv2-based architectures~\cite{oquab2023dinov2}, which generate robust feature embeddings by optimizing pretext tasks without manual annotations. While these models excel at morphology-centric tasks such as malignancy detection and histological subtyping, their capacity for molecular inference remains limited. Without explicit molecular supervision, vision-only models struggle to capture the subtle phenotypic shifts induced by underlying genomic alterations. Parallel efforts have explored vision-language pretraining (VLP) frameworks~\cite{radford2021learning, li2022blip, yu2022coca} to align image embeddings with clinical text extracted from pathology reports~\cite{lu2024visual, huang2023visual, ikezogwo2023quilt, xu2025multimodal}. However, VLP approaches are constrained by the low granularity of routine reports, which typically provide diagnostic summaries rather than spatially resolved, high-dimensional molecular annotations. Therefore, existing PFMs face a fundamental modality gap, underscoring the critical need for a novel pretraining paradigm guided by explicit, spatially anchored molecular signals to achieve comprehensive computational phenotyping.

Spatial transcriptomics (ST) has emerged as a transformative technology for profiling gene expression while preserving tissue architecture~\cite{marx2021method, rao2021exploring}. Unlike traditional bulk or dissociation-based sequencing, ST measures transcript levels directly on intact tissue sections, maintaining the spatial link between histology and molecular states. Despite its research value, routine clinical use of ST remains limited by high costs, complex library preparation, and strict tissue quality requirements. We therefore propose using ST data not as a standalone diagnostic test, but as a high-quality training supervision signal for model development. By learning the relationship between tissue morphology and molecular profiles from large-scale ST datasets, we can recalibrate the representation space of existing PFMs, endowing them the molecular awareness required for robust downstream clinical inference.

To achieve this, we introduce the Spatial Transcriptomics-guided Alignment framework for Molecular Profiling (\textbf{STAMP}). We first built HumanST-1k, a curated dataset of paired H\&E images and ST profiles collected from multiple public repositories, encompassing a wide range of anatomical sites, sequencing platforms, and spatial resolutions. Recognizing that raw spatial gene expression data is frequently sparse and highly susceptible to technical dropouts, STAMP avoids direct, unstable mapping from individual genes to image patches. Instead, it employs a biologically informed, pathway-guided alignment strategy that aggregates high-dimensional transcriptomic profiles into established biological pathways~\cite{subramanian2005gene}. This design reduces computational dimensionality, mitigates overfitting to sparse single-gene distributions, and leverages biological redundancy to buffer against platform-specific artifacts. Using parameter-efficient fine-tuning~\cite{hu2022lora, han2024parameter}, STAMP efficiently integrates these spatial molecular insights into underlying PFMs, enriching their representation space with latent morpho-molecular relationships.

We evaluated STAMP across multi-center clinical cohorts on a comprehensive biomarker panel across multiple cancer types. First, at the microenvironmental level, we demonstrate that STAMP facilitates the prediction of spatial gene expression across 8 cancer types. We further show that the ST-guided representations enable the recognition of spatially resolved functional domains. When deploying STAMP for clinical scenarios, we evaluated its performance across 18 clinically relevant markers, including diagnostic immunohistochemical biomarkers, actionable driver mutations, immunotherapy response indicators, and molecular prognostic signatures. Across 5 tumor types, 13 centers, 67 cohorts, 18,636 patients, and 37,229 slides, STAMP consistently enhances the predictive performance of underlying PFMs. Beyond retrospective internal and external validations, we confirmed its real-world translational value by deploying STAMP within prospective observational cohorts. By establishing conservative and reliable confidence thresholds, STAMP enables a triage workflow that reduces unnecessary confirmatory testing, thereby streamlining clinical workflows. Together, these results show that ST-guided alignment captures latent morphology-molecular relationships, offering a scalable and cost-effective pathway for computational phenotyping in precision oncology.

% We evaluated STAMP across multi-center clinical cohorts on a comprehensive biomarker panel across multiple cancer types. At the microenvironmental level, STAMP facilitates the prediction of spatial gene expression and recognition of spatially resolved functional domains. Expanding to clinical scenarios, STAMP achieves consistent performance improvements in the identification of diagnostic immunohistochemical biomarkers, the detection of actionable driver mutations, the profiling of immunotherapy response indicators, and the prediction of molecular prognostic signatures. Beyond retrospective internal and external validations, we confirmed its real-world translational value by deploying STAMP within prospective observational cohorts. By establishing reliable confidence thresholds, STAMP enables a triage workflow that safely reduces unnecessary confirmatory testing without compromising diagnostic accuracy, thereby streamlining clinical workflows. Together, these results show that ST-guided alignment effectively captures latent morphology-molecular relationships, offering a scalable and cost-effective pathway for computational phenotyping in precision oncology.

\begin{figure}[!htbp]
	\centering
	\includegraphics[width=\linewidth]{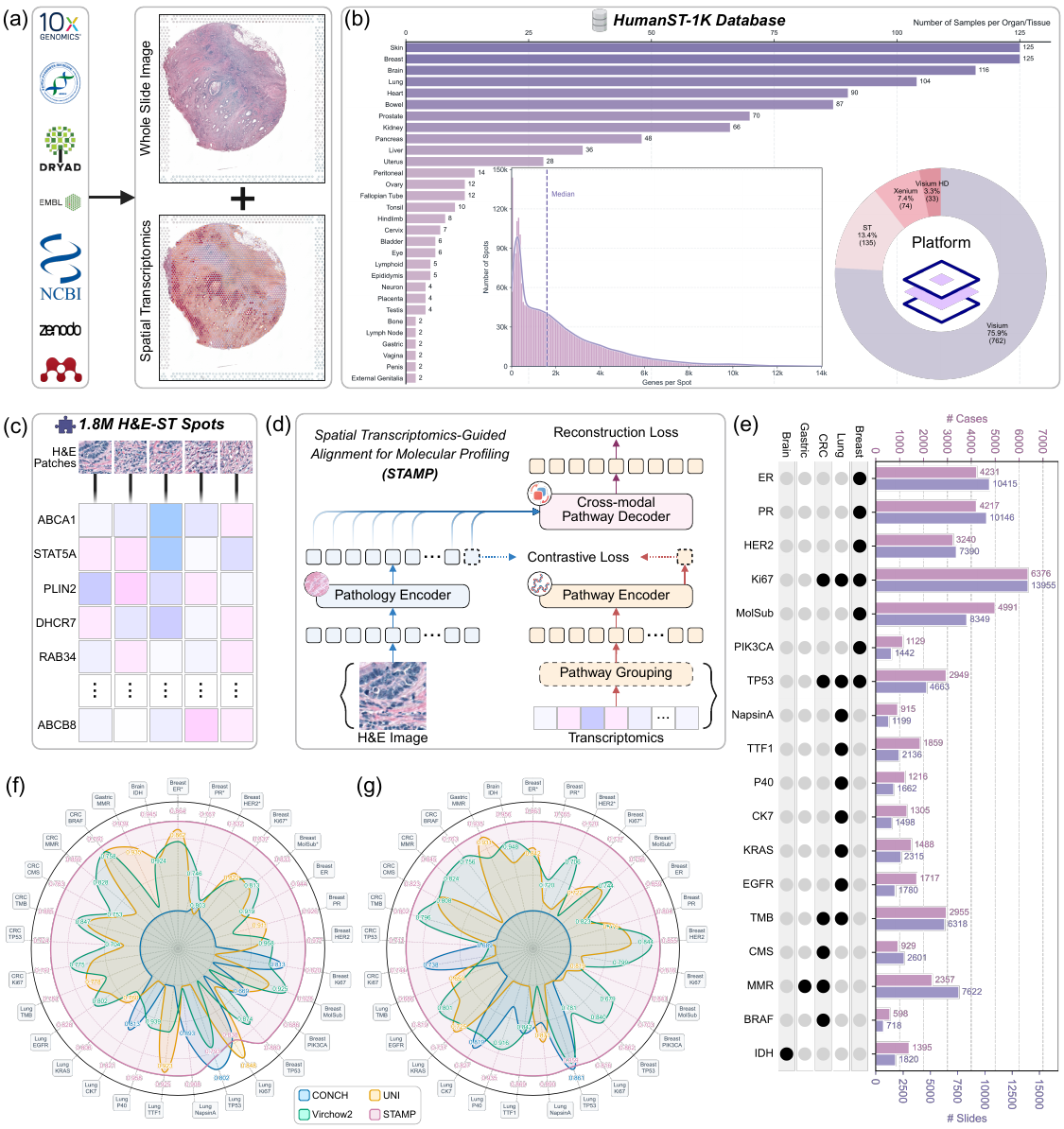}
	\caption{\textbf{Establishment and clinical validation of STAMP.} (\textbf{a}) Data curation process harmonizing paired H\&E WSIs and ST data from public repositories. (\textbf{b}) Demographics of the HumanST-1k dataset, detailing the pan-cancer distribution across organs, detected genes per spot, and spatial sequencing platforms. (\textbf{c}) Pre-training corpus comprising 2.1 million registered H\&E patches with paired spatial gene expression profiles. (\textbf{d}) STAMP architecture. Raw transcriptomic profiles undergo biologically informed pathway grouping to mitigate data sparsity. Using low-rank adaptation (LoRA), the pre-trained Virchow2 is aligned with this molecular modality via cross-modal contrastive and pathway reconstruction losses. (\textbf{e}) Summary of the downstream clinical evaluation cohorts. A dot matrix maps the evaluated biomarkers across five major cancer types, alongside bar charts quantifying the total cases and slides utilized per predictive task. (\textbf{f}, \textbf{g}) Performance benchmarking of STAMP against existing PFMs across the biomarker panel in internal testing (\textbf{f}) and independent external validation (\textbf{g}) cohorts.}
	\label{fig1}
\end{figure}

\section*{Results}
\subsection*{Development of STAMP framework and HumanST-1k atlas}
To construct a data foundation capable of supporting large-scale morpho-molecular alignment, we aggregated paired WSIs and ST profiles from multiple public repositories (Figure \ref{fig1}a). Given the high potential for data redundancy across decentralized databases, we implemented a strict duplicate-removal protocol using MD5 checksum verification, culminating in the establishment of the HumanST-1k dataset. There are 1,004 unique human ST samples in this atlas, spanning 30 organ/tissue types and 4 spatial sequencing platforms (Figure \ref{fig1}b). Each sample consists of a high-resolution H\&E WSI co-registered with a spatially resolved transcriptomic profile. By integrating transcriptomic profiles from diverse sequencing platforms, we ensured that the resulting model prioritizes the learning of biologically conserved features over platform-specific technical artifacts. To prevent data leakage, this atlas was partitioned into distinct pre-training (915 samples) and held-out evaluation sets (89 samples). A stringent spot-level quality control filter was applied to the pre-training split, excluding any spots with fewer than 100 detected pathway-related genes. This curation yielded a final training corpus of approximately 1.8 million paired H\&E patches and spatially resolved transcriptomic profiles (Figure \ref{fig1}c).

Building on this corpus, we developed the STAMP framework to align histological features with molecular states (Figure \ref{fig1}d). In this framework, we adopted Virchow2~\cite{vorontsov2024foundation, zimmermann2024virchow2}, one of the vision-centric and state-of-the-art PFMs~\cite{ma2025pathbench}, as the base visual encoder. To implement our biologically informed, pathway-guided alignment strategy, high-dimensional transcriptomic profiles are first grouped into biologically meaningful pathway representations~\cite{liberzon2015molecular}, and then a dedicated pathway encoder is used to extract functional representations. This molecular context is integrated into the visual domain via parameter-efficient fine-tuning, specifically low-rank adaptation (LoRA)~\cite{hu2022lora}, applied to the Virchow2 backbone. The model is optimized using a combination of a contrastive loss~\cite{radford2021learning} that aligns global morpho-molecular embeddings and a reconstruction loss that decodes pathway activity from visual features. This joint optimization effectively injects molecular awareness into the visual representations, enabling the model to capture latent morpho-molecular relationships that are critical for downstream clinical inference.

To systematically validate the clinical utility of STAMP, we architected a hierarchical, multi-tier evaluation framework escalating from fundamental molecular mapping to real-world clinical triage. We first evaluated the prediction of spatially resolved gene expression to verify that ST-guided pretraining embedded granular, pixel-to-transcript associative capabilities within the visual encoder. Building upon this localized foundation, we advanced to unsupervised tumor functional domain recognition, demonstrating the model's zero-shot capacity to aggregate low-level molecular cues into biologically coherent macro-architectures. Having established this intrinsic morpho-molecular correlation, we transitioned to clinical phenotyping across a benchmark of 24 predictive tasks covering 18 markers. This evaluation was designed to mirror the oncology care continuum, starting with initial diagnosis through immunohistochemical biomarkers, advancing to therapeutic stratification by identifying actionable driver mutations and immunotherapy response indicators, and culminating in prognostic risk assessment via the inference of molecular prognostic biomarkers. Performance comparisons were conducted against the Virchow2 baseline and other established PFMs~\cite{huang2023visual, chen2024towards, lu2024visual, chen2025visual} across both internal test sets and independent external validation cohorts. Finally, to bridge the gap between in silico benchmarking and true clinical deployment, a prospective observational cohort was established to simulate real-world diagnostic workflows and test STAMP's practical value as a clinical triage tool.

\subsection*{Inference of spatially resolved gene expression}
\begin{figure}[!htbp]
	\centering
	\includegraphics[width=\linewidth]{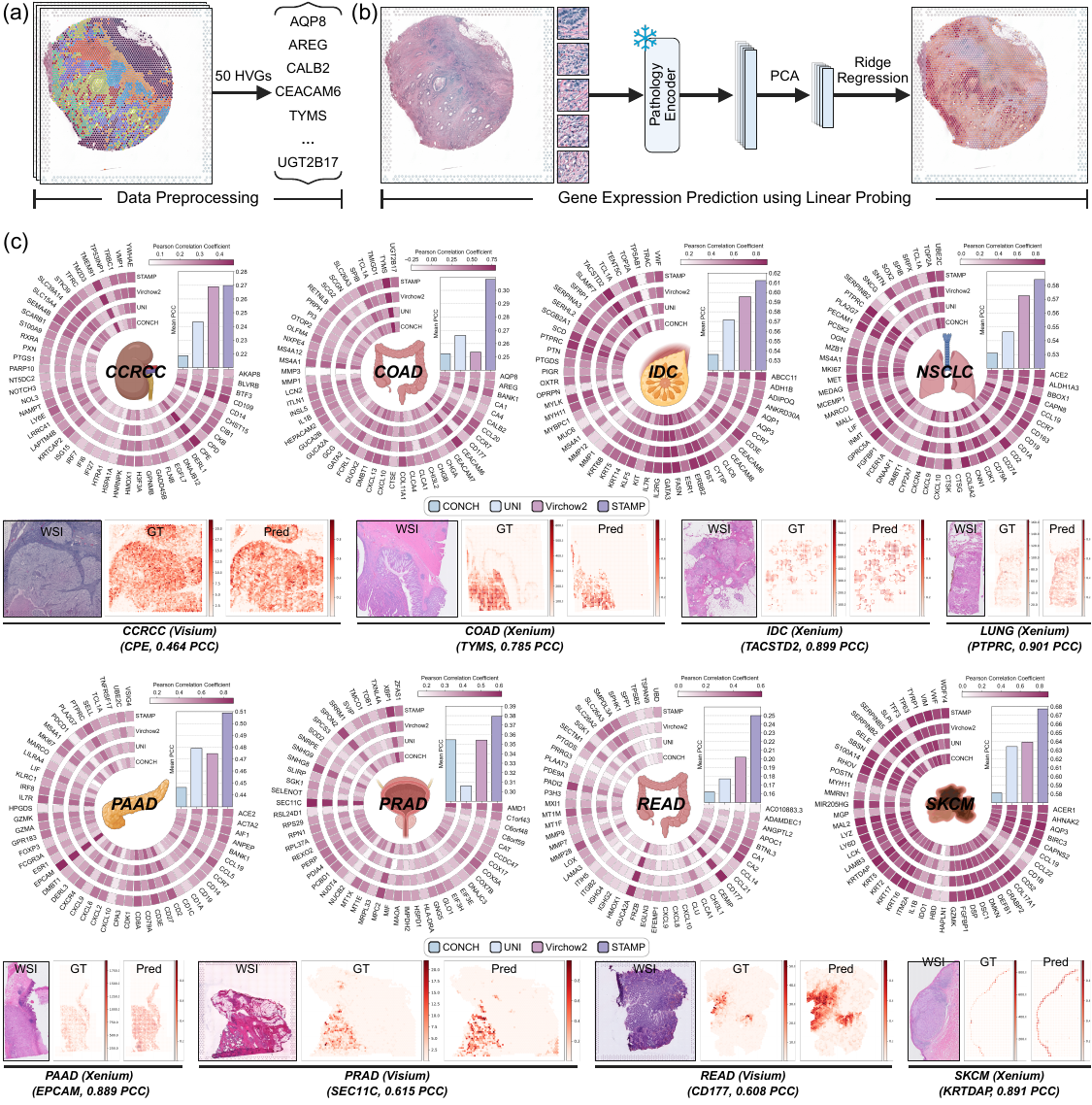}
	\caption{\textbf{Spatial gene expression inference via linear probing.} (\textbf{a}) Data preprocessing workflow illustrating the extraction of 50 highly variable genes (HVGs) from spatial transcriptomics datasets to serve as target molecular variables. (\textbf{b}) Linear probing pipeline for spatial gene expression prediction. The pathology encoder remains frozen. Extracted visual features undergo principal component analysis (PCA) for dimensionality reduction (256-dim), followed by ridge regression to infer spatially resolved expression profiles directly from H\&E patches. (\textbf{c}) Quantitative and qualitative evaluation of expression prediction across eight diverse oncology domains. Circular heatmaps benchmark the Pearson correlation coefficients (PCC) achieved by STAMP framework against existing PFMs across the 50 HVGs, with inset bar charts quantifying the overall mean PCC improvement relative to baselines. The bottom panels provide qualitative spatial maps demonstrating regional concordance among input H\&E WSI, ground truth (GT) transcriptomic profiles, and model predictions (Pred) for representative target genes across Visium and Xenium platforms.}
	\label{fig2}
\end{figure}

The spatial gene expression within the tumor microenvironment (TME) governs biological processes, including tumor progression, immune evasion, and therapeutic response~\cite{junttila2013influence, binnewies2018understanding, roma2019targeting, rao2021exploring}. Translating routine histology into quantitative transcriptomic maps requires vision models can accurately link localized morphological patterns to molecular readouts. To evaluate STAMP's capability in this domain, we assessed its performance in predicting spatially resolved gene expression directly from H\&E images. Following the established benchmark (\exttabref{hest-bench-dataset})~\cite{jaume2024hest}, we targeted the top 50 highly variable genes (HVGs) per cancer cohort (Figure \ref{fig2}a), which collectively capture the primary transcriptomic drivers of intratumoral heterogeneity and pathway activity. Feature representation quality was evaluated using a standardized linear probing pipeline (Figure \ref{fig2}b). The pathology encoder was frozen, and visual features extracted from H\&E patches were dimensionally reduced via principal component analysis (PCA) before being regressed against ground-truth transcriptomic profiles using ridge regression.

We applied this linear probing evaluation across eight diverse oncology cohorts, comprising clear cell renal cell carcinoma (CCRCC), colon adenocarcinoma (COAD), invasive ductal carcinoma (IDC), non-small cell lung cancer (NSCLC), pancreatic adenocarcinoma (PAAD), prostate adenocarcinoma (PRAD), rectal adenocarcinoma (READ), and skin cutaneous melanoma (SKCM). Predictive accuracy was quantified using the Pearson correlation coefficient (PCC) between model-predicted and experimentally measured gene expression values. Across all 400 evaluated HVGs, STAMP achieved a mean PCC of 0.449 (\exttabref{hest-bench}). This result consistently outperformed the Virchow2 baseline (0.420; $\Delta$+0.029) and established a clear performance margin over other PFMs, including UNI (0.403), CONCH (0.385), PLIP (0.320), \textit{etc}. Notably, STAMP also surpassed OmiCLIP~\cite{chen2025visual}, a model directly pre-trained on H\&E-ST dataset, by a substantial absolute margin of +0.185 (STAMP: 0.449 vs. OmiCLIP: 0.264). These performance gains proved highly consistent across individual cancer types (Figure \ref{fig2}c, Extended Data Table \ref{hest-bench-ccrcc}-\ref{hest-bench-skcm}). The most significant improvements relative to the Virchow2 baseline were observed in gastrointestinal malignancies, particularly COAD (+0.054) and READ (+0.048). Steady and consistent gains were further recorded in SKCM (+0.038), PAAD (+0.034), PRAD (+0.025), IDC (+0.016), and NSCLC (+0.012). These improvements across multiple cancer types confirm the effectiveness of STAMP in embedding molecular awareness into the visual encoder.

We further evaluated the spatial concordance of STAMP's predictions by visualizing predicted expression maps for representative genes across the eight cancer types (Figure \ref{fig2}c, lower panels). When applied to Xenium data, STAMP reconstructed the expected spatial localization of key lineage and functional markers. The model resolved the tumor-epithelial localization of \textit{TACSTD2} (0.899 PCC) in IDC, the leukocyte-enriched distribution of \textit{PTPRC} (0.901 PCC) in NSCLC, the epithelial compartmentalization of \textit{EPCAM} (0.889 PCC) in PAAD, and the epidermal differentiation signal of \textit{KRTDAP} (0.891 PCC) in SKCM. This spatial concordance was consistently maintained on spot-based Visium platforms, where STAMP predictions tracked the measured expression gradients of \textit{CPE} (0.464 PCC) in CCRCC, \textit{TYMS} (0.785 PCC) in COAD, \textit{SEC11C} (0.615 PCC) in PRAD, and \textit{CD177} (0.608 PCC) in READ. By recovering transcriptomic profiles for highly variable markers across multiple cancer types and sequencing technologies, these results demonstrate that ST-guided alignment facilitates visual encoder in capturing subtle morpho-molecular relationships that are critical for spatial gene expression inference. This establishes a promising computational foundation for inferring targeted molecular states directly from routine H\&E histology.

\subsection*{Recognition of spatially resolved functional domains}
Clinical interpretation of tumor sections relies on recognizing functional domains within the TME~\cite{keren2018structured}. To determine whether STAMP can resolve spatially organized tissue domains without physical sequencing, we evaluated its capacity in unsupervised spatial clustering. Four cancer types with paired H\&E-ST data and pathologist-annotated ground truth were selected for this benchmark (\exttabref{spatial-domain-dataset}), including breast cancer~\cite{andersson2020spatial}, lung cancer~\cite{dawo_10x_2025}, kidney cancer~\cite{dawo_10x_2025}, and prostate cancer~\cite{erickson2022spatially}. For each spot, we extracted a patch-level embedding from the co-registered H\&E image and performed unsupervised clustering after PCA-based dimensionality reduction. The spatial concordance between morphology-inferred clusters and expert annotations was quantified using the adjusted Rand index (ARI) for overall partition similarity, normalized mutual information (NMI) for distribution overlap, homogeneity (HOM) for cluster purity, and completeness (COM) for structural integrity.

\begin{figure}[!htbp]
	\centering
	\includegraphics[width=\linewidth]{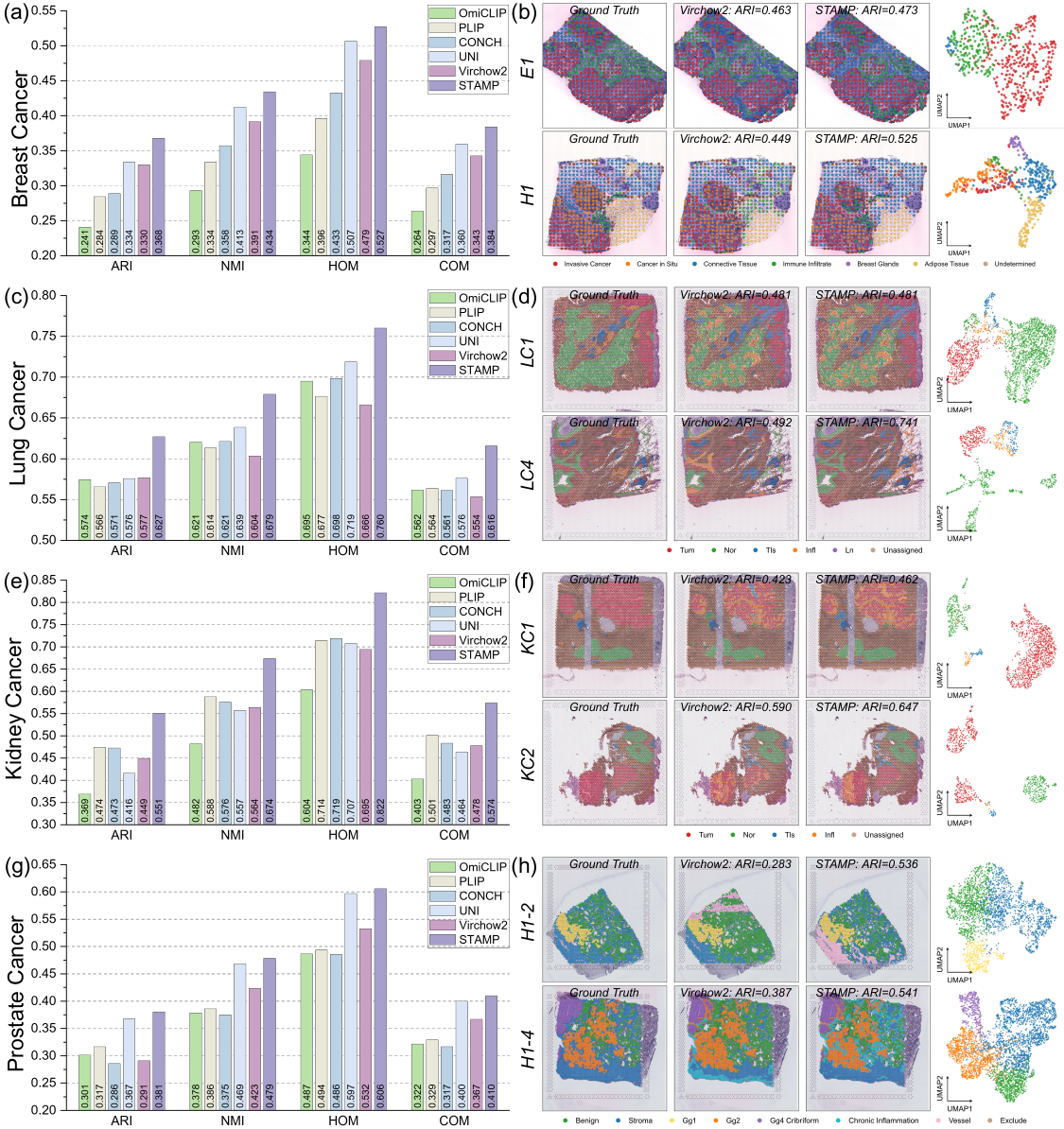}
	\caption{\textbf{Spatial domain recognition via unsupervised clustering.} Adjusted Rand Index (ARI), Normalized Mutual Information (NMI), Homogeneity (HOM), and Completeness (COM) scores quantifying the spatial concordance between unsupervised clusters derived from model embeddings and pathologist-annotated ground truth across four cancer types. (\textbf{a} and \textbf{b}) Overall performance benchmarking and representative spatial cluster maps for breast cancer. (\textbf{c} and \textbf{d}) Overall performance benchmarking and representative spatial cluster maps for lung cancer. (\textbf{e} and \textbf{f}) Overall performance benchmarking and representative spatial cluster maps for kidney cancer. (\textbf{g} and \textbf{h}) Overall performance benchmarking and representative spatial cluster maps for prostate cancer.}
	\label{fig3}
\end{figure}

Across the comprehensive multi-organ evaluation, unsupervised spatial partitions derived from STAMP yielded the highest alignment with pathologist's annotations, consistently establishing a robust discriminative lead over the Virchow2 baseline and other established multimodal PFMs. On breast cancer, pathologist's annotations encompass invasive cancer, cancer in situ, breast glands, connective tissue, immune infiltrate, and adipose tissue. STAMP achieved a mean ARI of 0.368, the highest among all tested models (Figure \ref{fig3}a, \exttabref{spatial-breast}), outperforming UNI (0.334), Virchow2 (0.330), CONCH (0.289), PLIP (0.285), and OmicCLIP (0.241) with corresponding advantages in NMI (0.434 vs. 0.391 for Virchow2) and HOM (0.527 vs. 0.479), indicating that STAMP enhances the model's ability to capture the spatial continuity of TME. Spatial cluster maps on representative sections illustrate this advantage visually. STAMP cluster boundaries trace the pathologist-delineated interface between invasive tumor, reactive connective tissue, and immune infiltrate(Figure \ref{fig3}b).

On lung cancer, sections span four to five annotation classes, \textit{i.e.}, tumor (Tum), normal (Nor), tertiary lymphoid structures (Tls), immune infiltration (Infl), and lymph node (Ln) tissue. STAMP attained a mean ARI of 0.627, compared with 0.577 for Virchow2 and 0.576 for UNI along with the highest mean NMI (0.679 vs. 0.604 for Virchow2) and HOM (0.760 vs. 0.700) (Figure \ref{fig3}c, \exttabref{spatial-lung}). The spatial visualizations in sections LC1 and LC4 reveal that STAMP more reliably distinguishes the difference between tumor, normal tissue, and immune infiltrate, with more coherent cluster boundaries that respect the pathologist-annotated interfaces. In contrast, Virchow2 clusters tend to merge immune infiltration and normal regions and fragment tumor across multiple spurious sub-clusters (Figure \ref{fig3}d). On kidney cancer, STAMP achieved a mean ARI of 0.551, surpassing PLIP (0.474), CONCH (0.473), Virchow2 (0.449), UNI (0.416), and OmicCLIP (0.369) by absolute margins of 7.7\%-18.2\% (Figure \ref{fig3}e, \exttabref{spatial-kidney}). STAMP also achieved the highest mean HOM (0.822 vs. 0.694 for Virchow2), indicating substantially purer cluster compositions with each predicted cluster corresponding to a single annotated domain class, as spatial visualizations in sections KC1 and KC2 confirm (Figure \ref{fig3}f).

The prostate dataset is the most annotation-granular benchmark in this evaluation, with up to eight domain classes per section reflecting the histological complexity. Gleason-grade tumor subclasses (Gg1, Gg2, Gg4 Cribriform) are defined by subtle differences in glandular architecture and nuclear morphology within spatially contiguous but morphologically overlapping territories, interspersed with benign glands, fibromuscular stroma, chronic inflammation, and vascular structures. This makes the task demanding for all models, and mean ARI values were relatively lower. Nevertheless, STAMP ranked first with a mean ARI of 0.381, outperforming UNI (0.367), PLIP (0.317), OmicCLIP (0.301), Virchow2 (0.291), and CONCH (0.286) (Figure \ref{fig3}g, \exttabref{spatial-prostate}). The performance gap ($\Delta$+9.0\% ARI) relative to Virchow2 is particularly informative. Due to the spatial intermixing of glandular subtypes, patch-level morphology alone provides limited discriminative signal for Gleason-grade distinctions. The substantial ARI improvement indicates that STAMP enables visual encoder to capture subtle molecular cues that are not visually discernible. STAMP's NMI (0.479) and HOM (0.606) substantially exceeded those of Virchow2 (0.423 and 0.532), and COM (0.410 vs. 0.367) was similarly elevated, confirming a structurally superior organization of the embedding space. Spatial visualizations, which contain the most clearly compartmentalized Gleason-grade regions in this cohort, show that STAMP clusters respect the spatial boundaries between Gg1, Gg2, and Gg4 Cribriform tumor subclasses and maintain coherent separation from benign glands, stroma, chronic inflammation, and vessel compartments, with markedly reduced fragmentation of Gleason-grade regions into spurious sub-clusters compared with Virchow2 (Figure \ref{fig3}h).

These results collectively demonstrate that, across four cancer types, STAMP consistently outperforms existing PFM models in unsupervised spatially resolved domain recognition. In particular, compared to the Virchow2 baseline model, STAMP's alignment strategy enables it to capture subtle cues that are difficult to discern morphologically, allowing the model to reconstruct the spatial structure of TME and resolve complex histological patterns crucial for clinical interpretation.

\subsection*{Identification of diagnostic immunohistochemical biomarkers}
Accurate determination of protein-level biomarkers via IHC remains the clinical cornerstone for tumor subtyping and therapeutic stratification. However, routine diagnostic workups frequently require panels of multiple IHC assays to establish definitive molecular classifications. The execution of these numerous stains imposes substantial laboratory bottlenecks, extending turnaround times and increasing operational costs (1~2 days per IHC stain) that delay treatment initiation. Besides, the increasing adoption of early diagnosis and neoadjuvant treatment has led to a growing reliance on core needle biopsies for initial diagnostic decisions. These limited specimens are highly susceptible to tissue exhaustion during sequential IHC sectioning (at least 5 serial tissue sections at 4-5 $\mu$m thickness in breast cancer), which can ultimately preclude patients from undergoing subsequent comprehensive genomic profiling (NGS). To address this logistical constraint, computational inference of protein biomarkers directly from routine H\&E WSIs has emerged as a high-value translational objective~\cite{couture2018image, valieris2024weakly}. We therefore assessed STAMP's predictive capacity for diagnostic IHC biomarkers across multi-institutional breast and lung cancer cohorts.

\begin{figure}[!htbp]
	\centering
	\includegraphics[width=\linewidth]{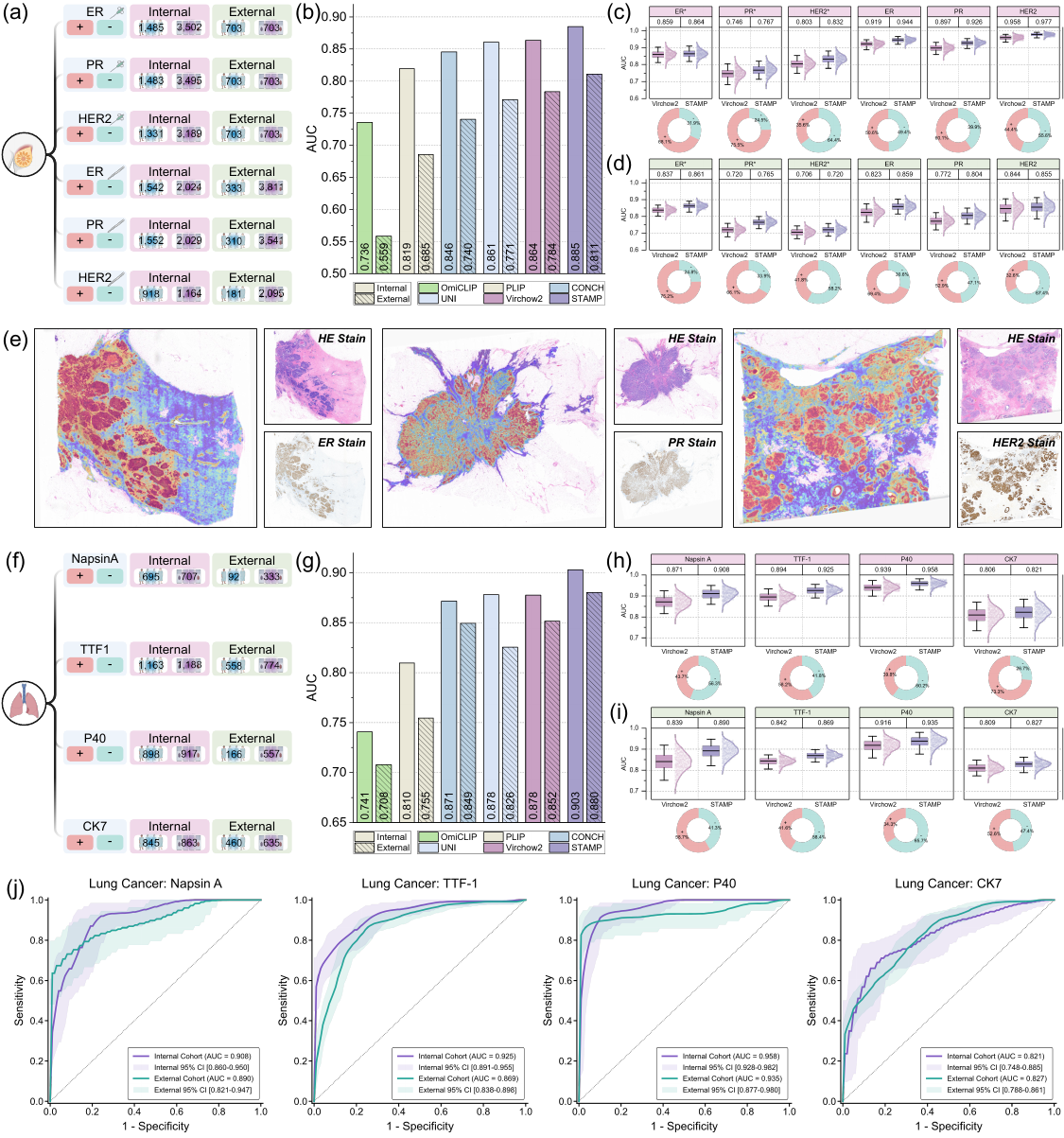}
	\caption{\textbf{Inference of diagnostic immunohistochemical biomarkers in breast and lung cancer.} (\textbf{a}) Summary of the clinical biomarkers and cohort design for breast cancer across biopsy and resection specimens. (\textbf{b}) Overall predictive performance for breast cancer biomarkers across internal testing and external validation cohorts. c, d, Performance distributions comparing STAMP against the Virchow2 baseline on the internal (\textbf{c}) and external (\textbf{d}) breast cancer cohorts, where `*' denote needle biopsy specimens. (\textbf{e}) Spatially resolved predictive heatmaps for ER, PR, and HER2. The high-attention regions inferred directly from routine H\&E WSIs exhibit strong spatial concordance with ground-truth IHC positive areas. (\textbf{f}) Summary of the clinical biomarkers and cohort design for lung cancer. (\textbf{g}) Overall predictive performance for lung cancer biomarkers across internal and external cohorts. (\textbf{h, i}) Performance distributions comparing STAMP and Virchow2 for the lung cancer diagnostic panel across internal (\textbf{h}) and external (\textbf{i}) cohorts. (\textbf{j}) Receiver operating characteristic (ROC) curves for individual lung cancer biomarkers in both validation settings, demonstrating the robust generalization of STAMP-derived representations.}
	\label{fig4}
\end{figure}

In breast cancer, we targeted the identification of ER, PR, and HER2 statuses~\cite{allison2020estrogen, wolff2018human}, in which these biomarkers are routinely examined in clinical practice to classify breast tumors into molecular subtypes that have distinct prognostic implications and therapeutic options. We evaluated STAMP's performance in predicting these biomarkers across multiple cohorts encompassing diverse surgical resection and core needle biopsy specimens (Figure \ref{fig4}a). We deployed internal test sets and independent external validation cohorts to quantify generalizability across institutional domain shifts and staining variations. Benchmarking across the combined internal cohorts demonstrated that pathway-guided cross-modal alignment consistently enhances biomarker prediction accuracy. STAMP attained a mean AUC (area under ROC curve) of 0.885, yielding a steady improvement over the Virchow2 baseline (0.864) and establishing a clear performance margin over compared PFMs, including UNI (0.861), CONCH (0.846), PLIP (0.819), and OmiCLIP (0.736) (Figure \ref{fig4}b). This discriminative advantage translated to external validation, where STAMP maintained a mean AUC of 0.811 compared to 0.784 for the baseline model. At the specimen level, STAMP demonstrated consistent predictive gains across all three biomarkers in core needle biopsy cohorts, where restricted tissue area poses inherent challenges for morphological inference (Figure \ref{fig4}c, d). In internal validation, STAMP achieved AUCs of 0.864 (95\% CI: 0.819-0.908) for ER, 0.767 (95\% CI: 0.714-0.819) for PR, and 0.832 (95\% CI: 0.778-0.880) for HER2, corresponding to absolute improvements of $0.5\%$ ($P<0.001$), $2.1\%$ ($P<0.001)$, and $2.9\%$ ($P<0.001$) over the Virchow2 baseline, respectively (\exttabref{breast-diagnostic-internal-biopsy}). These gains were preserved during external validation, where STAMP attained AUCs of 0.861 (95\% CI: 0.826-0.890) for ER, 0.765 (95\% CI: 0.727-0.800) for PR, and 0.720 (95\% CI: 0.682-0.758) for HER2, securing absolute improvements of $2.4\%$ ($P<0.001$), $4.5\%$ ($P<0.001$), and $1.4\%$ ($P<0.001$) over baseline model (\exttabref{breast-diagnostic-external-biopsy}).

In contrast, surgical resection specimens can offer expansive spatial context and preserved tissue architecture, enabling internal performance to peak at AUCs of 0.944 (95\% CI: 0.919-0.965) for ER, 0.926 (95\% CI: 0.895-0.953) for PR, and 0.977 (95\% CI: 0.957-0.990) for HER2, representing absolute gains of $2.5\%$ ($P<0.001$), $2.9\%$ ($P<0.001$), and +$1.9\%$ ($P<0.001$) over Virchow2, respectively (\exttabref{breast-diagnostic-internal-resection}). External validation confirmed the stability of these high performance levels, with STAMP reaching AUCs of 0.859 (95\% CI: 0.815-0.903) for ER, 0.804 (95\% CI: 0.755-0.849) for PR, and 0.855 (95\% CI: 0.786-0.912) for HER2, corresponding to absolute improvements of $3.6\%$ ($P<0.001$), $3.2\%$ ($P<0.001$), and $1.1\%$ ($P<0.001$) over Virchow2 (\exttabref{breast-diagnostic-external-resection}). The elevated AUCs in resection specimens compared to biopsies reflect the advantage of comprehensive tissue sampling and intact histological architecture for molecular inference. To verify that these predictions originate from biologically grounded morphological features rather than dataset-specific artifacts or slide-level confounders, we performed spatial concordance analysis using serial H\&E and IHC sections derived from identical tissue blocks (Figure \ref{fig4}e). Predictive heatmaps generated exclusively from H\&E inputs exhibited spatial alignment with experimentally determined IHC staining patterns across ER, PR, and HER2. Regions of high attention consistently co-localized with IHC-positive tumor compartments, while low-probability areas corresponded to stromal regions or IHC-negative tissue zones. This high spatial correspondence demonstrates that STAMP learns localized histological patterns associated with protein expression rather than non-specific slide-level biases.

We applied a parallel evaluation framework to lung cancer, focusing on four markers essential for the histological subtyping of non-small cell lung cancer (NSCLC)~\cite{travis20152015, yatabe2019best}, specifically Napsin A, TTF-1, p40, and CK7 (Figure \ref{fig4}f). Cross-modal supervision yielded consistent performance elevations across the panel, with STAMP achieving an aggregate internal AUC of 0.903 against 0.878 for Virchow2, while also surpassing UNI (0.878), CONCH (0.871), PLIP (0.810), and OmiCLIP (0.741) (Figure \ref{fig4}g). External cohorts showed parallel improvements, where STAMP reached 0.880 compared to the baseline model's 0.852. Marker-specific analysis revealed distinct performance patterns across the four targets (Figure \ref{fig4}h, i). The squamous lineage marker p40 achieved the highest discriminative accuracy, with STAMP attaining AUCs of 0.958 (95\% CI: 0.928-0.982) internally and 0.935 (95\% CI: 0.877-0.980) externally, reflecting the robust morphological signatures associated with squamous differentiation~\cite{bishop2012p40}. For glandular markers, Napsin A prediction improved from 0.871 (95\% CI: 0.815-0.924) to 0.908 (95\% CI: 0.860-0.950) internally and from 0.839 (95\% CI: 0.751-0.918) to 0.890 (95\% CI: 0.821-0.947) externally, representing the largest absolute gains among all lung cancer targets (+3.7\% and +5.1\%, respectively; $P<0.001$ for both; \exttabref{lung-diagnostic-internal}, \exttabref{lung-diagnostic-external}). TTF-1 also showed significant improvements, with STAMP reaching 0.925 (95\% CI: 0.891-0.955) internally and 0.869 (95\% CI: 0.838-0.898) externally, corresponding to gains of $3.1\%$ and $2.7\%$ over baseline ($P<0.001$). CK7, despite its inherently heterogeneous expression patterns, maintained stable performance with AUCs of 0.821 (internal) and 0.827 (external), securing improvements of $1.5\%$ and $1.8\%$ ($P<0.001$). The ROC curves underscored these findings, with p40 demonstrating near-complete class separation (Figure \ref{fig4}j). Notably, the substantial improvements for Napsin A highlights the particular value of pathway-guided alignment for generalizing glandular differentiation markers across institutional variations in tissue processing and staining protocols.

The IHC serves as the clinical standard for evaluating protein-level biomarker status. These protein expressions are fundamentally governed by underlying transcriptomic profiles. By learning the morpho-molecular relationships that govern protein expression, STAMP successfully captures the histological signatures associated with diagnostic biomarkers, enabling accurate inference from routine H\&E WSIs. The consistent performance gains across both breast and lung cancer cohorts, as well as the spatial concordance with ground-truth IHC patterns, confirm that ST-guided cross-modal supervision effectively enhances the model's capacity to recognize biologically meaningful morphological cues linked to protein expression.

\begin{figure}[!tbp]
	\centering
	\includegraphics[width=\linewidth]{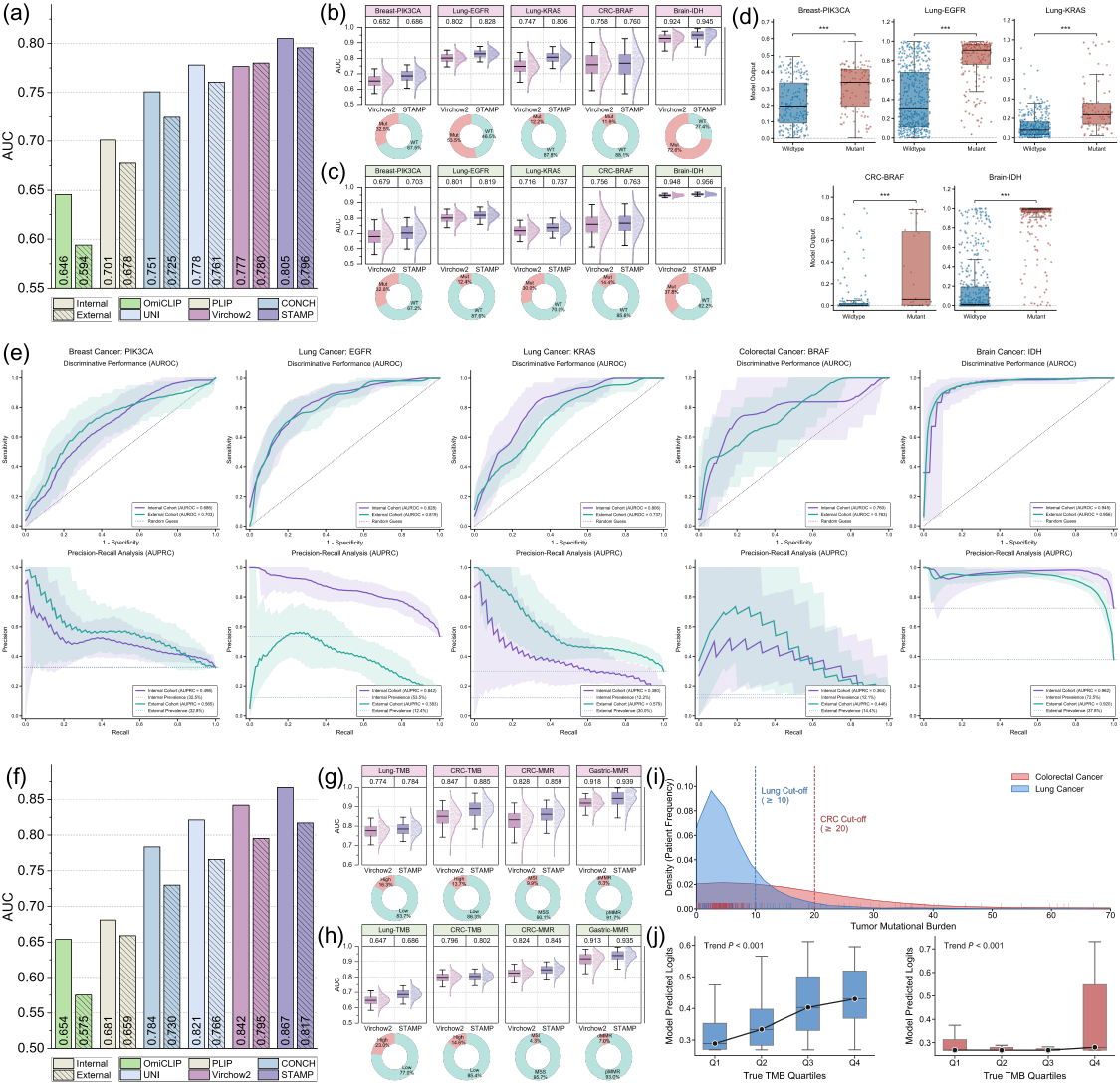}
	\caption{\textbf{Identification of actionable driver mutations and immunotherapy response indicators.} (\textbf{a}) Overall predictive performance for actionable driver mutations (PIK3CA in breast cancer, EGFR and KRAS in lung cancer, BRAF in colorectal cancer, and IDH in brain cancer) across internal and external cohorts. \textbf{b, c} Performance distributions of STAMP versus the Virchow2 baseline for individual driver mutations on the internal (\textbf{b}) and external (\textbf{c}) cohorts. (\textbf{d}) Model output distributions for wild-type versus mutant phenotypes. (\textbf{e}) Receiver operating characteristic (ROC) and precision-recall (PR) curves for targeted mutations in both validation settings. (\textbf{f}) Overall predictive performance for immunotherapy response markers, including tumor mutational burden (TMB) and mismatch repair (MMR) status, across lung, CRC, and gastric cancer cohorts. \textbf{g, h} Performance distributions of STAMP versus Virchow2 for individual immunotherapy biomarkers on the internal (\textbf{g}) and external (\textbf{h}) cohorts. (\textbf{i}) Density distributions of TMB values and clinical thresholds for TMB-High versus TMB-Low stratification in the lung and CRC cohorts. (\textbf{j}) STAMP predictive score distributions stratified by TMB quartiles in the lung (left) and CRC (right) cohorts.}
	\label{fig5}
\end{figure}

\subsection*{Detection of actionable driver mutations}
Genotype-directed oncology relies on accurate detection of actionable driver mutations to guide targeted therapy selection. While NGS serves as the clinical reference standard for mutational profiling, its routine deployment faces practical constraints including high per-test costs, multi-day turnaround times, and strict requirements for sufficient tumor cellularity and nucleic acid quality~\cite{mosele2020recommendations}. Inferring mutational status directly from routine H\&E WSIs offers a computationally efficient pre-screening alternative that could accelerate molecular triage~\cite{coudray2018classification, kather2020pan}. This task remains technically challenging because many genomic alterations induce only subtle, focal, or highly heterogeneous histological patterns that evade conventional morphological assessment. We therefore evaluated STAMP's capacity to predict five clinically critical mutations across four tumor types: \textit{PIK3CA} in breast cancer, \textit{EGFR} and \textit{KRAS} in lung cancer, \textit{BRAF} in colorectal cancer, and \textit{IDH} in brain cancer.

Integrating ST priors yielded systematic performance gains for mutation profiling across the multi-cancer panel (Figure \ref{fig5}a). STAMP achieved a mean internal AUC of 0.805, representing a steady improvement over the Virchow2 baseline (0.777). Benchmarking against other PFMs demonstrated consistent superiority, with STAMP outperforming UNI (0.778), CONCH (0.751), PLIP (0.701), and OmiCLIP (0.646). External validation preserved this advantage, with STAMP attaining an AUC of 0.796 compared to 0.780 for Virchow2. Task-level analysis revealed that the magnitude of improvement correlated with the known morphological penetrance of each mutation (Figure \ref{fig5}b-c, Extended Data Tables \ref{targeted-internal}-\ref{targeted-external}). Mutations with strong histopathological correlates demonstrated the highest accuracy and the most robust external generalization. \textit{IDH} prediction in brain tumors reached AUCs of 0.945 (95\% CI: 0.873-0.992) internally and 0.956 (95\% CI: 0.942-0.969) externally, representing absolute improvements of $2.1\%$ and $0.8\%$ over Virchow2 ($P<0.001$ for both). For lung cancer, STAMP delivered substantial and consistent gains for both \textit{EGFR} and \textit{KRAS}. \textit{EGFR} prediction improved from 0.802 (95\% CI: 0.744-0.853) to 0.828 (95\% CI: 0.777-0.876) internally and from 0.801 (95\% CI: 0.737-0.859) to 0.819 (95\% CI: 0.759-0.872) externally, corresponding to absolute gains of +2.6\% and +1.8\% ($P<0.001$). \textit{KRAS} showed the largest internal improvement among all targets, rising from 0.747 (95\% CI: 0.645-0.838) to 0.806 (95\% CI: 0.733-0.874), a +5.9\% absolute increase ($P<0.001$). External validation maintained a +2.1\% advantage (0.716 to 0.737; $P<0.001$), confirming that the model captures morphological signatures associated with this clinically prevalent alteration. \textit{BRAF} prediction in colorectal cancer exhibited modest but consistent gains, improving from 0.758 to 0.760 internally (+0.2\%) and from 0.756 to 0.763 externally (+0.7\%; $P<0.001$ for external). For \textit{PIK3CA} in breast cancer, where morphological signatures are inherently subtle and heterogeneous, STAMP still secured meaningful improvements of +3.4\% internally (0.652 to 0.686) and +2.4\% externally (0.679 to 0.703; $P<0.001$ for both).

Analysis of predicted score distributions provided additional insight into model behavior across mutation types (Figure \ref{fig5}d). For all five targets, STAMP generated statistically distinct probability distributions between mutant and wild-type groups ($P<0.001$ by Mann-Whitney U test), reflecting mutation-specific biology and histological penetrance. \textit{IDH} mutant cases formed a near-bimodal pattern with minimal overlap against wild-type samples, consistent with the nuclear and architectural alterations associated with this alteration in gliomas. In contrast, \textit{PIK3CA} and \textit{KRAS} displayed broader score overlap between classes, yet mutant populations maintained consistently higher median predicted probabilities across all cancer types. This pattern aligns with the known heterogeneity of these alterations across tumor subtypes and microenvironmental contexts. Receiver operating characteristic (ROC) and precision-recall curves (PRC) further contextualized performance under realistic clinical class imbalance (Figure \ref{fig5}e). For mutations with clear morphological expression (\textit{IDH}, \textit{EGFR}), ROC curves showed stable discriminative power across internal and external cohorts. PRC addressed the challenge of low mutation prevalence in clinical populations. For \textit{IDH}, precision remained above 0.85 across recall thresholds from 0.0 to 0.9, reflecting the model's ability to prioritize true positives with minimal false alarms. For less prevalent alterations with weaker morphological signals (\textit{PIK3CA}, \textit{KRAS}, \textit{BRAF}), precision decreased at higher recall thresholds, reflecting the expected trade-off between sensitivity and false-positive rate. Critically, STAMP's PR curves remained above the baseline mutation prevalence across broad recall ranges in both internal and external cohorts. This indicates that the model enriches for true mutation carriers relative to random selection, even for targets with subtle histological correlates.

These findings establish STAMP as a promising computational predictor of actionable driver mutations from routine H\&E histology. While morphology-based inference is not intended to replace definitive NGS testing, the demonstrated enrichment for true positives shows that STAMP has potential utility as a rapid triage tool to prioritize cases for expedited molecular confirmation. By identifying high-probability mutation carriers directly from H\&E WSIs, STAMP could enable more efficient allocation of limited sequencing resources and accelerate time-to-treatment decisions for patients with actionable alterations.

\subsection*{Profiling of immunotherapy response indicators}
Immune checkpoint inhibitors have improved outcomes across multiple solid tumors, yet their clinical application is restricted to a subset of patients due to high treatment costs, potential immune-related toxicities, and highly variable response rates. Mismatch repair deficiency (dMMR) and high tumor mutational burden (TMB-H) are FDA-approved pan-tumor biomarkers that reflect elevated neoantigen load and enhanced tumor immunogenicity, informing patient eligibility for immunotherapy~\cite{le2017mismatch, marabelle2020association}. Standard assessment requires NGS or multiplexed IHC, both resource-intensive processes that can delay treatment initiation. Inferring MMR and TMB status from routine H\&E WSIs therefore represents a high-value clinical objective, enabling rapid, cost-effective patient stratification without additional wet-lab testing. To this end, we evaluated STAMP on four clinically relevant prediction tasks including TMB in lung and colorectal cancer, as well as MMR status in gastric and colorectal cancer.

Across the immunotherapy marker set (Figure \ref{fig5}f), internally, STAMP secured a mean AUC of 0.867, outpacing the Virchow2 (0.842) by a consistent margin while simultaneously surpassing established vision-centric and multimodal encoders, including UNI (0.821), CONCH (0.784), PLIP (0.681), and OmiCLIP (0.654). On the external validation, STAMP maintained its performance edge, achieving an aggregate AUC of 0.817 compared to 0.795 for the baseline. The performance elevation across all four immunotherapy biomarkers indicates that STAMP facilitates visual encoder to learn histological patterns associated with genomic instability and neoantigen load, which are critical determinants of immunotherapy responsiveness. Specifically, MMR status, characterized by well-documented histological correlates including prominent tumor-infiltrating lymphocytes and mucinous architectural shifts, generated the most robust classification signals. STAMP resolved MMR prediction in gastric cancer with an AUC of 0.939 (95\% CI: 0.843-0.997) in internal testing and 0.935 (0.852-0.994) externally, yielding absolute gains of +2.1\% and +2.2\% over Virchow2 ($P<0.001$ for both; \exttabref{immune-internal}, \exttabref{immune-external}). A comparable improvement emerged for MMR prediction in colorectal cancer, where STAMP attained an AUC of 0.859 (95\% CI: 0.764-0.934) internally and 0.845 (95\% CI: 0.780-0.897) externally, corresponding to baseline improvements of +3.1\% and +2.1\% ($P<0.001$). In contrast, TMB prediction is more challenging and shows variable performance across cancer types. For TMB prediction in colorectal cancer, STAMP recorded 0.885 (0.785-0.970) internally and 0.802 (0.744-0.853) externally, exceeding the baseline by +3.8\% and +0.6\% ($P<0.001$ for internal). However, STAMP only achieved modest performance for TMB prediction in lung cancer, with AUCs of 0.784 (0.721-0.846) internally and 0.686 (0.623-0.742) externally.

To contextualize these performance differences within a biological framework, we examined cohort-specific clinical cutoffs and the underlying continuous TMB distributions (Figure \ref{fig5}i). The TMB threshold is typically set much lower in lung cancer ($\ge 10$ mut/Mb) than in colorectal cancer ($\ge 20$ mut/Mb). In lung cancer, TMB values follow a continuous and unimodal distribution shaped, meaning the $\ge 10$ mut/Mb cutoff results in considerable morphological overlap near the decision boundary~\cite{chalmers2017analysis, mcgrail2021high}. Conversely, CRC exhibits a relatively even distribution across a broad spectrum of TMB values in which the $\ge 20$ mut/Mb threshold isolates a biologically divergent hypermutated population~\cite{cancer2012comprehensive}. To assess whether STAMP learns this underlying biological continuum, we evaluated model prediction scores across distinct TMB quartiles (Figure \ref{fig5}j). The exact distribution of these scores mirrored the cancer-type-specific mutational biology. Lung cancer predictions showed a steady, incremental rise from Q1 to Q4, consistent with the gradual and continuous accumulation of somatic mutations ($P < 0.001$, Kruskal-Wallis H-test). In contrast, CRC scores remained relatively low and flat through Q1-Q3, exhibiting a sudden, marked increase exclusively in Q4, aligning with the hypermutated phenotype characteristic of tumors. Collectively, these findings suggest that ST alignment enables a vision-based model to encode genuine histomolecular signatures of genomic instability, establishing STAMP as a potential screening adjunct prior to immunotherapy administration.

% In both lung and colorectal cohorts, STAMP scores increased monotonically across all quartiles ($P < 0.001$, Kruskal-Wallis H-test). The exact distribution of these scores mirrored the cancer-type-specific mutational biology. Lung cancer predictions showed a steady, incremental rise from Q1 to Q4, consistent with the gradual and continuous accumulation of somatic mutations. In contrast, CRC scores remained relatively low and flat through Q1-Q3, exhibiting a sudden, marked increase exclusively in Q4, aligning with the hypermutated phenotype characteristic of tumors. Collectively, these findings suggest that ST alignment enables a vision-based model to encode genuine histomolecular signatures of genomic instability, establishing STAMP as an effective screening adjunct prior to immunotherapy administration.

\begin{figure}[!htbp]
	\centering
	\includegraphics[width=\linewidth]{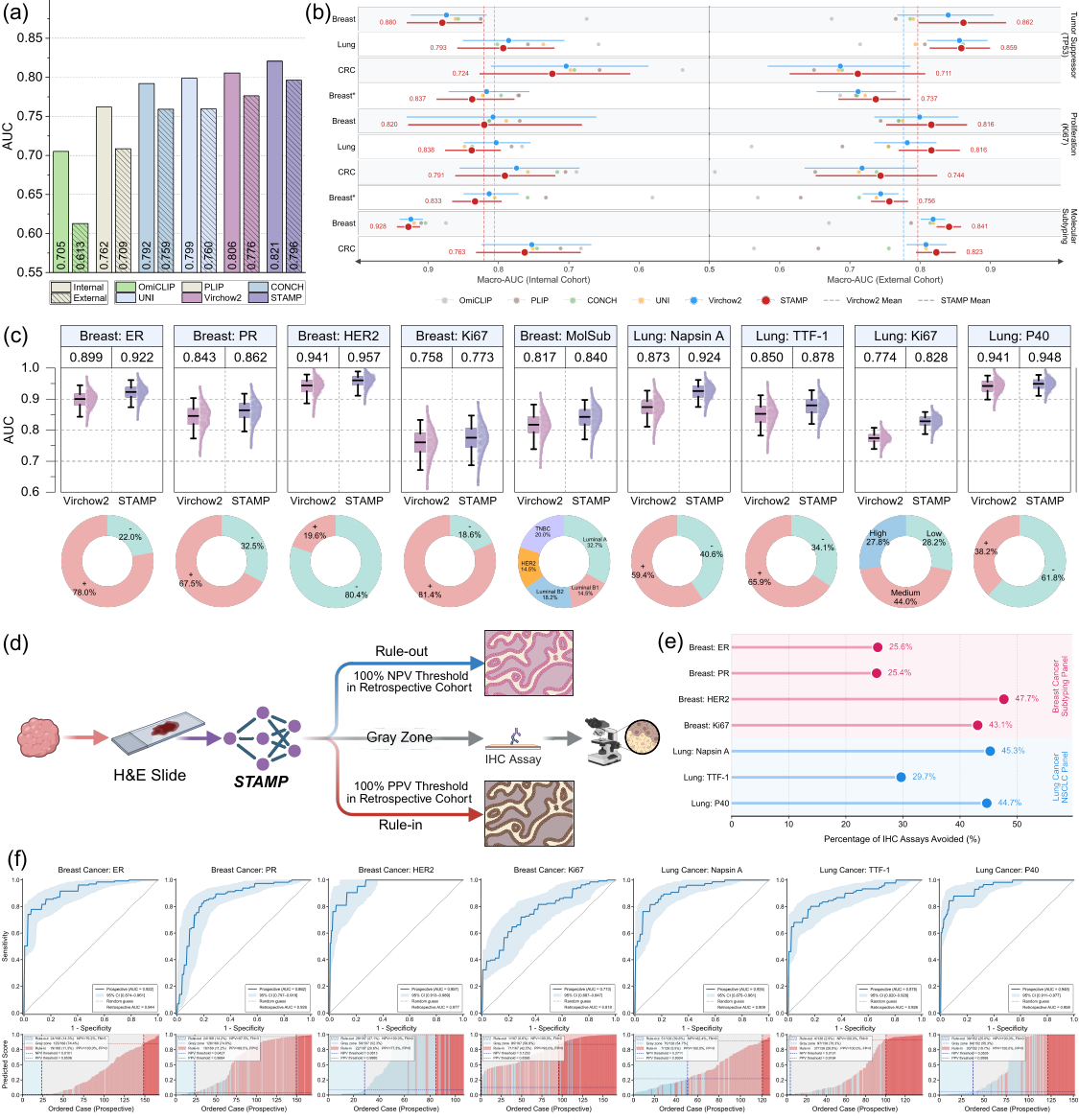}
	\caption{\textbf{Evaluation of molecular prognostic signatures and assessment of prospective clinical utility.} (\textbf{a}) Overall predictive performance for molecular prognostic signatures, encompassing TP53 mutation status, Ki-67 proliferation index, and molecular subtyping, across internal and external cohorts. (\textbf{b}) Detailed performance benchmarking of STAMP versus baseline models for individual prognostic targets on the internal (left) and external (right) cohorts. (\textbf{c}) Performance distributions comparing STAMP and the Virchow2 baseline across the diagnostic and prognostic panel in the prospective observational cohorts. (\textbf{d}) Schematic workflow for evaluating the clinical utility of STAMP in the prospective setting. (\textbf{e}) Quantitative assessment of clinical benefits and workflow efficiency gains enabled by the STAMP predictions. (\textbf{f}) Receiver operating characteristic (ROC) curves and predicted logit waterfall plots demonstrating individual biomarker performance and patient stratification in the prospective cohorts.}
	\label{fig6}
\end{figure}

\subsection*{Prediction of molecular prognostic signatures}
Accurate prognostic stratification is fundamental to precision oncology, as it dictates surveillance intensity, adjuvant therapy selection, and long-term patient management. In clinical practice, there are several well-established molecular signatures that provide critical insights into tumor aggressiveness, metastatic potential, and overall patient outcomes, including but not limited to, the mutation status of \textit{TP53}, the Ki-67 proliferation index, and molecular subtypes. Here, we evaluated STAMP's ability to predict these prognostic signatures across breast, lung, and colorectal cancers.

Over these molecular prognostic signatures, STAMP consistently outperformed the Virchow2 baseline and other PFMs in both internal and external validation settings (Figure \ref{fig6}a). On the internal test sets, STAMP achieved a mean AUC of 0.821, yielding a consistent absolute improvement over the Virchow2 baseline (mean AUC 0.806). It also outperformed other PFMs including UNI (0.799), CONCH (0.792), PLIP (0.762), and OmiCLIP (0.705). Most importantly, this prognostic sensitivity generalized well to the independent external validation sets. Task-level analysis uncovers the mechanism of these aggregate improvements across distinct prognostic dimensions (Figure \ref{fig6}b, Extended Data Table \ref{prognostic-internal}-\ref{prognostic-external}).  For the assessment of tumor proliferation, STAMP demonstrated high precision in stratifying the clinically actionable high versus low Ki-67 expression status. Notably, in lung cancer, STAMP unlocked an absolute AUC gain of +3.5\% over Virchow2 in both the internal (0.838 versus 0.803) and external (0.816 versus 0.781) cohorts. It also secured external enhancements for predicting clinically high Ki-67 in colorectal cancer (+2.7\%, AUC 0.744) and in challenging breast cancer core needle biopsies (+2.5\%, AUC 0.737). Similarly, predicting the mutation of \textit{TP53}, a gene notorious for causing diffuse genomic instability and varied morphological alterations, benefited from ST alignment. STAMP yielded meaningful absolute margins in predicting \textit{TP53} mutations externally, achieving AUCs of 0.711 in colorectal cancer (+2.5\% over baseline) and 0.862 in breast cancer (+2.2\% over baseline).

Furthermore, STAMP showed proficiency in predicting molecular subtypes. For breast cancer molecular subtyping~\cite{parker2009supervised}, STAMP demonstrated consistent superiority. On challenging core needle biopsies, STAMP achieved an internal AUC of 0.833 (a +2.0\% absolute increase over Virchow2's 0.813) and generalized to the external cohort with an AUC of 0.756 (+1.2\% over baseline). When evaluating surgical resection specimens, predictive performance internally peaked at an AUC of 0.928 (+0.3\% over baseline), and secured a substantial absolute gain of +2.3\% during independent external validation (AUC 0.841 versus 0.818). Parallel enhancements were observed in predicting consensus molecular subtypes (CMS)~\cite{guinney2015consensus} of colorectal cancer, with STAMP yielding an internal AUC of 0.763 (+1.0\% over baseline) and extending this margin externally to reach 0.823 (a +1.5\% absolute gain over Virchow2's 0.808). By learning the histological patterns associated with these prognostic signatures, STAMP effectively captures the underlying morpho-molecular relationships that govern tumor biology and clinical behavior. The performance gains across multiple prognostic dimensions and cancer types, as well as the robust external validation, confirm that ST supervision enhances the model's ability to recognize histological features linked to these prognostic markers.

\subsection*{Clinical utility in prospective observational cohorts}
While retrospective evaluations establish foundational model generalizability, the translational benchmark is its deployment within real-world, prospective clinical environments. To verify the translational robustness of the molecular inductive biases injected by ST alignment, we deployed STAMP in a prospective observational workflow. We evaluated its predictive performance on a comprehensive panel of nine actionable markers routinely assessed in daily practice: ER, PR, HER2, Ki-67, and molecular subtyping in breast cancer, alongside Napsin A, TTF-1, p40, and Ki-67 in lung cancer. On these prospective cohorts, STAMP consistently maintained an absolute advantage across the entire prospective panel (Figure \ref{fig6}c, \exttabref{prospective}), confirming that cross-modal ST supervision confers intrinsic resilience to real-world clinical variance. In breast cancer, STAMP achieved outstanding prospective AUCs of 0.957 for HER2 and 0.922 for ER, delivering absolute improvements of +1.6\% and +2.3\% over Virchow2 (0.941 and 0.899), respectively. It also provided steady baseline enhancements for PR (+1.9\%, AUC 0.862), Ki-67 (+1.5\%, AUC 0.773), and molecular subtyping (+2.3\%, AUC 0.840). In the lung cancer cohort, the ST-guided alignment effectively unlocked latent predictive potential, yielding profound absolute AUC gains of +5.1\% for Napsin A (0.924 versus 0.873) and +5.4\% for Ki-67 (0.828 versus 0.774) over Virchow2, while reaching an exceptional AUC of 0.948 for the squamous marker p40.

To translate these prospective AUCs into tangible clinical workflow optimizations, we conceptualized an AI-driven triage pipeline designed to safely minimize reliance on physical IHC assays (Figure \ref{fig6}d). In this pipeline, we established dual confidence thresholds for each biomarker based on the predicted logit distributions in retrospective validation. Specifically, we use confidence bounds calibrated to target 100\% negative predictive value (100\% NPV, Rule-out zone) and 100\% positive predictive value (100\% PPV, Rule-in zone) on the retrospective cohorts. Cases falling within these confident extremes are computationally diagnosed without further wet-lab assays, whereas biologically equivocal cases falling into the intermediate ``gray zone'' are reflexed to conventional IHC testing and pathologist review. By applying these thresholds to the prospective cohorts, we can estimate the potential reduction in IHC testing volume based on STAMP's predictions (Figure \ref{fig6}e). In breast cancer, STAMP's confident predictions could bypass 25.6\% of ER (5 FNs and 0 FP), 25.4\% of PR (3 FNs and 2 FPs), 47.7\% of HER2 (0 FN and 5 FPs), and 43.1\% (0 FN and 5 FPs) of Ki-67 assays. In lung cancer, STAMP's confident predictions could bypass 45.3\% (9 FNs and 0 FP) of Napsin A, 29.7\% (0 FN and 0 FP) of TTF-1, and 44.7\% (0 FN and 0 FP) of p40 stains. The ROC curves and predicted logit distributions further illustrate the model's discriminative power and patient stratification capabilities in the prospective setting (Figure \ref{fig6}f). Although we adopted conservative thresholds to minimize the risk of false negatives and false positives, there still exists a significant proportion of cases that could be confidently classified by STAMP, thereby reducing the need for additional IHC testing. Notably, as for p40 and CK7 in lung cancer, our model can safely rule in or rule out a substantial fraction of cases with 100\%NPV and 100\% PPV, which could directly expedite diagnostic workflows by eliminating the need for confirmatory IHC testing in these patients. As for other markers, the over-treatment associated with false positives or missed diagnosis associated with false negatives can be mitigated by using more conservative thresholds to further reduce the risk of misclassification, albeit at the cost of fewer cases being confidently classified. These findings underscore the potential of STAMP to enhance clinical efficiency by reducing unnecessary testing and accelerating diagnostic workflows, ultimately facilitating more timely and personalized patient care.

\section*{Discussion}
The deployment of PFMs has substantially advanced automated histological assessment, yet inferring molecular status and genomic alterations from routine H\&E slides remains a significant translational challenge. Conventional PFMs are predominantly trained using vision-centric self-supervised learning or aligned with clinical diagnostic reports. While these approaches capture general tissue morphology, they often lack direct biological context and struggle to resolve the subtle, spatially distributed histological features linked to specific genomic drivers. To address this limitation, this study presented STAMP, a framework that leverages ST as a biologically grounded, cross-modal supervisory signal. Unlike vision-language models trained on diagnostic texts that lack spatial resolution, ST provides localized, region-specific transcriptomic profiles. Importantly, rather than relying on sparse, noise-prone raw gene counts, the framework aggregates transcriptomic profiles into functionally biological pathways. This approach reduces feature dimensionality, mitigates technical noise, and encourages the visual encoder to learn functionally coherent and robust spatial representations~\cite{sanchez2018oncogenic}. Across 24 clinical tasks spanning multi-center retrospective and prospective cohorts, STAMP consistently elevated the predictive performance of pre-trained PFMs. These results confirm that there are morpho-molecular correlations contained within routine histology, and such latent relationships can be bridged by the guidance of spatially resolved transcriptomic data.

Modern precision oncology requires multidimensional tumor characterization, including diagnostic immunohistochemical biomarkers, actionable driver mutations, immunotherapy response indicators, and molecular prognostic signatures. In current practice, patients typically undergo sequential IHC panels followed by NGS. This workflow is constrained by high costs, complex logistics, and prolonged turnaround times. For patients who need rapid treatment decisions, such delays can postpone appropriate interventions. By predicting molecular profiles from H\&E slides, STAMP provides a scalable, tissue-conserving computational alternative. Importantly, our systematic evaluation clarifies the biological boundaries of morphological predictability. The model performs reliably for genomic alterations that produce clear histological phenotypes, whereas prediction accuracy declines for functionally silent or microscopically subtle molecular events~\cite{kather2020pan}. We therefore propose using STAMP as a computational triage and decision-support tool, rather than a complete replacement for molecular assays. By quantifying predictive uncertainty, STAMP can identify high-probability cases for confirmatory testing, which can help optimize NGS resource allocation and preserve limited biopsy tissue. Furthermore, for complex immunological phenotypes, STAMP quantifies continuous biological gradients rather than relying on fixed binary thresholds. This continuous stratification may improve patient selection for immune checkpoint inhibitors, where standard binary classifications often fail to reflect the full range of therapeutic responses~\cite{yarchoan2017tumor, litchfield2021meta}.

% TODO: 继续check
To evaluate real-world clinical utility, our prospective observational study moved beyond retrospective benchmarking to assess STAMP within routine diagnostic workflows. By testing the model on a comprehensive panel of breast and lung cancer biomarkers, we demonstrated that the molecular inductive biases learned through ST alignment provide stable performance across prospective clinical settings. A key finding from this cohort is the practical validation of an uncertainty-aware triage framework. Instead of applying single arbitrary cutoffs, we established calibrated decision boundaries designed to maximize positive and negative predictive values within predefined confidence intervals. This dual-threshold approach creates a structured system for ruling in and ruling out specific molecular alterations. Cases falling within these high-confidence intervals can be predicted with substantial certainty, potentially allowing laboratories to defer routine confirmatory testing~\cite{kleppe2021designing}. In contrast, cases with intermediate prediction probabilities are automatically flagged for standard IHC testing and review by expert pathologists. This workflow directly links algorithmic uncertainty to appropriate clinical actions. Our prospective data indicate that this approach can safely reduce the number of standard physical assays for highly predictive markers, while maintaining diagnostic accuracy. From a translational standpoint, this targeted reduction in laboratory testing shifts the operational workflow of precision oncology. It helps reduce the financial and logistical demands of comprehensive IHC profiling and shortens turnaround times for time-sensitive treatment decisions~\cite{bera2019artificial, niazi2019digital}. In practice, STAMP functions as a workflow optimization tool that conserves laboratory resources and allows pathologists to focus their expertise on complex or ambiguous cases.

Although STAMP establishes a structured approach for aligning morphology with molecular profiles, several areas require further development. As precision oncology advances, a growing number of novel therapeutic targets and biomarkers are being identified across different cancer types. While our study validated a comprehensive set of established clinical markers, this panel represents only a portion of the broader molecular landscape. Future work should systematically evaluate these emerging targets to map the predictive boundaries of computational pathology. This will help clarify which molecular changes consistently produce detectable morphological features on H\&E slides, and which remain below the resolution of routine histology. To extend these predictive boundaries, the underlying cross-modal training framework will need to expand. The ongoing development of spatial multi-omics technologies provides opportunities to move beyond transcriptomics alone~\cite{vandereyken2023methods}. Integrating spatial proteomics and spatial metabolomics could introduce additional biochemical context into the visual encoder, offering more spatial supervision signals. Additionally, the scale of cross-modal training data could be expanded by incorporating non-human datasets. Large public repositories of spatial data from mouse tumor models are currently underutilized. However, direct cross-species integration requires careful handling of biological and technical differences. By applying rigorous cross-species gene mapping and domain adaptation techniques, future versions of STAMP could incorporate evolutionarily conserved pathways to increase pre-training data volume and improve model generalization. In summary, by providing a scalable and biologically informed framework for morpho-molecular alignment, STAMP facilitates broader access to molecular profiling, supporting more timely, precise, and tissue-efficient clinical decision-making in modern oncology.

\section*{Methods}
\subsection*{HumanST-1k Dataset Curation and Pre-processing}
To establish a foundational resource for morpho-molecular alignment, we assembled the HumanST-1k dataset by integrating publicly available human (\textit{Homo sapiens}) ST profiles with their corresponding WSIs. Data were aggregated from multiple repositories, including 10$\times$ Genomics datasets, the China National Center for Bioinformation (CNCB), Dryad, the European Molecular Biology Laboratory (EMBL), the National Center for Biotechnology Information (NCBI), Zenodo, Mendeley, \textit{etc}. The resulting collection encompasses several spatial sequencing platforms with varying capture resolutions, ranging from legacy Spatial Transcriptomics (ST) arrays and standard Visium to ultra-high-resolution Visium HD and \textit{in situ} sequencing platform Xenium. Because accurate phenotype mapping requires sufficient histological context, we retained only samples with WSI dimensions of at least 3,000 $\times$ 3,000 pixels. To align the spatial granularities across platforms, we standardized ultra-high-resolution data by binning transcript counts into 100 $\mu$m pseudo-spots. This procedure approximates the spatial footprint of conventional spot-based arrays, creating a consistent coordinate system for downstream alignment.

Given that identical clinical cohorts are frequently deposited across public databases, we implemented a strict duplicate-removal protocol via MD5 checksum verification to preclude data redundancy. A sample was classified as redundant and excluded only when the cryptographic hashes for both the raw WSI file and its corresponding spatial expression matrix were strictly identical. Following deduplication, we standardized the molecular annotations by mapping all transcript identifiers to official HGNC (HUGO Gene Nomenclature Committee) gene symbols. These curation steps resulted in a finalized dataset of 1,004 non-redundant H\&E-ST sample pairs. Dataset partitioning was then performed to guarantee the integrity of downstream evaluations and absolutely prevent data leakage. Samples allocated for gene expression prediction and tumor domain recognition were held out, while the remaining samples constituted the pre-training corpus. Prior to model pre-training, a stringent spot-level quality control filter was applied to maximize the signal-to-noise ratio. Specifically, any spatial spot or pseudo-spot expressing fewer than 100 detected pathway-related genes was discarded, eliminating technical sequencing dropouts and acellular tissue regions. This curation pipeline ultimately produced a pre-training corpus containing approximately 1.8 million paired H\&E image patches coupled with their spatially resolved transcriptomic profiles.

\subsection*{STAMP Architecture and Pre-training Paradigm}
To endow underlying PFMs with molecular sensitivity, we designed the \textbf{S}patial \textbf{T}ranscriptomics-guided \textbf{A}lignment for \textbf{M}olecular \textbf{P}rofiling (\textbf{STAMP}) framework. This framework consists of three core components: (1) a pathway-informed representation module that transforms sparse transcriptomic profiles into robust functional embeddings; (2) a parameter-efficient fine-tuning strategy that enables the visual encoder to learn molecularly sensitive representations without catastrophic forgetting; and (3) a cross-modal pathway decoder that constraints the visual features to encode molecular information. These components are jointly optimized through a compound objective function that balances semantic alignment with molecular reconstruction, effectively bridging the modality gap between histological morphology and spatially resolved transcriptomics.

\noindent\textbf{Biologically Informed Pathway Representation.} Raw ST profiles are typically characterized by high dimensionality, inherent sparsity, and technical noise. To associate morphological features with robust functional states rather than noisy gene-level counts, we aggregate transcriptomic profiles into $K=50$ Hallmark biological pathways from MSigDB~\cite{liberzon2015molecular}. Let $\mathcal{G} = \{g_1, \dots, g_{|\mathcal{G}|}\}$ denote the set of all detected genes. After applying a log-transform to raw counts for variance stabilization, we define the pathway collection $\mathcal{P} = \{P_1, \dots, P_K\}$, where each $P_k \subseteq \mathcal{G}$ represents a curated gene set. For pathway $k$, let $\mathbf{e}_k \in \mathbb{R}^{|P_k|}$ denote the log-transformed expression vector restricted to genes in $P_k$. To handle variable pathway sizes, we employ pathway-specific self-normalizing networks~\cite{klambauer2017self} that project $\mathbf{e}_k$ into a unified $d$-dimensional latent space:
\begin{equation*}
	\mathbf{p}_k = \phi_k(\mathbf{e}_k) \in \mathbb{R}^d, \quad k = 1, \dots, K,
\end{equation*}
where $\phi_k(\cdot)$ denotes a learnable mapping consisting of linear transformation followed by SELU activation. A Transformer encoder $\Psi(\cdot)$~\cite{vaswani2017attention} then models inter-pathway dependencies:
\begin{equation*}
	[\mathbf{m}, \mathbf{h}^{\text{mol}}_1, \dots, \mathbf{h}^{\text{mol}}_K] = \Psi\big([\mathbf{p}_{\text{cls}}, \mathbf{p}_1, \dots, \mathbf{p}_K]\big),
\end{equation*}
where $\mathbf{p}_{\text{cls}} \in \mathbb{R}^d$ is a learnable molecular class token, $\mathbf{m} \in \mathbb{R}^d$ is its final hidden state serving as the global molecular representation, and $\{\mathbf{h}^{\text{mol}}_k\}_{k=1}^K$ are pathway-level contextual embeddings.

\noindent\textbf{Visual Encoder with Parameter-Efficient Fine-Tuning.}
We adopt Virchow2~\cite{vorontsov2024foundation, zimmermann2024virchow2} as the frozen visual backbone. As a state-of-the-art PFM built upon the DINOv2 self-supervised architecture~\cite{oquab2023dinov2}, Virchow2 was pre-trained on 2 billion patches from 3 million WSIs, providing strong generalization capabilities across diverse cancer types and morphological patterns. To enable molecular adaptation without catastrophic forgetting, we integrate low-rank adaptation (LoRA)~\cite{hu2022lora} exclusively into the query and value projection matrices of multi-head self-attention blocks. For a frozen weight matrix $\mathbf{W}_0 \in \mathbb{R}^{d_{\text{in}} \times d_{\text{out}}}$, the adapted weight is:
\begin{equation}
	\mathbf{W} = \mathbf{W}_0 + \frac{\alpha}{r} \mathbf{B}\mathbf{A},
	\label{eq:lora_update}
\end{equation}
where $\mathbf{B} \in \mathbb{R}^{d_{\text{in}} \times r}$ and $\mathbf{A} \in \mathbb{R}^{r \times d_{\text{out}}}$ are trainable matrices, bottleneck rank $r=8$, scaling factor $\alpha=16$, and LoRA dropout $0.1$. This reduces trainable parameters from 632M to $\sim$2M (0.31\%). For an H\&E patch $\mathbf{x}^{\text{vis}} \in \mathbb{R}^{H \times W \times C}$, the adapted encoder yields:
\begin{equation}
	[\mathbf{v}, \mathbf{h}^{\text{vis}}_1, \dots, \mathbf{h}^{\text{vis}}_N] = \Phi_{\text{LoRA}}(\mathbf{x}^{\text{vis}}),
	\label{eq:visual_encoder}
\end{equation}
where $\mathbf{v} \in \mathbb{R}^d$ is the visual class token state (linearly projected to match the molecular embedding dimension), and $\{\mathbf{h}^{\text{vis}}_n\}_{n=1}^N$ are patch-level visual embeddings.

\noindent\textbf{Cross-modal Decoder for Pathway Reconstruction.}
While the global contrastive objective establishes alignment between morphological and transcriptomic representations, it operates in the latent space. To encourage fine-grained molecular representation within the visual encoder, we introduced an auxiliary cross-attention decoder. This module is designed to guide the visual tokens in reconstructing the transcriptional signals underlying each biological pathway. Specifically, we leverage the standard cross-attention mechanism. Learnable pathway queries $\{\mathbf{q}_k \in \mathbb{R}^d\}_{k=1}^K$ serve as semantic anchors for each Hallmark process. Within Transformer decoder layers, these queries attend to visual embeddings via cross-attention:
\begin{equation}
	\mathbf{z}_k = \text{CrossAttn}\big(\mathbf{q}_k, \{\mathbf{h}^{\text{vis}}_n\}_{n=1}^N\big) \in \mathbb{R}^d,
	\label{eq:cross_attention}
\end{equation}
where $\text{CrossAttn}(\cdot)$ denotes multi-head cross-attention with $\mathbf{q}_k$ as queries and $\{\mathbf{h}^{\text{vis}}_n\}$ as keys/values. A pathway-specific prediction head $\psi_k(\cdot)$ then reconstructs the log-transformed expression vector:
\begin{equation}
	\hat{\mathbf{e}}_k = \psi_k(\mathbf{z}_k) \in \mathbb{R}^{|P_k|}.
	\label{eq:reconstruction_head}
\end{equation}

\noindent\textbf{Cross-Modal Alignment and Optimization Objectives.} To achieve robust morpho-molecular alignment, the STAMP pre-training paradigm jointly optimizes a compound objective function. This objective balances global semantic matching across modalities with the reconstruction of specific biological pathways, ensuring the visual encoder captures both overarching tissue phenotypes and fine-grained transcriptomic details. To establish global semantic correspondence, the alignment objective ($\mathcal{L}_{\text{con}}$) adopts the standard symmetric cross-entropy loss paradigm. Let $\mathbf{v}_i \in \mathbb{R}^d$ and $\mathbf{m}_i \in \mathbb{R}^d$ denote the L2-normalized global visual and molecular embeddings for the $i$-th sample within a training mini-batch of size $B$. Here, $\mathbf{v}_i$ is derived by linearly projecting the visual class token state ($\mathbf{v}_{\text{cls}}$) into the shared latent space, aligning its dimensionality with $\mathbf{m}_i$. Let $\ell_{v \rightarrow m}^{(i)}$ denote the asymmetric InfoNCE loss~\cite{oord2018representation, radford2021learning} that maximizes the temperature-scaled cosine similarity between the matched pair $(\mathbf{v}_i, \mathbf{m}_i)$ against all negative samples in the batch. The symmetric contrastive alignment loss $\mathcal{L}_{\text{con}}$ is then seamlessly computed by averaging the bidirectional losses,
\begin{equation*}
	\begin{aligned}
		\mathcal{L}_{\text{con}} = \frac{1}{2B} \sum_{i=1}^{B} \left( \ell_{v \rightarrow m}^{(i)} + \ell_{m \rightarrow v}^{(i)} \right).
	\end{aligned}
\end{equation*}

While the contrastive objective facilitates broad feature alignment, it may miss fine-grained quantitative biological variations. To explicitly enforce granular precision, the auxiliary pathway reconstruction loss ($\mathcal{L}_{\text{rec}}$) evaluates the fidelity of the cross-modal decoder. For a given sample, this loss is computed by calculating the Mean Squared Error (MSE) between the predicted gene expression sub-vectors $\hat{\mathbf{e}}^{(k)}$ and their corresponding ground-truth targets $\mathbf{e}^{(k)}$ across all $K$ pathways:
\begin{equation*}
	\begin{aligned}
		\mathcal{L}_{\text{rec}} = \frac{1}{K} \sum_{k=1}^{K} || \hat{\mathbf{e}}^{(k)} - \mathbf{e}^{(k)} ||_2^2,
	\end{aligned}
\end{equation*}

Ultimately, the STAMP architecture is trained end-to-end by minimizing the total compound objective, which integrates these two complementary constraints,
\begin{equation*}
	\begin{aligned}
		\mathcal{L}_{\text{Total}} = \mathcal{L}_{\text{con}} + \lambda \mathcal{L}_{\text{rec}},
	\end{aligned}
\end{equation*}
where $\lambda$ is an empirically derived hyperparameter determining the relative optimization contribution. This joint optimization paradigm ensures that the foundational visual features are endowed with robust molecular sensitivity.

\subsection*{Clinical Cohort Definitions and Data Processing}
To validate the effectiveness of the STAMP framework, we conducted a comprehensive evaluation across multiple biological scales and translational phases. This evaluation ecosystem ranges from pan-cancer ST datasets for microenvironmental benchmarking, to multi-centric WSI cohorts for retrospective clinical biomarker inference, and ultimately to real-world prospective cohorts for clinical triage simulation.

\noindent\textbf{Spatial Transcriptomics Prediction.} We conducted this evaluation using the established HEST-Benchmark~\cite{jaume2024hest}. There are eight distinct cancer types involved in this task, comprising clear cell renal cell carcinoma (CCRCC; 24 Visium samples, 73,167 spots), colon adenocarcinoma (COAD; 4 Xenium samples, 15,651 spots), invasive ductal carcinoma (IDC; 4 Xenium samples, 35,536 spots), lung cancer (LUNG; 2 Xenium samples, 5,206 spots), pancreatic adenocarcinoma (PAAD; 3 Xenium samples, 7,571 spots), prostate adenocarcinoma (PRAD; 23 Visium samples, 62,710 spots), rectal adenocarcinoma (READ; 4 Visium samples, 8,407 spots), and skin cutaneous melanoma (SKCM; 2 Xenium samples, 3,034 spots). For each cohort, we extracted the top 50 highly variable genes (HVGs) based on their expression variance across all spatial spots. These HVGs served as the targeted molecular variables for gene expression prediction.

\noindent\textbf{Spatial Domain Recognition.} We benchmarked spatially resolved domain recognition across four publicly available ST datasets spanning distinct cancer types. The breast cancer dataset~\cite{andersson2020spatial} comprises 8 samples from HER2-positive invasive breast carcinoma, with expert annotations delineating tumor epithelium, stroma, adipose tissue, immune infiltrates, and connective tissue. The lung cancer dataset~\cite{dawo_10x_2025} includes 5 samples, with domain classes spanning tumor glands, alveolar parenchyma, bronchiolar epithelium, vessels, and stroma. The kidney cancer dataset~\cite{dawo_10x_2025} contains 3 samples, annotated with tumor, stroma, normal tubular epithelium, and vascular compartments. The prostate dataset~\cite{erickson2022spatially} consists of 7 samples annotated with Gleason-grade tumor classes (GG1, GG2, GG4 Cribriform), benign glands, stroma, chronic inflammation, and vascular tissue. Together, these 23 samples present a broad spectrum of tissue architectures, annotation granularities, and domain class imbalances, providing a comprehensive multi-disease benchmark.

\noindent\textbf{Core Diagnostic Biomarkers.} For breast cancer, we evaluated the prediction of estrogen receptor (ER), progesterone receptor (PR), and human epidermal growth factor receptor 2 (HER2) status. These core biomarkers govern the molecular subtyping of breast carcinomas and form the definitive basis for targeted therapeutic interventions. To account for tissue volume disparities in routine clinical workflows, we evaluated these markers independently on core needle biopsy and surgical resection specimens.

\begin{itemize}
	\item \textbf{ER Status: } ER is the defining transcriptional driver of hormone-receptor-positive breast cancer and the primary prerequisite for determining patient eligibility for endocrine therapies.
	      \begin{itemize}
		      \item \textit{Biopsy Specimens:} The internal cohort consisted of 1,485 samples (3,502 slides) collected from private H3 center, with 1,011 ER-positive and 474 ER-negative cases. The internal cohort was randomly stratified into training (1,039 samples), validation (149 samples), and testing (297 samples) sets. The external cohort included 703 samples (703 slides) collected from private H9 center, with 529 ER-positive and 174 ER-negative cases.
		      \item \textit{Resection Specimens:} The internal cohort consisted of 1,542 samples (2,024 slides) collected from private H2 center, with 780 ER-positive and 762 ER-negative cases. The internal cohort was randomly stratified into training (1,083 samples), validation (155 samples), and testing (304 samples) sets. The external cohort included 333 samples (3,811 slides) collected from private H1 center, with 231 ER-positive and 102 ER-negative cases.
	      \end{itemize}

	\item \textbf{PR Status: } As an estrogen-regulated gene, PR serves as a critical surrogate marker for an intact and functional ER signaling pathway, offering independent prognostic value regarding survival and therapeutic response.
	      \begin{itemize}
		      \item \textit{Biopsy Specimens:} The internal cohort consisted of 1,483 samples (3,495 slides) collected from private H3 center, with 1,119 PR-positive and 364 PR-negative cases. The internal cohort was randomly stratified into training (1,037 samples), validation (149 samples), and testing (297 samples) sets. The external cohort included 703 samples (703 slides) collected from private H9 center, with 465 PR-positive and 238 PR-negative cases.
		      \item \textit{Resection Specimens:} The internal cohort consisted of 1,552 samples (2,029 slides) collected from private H2 center, with 932 PR-positive and 620 PR-negative cases. The internal cohort was randomly stratified into training (1,088 samples), validation (156 samples), and testing (308 samples) sets. The external cohort included 310 samples (3,541 slides) collected from private H1 center, with 164 PR-positive and 146 PR-negative cases.
	      \end{itemize}

	\item \textbf{HER2 Status: } HER2 is a transmembrane tyrosine kinase receptor. Its amplification or overexpression drives a highly aggressive breast cancer phenotype that is uniquely susceptible to targeted anti-HER2 therapies (e.g., trastuzumab).
	      \begin{itemize}
		      \item \textit{Biopsy Specimens:} The internal cohort consisted of 1,331 samples (3,189 slides) collected from private H3 center, with 474 HER2-positive and 857 HER2-negative cases. The internal cohort was randomly stratified into training (931 samples), validation (133 samples), and testing (267 samples) sets. The external cohort included 703 samples (703 slides) collected from private H9 center, with 294 HER2-positive and 409 HER2-negative cases.
		      \item \textit{Resection Specimens:} The internal cohort consisted of 918 samples (1,164 slides) collected from private H2 center, with 408 HER2-positive and 510 HER2-negative cases. The internal cohort was randomly stratified into training (643 samples), validation (92 samples), and testing (183 samples) sets. The external cohort included 181 samples (2,095 slides) collected from private H1 center, with 59 HER2-positive and 122 HER2-negative cases.
	      \end{itemize}
\end{itemize}

For lung cancer, we focused on four core immunohistochemical biomarkers essential for the definitive histological subtyping of non-small cell lung cancer (NSCLC), enabling the  differentiation between adenocarcinomas and squamous cell carcinomas.
\begin{itemize}
	\item \textbf{Napsin A Status Prediction: } Napsin A is a functional aspartic proteinase involved in surfactant processing. It serves as a highly sensitive and specific diagnostic marker for primary lung adenocarcinoma. The internal cohort consisted of 695 samples (707 slides) collected from private H5 center, with 304 Napsin A-positive and 391 Napsin A-negative cases. The internal cohort was randomly stratified into training (485 samples), validation (70 samples), and testing (140 samples) sets. The external cohort included 92 samples (333 slides) collected from private H2 center, with 54 Napsin A-positive and 38 Napsin A-negative cases.

	\item \textbf{TTF-1 Status Prediction: } Thyroid transcription factor-1 (TTF-1) is a master transcriptional regulator of lung development. It acts as the cornerstone marker for distinguishing primary lung adenocarcinoma from both squamous cell carcinoma and distant extrapulmonary metastases. The internal cohort consisted of 1,163 samples (1,188 slides) collected from private H5 center, with 677 TTF-1-positive and 486 TTF-1-negative cases. The internal cohort was randomly stratified into training (816 samples), validation (117 samples), and testing (230 samples) sets. The external cohort included 558 samples (774 slides) collected from private H1 center, with 232 TTF-1-positive and 326 TTF-1-negative cases.

	\item \textbf{p40 Status Prediction: } p40, an isoform of p63, is exceptionally specific to basal and squamous epithelial lineages, making it the most reliable biomarker for the definitive diagnosis of pulmonary squamous cell carcinoma. The internal cohort consisted of 898 samples (917 slides) collected from private H5 center, with 357 p40-positive and 541 p40-negative cases. The internal cohort was randomly stratified into training (629 samples), validation (90 samples), and testing (179 samples) sets. The external cohort included 166 samples (557 slides) collected from private H2 center, with 57 p40-positive and 109 p40-negative cases.

	\item \textbf{CK7 Status Prediction: } Cytokeratin 7 (CK7) is a low-molecular-weight cytokeratin broadly expressed in glandular epithelia. In pulmonary pathology, it is routinely utilized alongside TTF-1 and p40 to verify non-squamous NSCLC lineages. The internal cohort consisted of 845 samples (863 slides) collected from private H5 center, with 619 CK7-positive and 226 CK7-negative cases. The internal cohort was randomly stratified into training (591 samples), validation (84 samples), and testing (170 samples) sets. The external cohort included 460 samples (774 slides) collected from private H1 center, with 242 CK7-positive and 218 CK7-negative cases.
\end{itemize}

\noindent\textbf{Actionable Driver Mutations.}
Beyond diagnostic phenotyping, the identification of underlying actionable driver mutations is the cornerstone of genotype-directed precision oncology. We assessed the capacity of STAMP to infer five clinically critical mutational statuses directly from H\&E morphology across breast, lung, colorectal, and brain cancers.

\begin{itemize}
	\item \textbf{\textit{PIK3CA} Mutation in Breast Cancer: } \textit{PIK3CA} is the most frequently mutated oncogene in breast cancer and serves as an indispensable predictive biomarker for targeted PI3K$\alpha$ inhibitors (e.g., alpelisib), particularly in hormone receptor-positive disease.
	      The internal cohort (TCGA-BRCA) comprised 1,013 samples across 1,080 slides (329 mutant, 684 wild-type). This cohort was stratified into training (708 samples), validation (102 samples), and testing (203 samples) sets. The independent external cohort (CPTAC-BRCA) included 116 samples across 362 slides (38 mutant, 78 wild-type).

	\item \textbf{\textit{EGFR} Mutation in Lung Cancer: } \textit{EGFR} mutations are canonical oncogenic drivers that definitively predict robust clinical responses to specific tyrosine kinase inhibitors (TKIs). Given the exceedingly low mutation prevalence of \textit{EGFR} in squamous cell carcinomas, our evaluation was exclusively restricted to lung adenocarcinomas (LUAD).
	      The internal cohort, collected from the private H6 center, consisted of 1,265 samples across 1,265 slides (677 mutant, 588 wild-type), partitioned into training (884 samples), validation (127 samples), and testing (254 samples) sets. The external cohort (TCGA-LUAD) comprised 452 samples across 515 slides (56 mutant, 396 wild-type).

	\item \textbf{\textit{KRAS} Mutation in Lung Cancer: } Historically considered an ``undruggable'' target, \textit{KRAS} is the most common oncogenic driver in lung adenocarcinoma and is now actionable via novel allele-specific covalent inhibitors. Similar to \textit{EGFR}, this task was restricted to LUAD cohorts.
	      The internal cohort, collected from the private H6 center, included 1,265 samples across 1,265 slides (154 mutant, 1,111 wild-type; stratified into 884 training, 127 validation, and 254 testing samples). The external cohort (CPTAC-LUAD) consisted of 223 samples across 1,050 slides (67 mutant, 156 wild-type).

	\item \textbf{\textit{BRAF} Mutation in Colorectal Cancer (CRC): } The \textit{BRAF} mutation (predominantly V600E) delineates a distinct, highly aggressive molecular subtype in CRC. It is a critical prognostic indicator that necessitates specific combinatorial targeted therapeutic regimens.
	      The combined internal cohort (incorporating TCGA-COAD and TCGA-READ) comprised 494 samples across 501 slides (59 mutant, 435 wild-type). The internal data was randomly stratified into training (345 samples), validation (50 samples), and testing (99 samples) sets. The external validation cohort (CPTAC-COAD) included 104 samples across 217 slides (15 mutant, 89 wild-type).

	\item \textbf{\textit{IDH} Mutation in Brain Tumors: } While Isocitrate dehydrogenase (\textit{IDH}) mutations were traditionally utilized as diagnostic and prognostic classifiers, they have recently emerged as transformative therapeutic targets in neuro-oncology. The development of IDH-specific inhibitors, most notably the FDA-approved dual IDH1/2 inhibitor \textit{Vorasidenib}, has shifted the treatment paradigm for low-grade gliomas toward early targeted intervention. The combined internal cohort (incorporating TCGA-LGG and TCGA-GBM) consisted of 543 samples across 968 slides (394 mutant, 149 wild-type), which was systematically divided into training (379 samples), validation (55 samples), and testing (109 samples) sets. The independent external cohort (eBrains) comprised 852 samples across 852 slides (322 mutant, 530 wild-type).
\end{itemize}

\noindent\textbf{Immunotherapy Response Indicators.}
The advent of immune checkpoint inhibitors (ICIs) has revolutionized oncology, yet identifying patients who will derive enduring clinical benefit remains a profound challenge. Tumor Mutational Burden (TMB) and Mismatch Repair (MMR) deficiency (or Microsatellite Instability, MSI) currently serve as the definitive, FDA-approved agnostic biomarkers for immunotherapy response. We evaluated STAMP’s capability to infer these complex, genome-wide instability signatures directly from morphological architectures across lung, colorectal, and gastric cancers.

\begin{itemize}
	\item \textbf{Tumor Mutational Burden (TMB) in Lung Cancer: } High TMB reflects a substantial neoantigen load, correlating strongly with enhanced immunogenicity and favorable responses to PD-1/PD-L1 blockade in non-small cell lung cancer (NSCLC). In accordance with established clinical guidelines, TMB-High status was strictly defined using the threshold of $\ge 10$ mutations/megabase (mut/Mb).
	      The internal cohort, collected from private H6 center, consisted of 1,523 samples across 1,523 slides (249 TMB-High, 1,274 TMB-Low). This cohort was stratified into training (1,065 samples), validation (153 samples), and testing (305 samples) sets. The independent external cohort (CPTAC-NSCLC, incorporating both LUAD and LSCC) comprised 330 samples across 1,515 slides (76 TMB-High, 254 TMB-Low).

	\item \textbf{Tumor Mutational Burden (TMB) in Colorectal Cancer (CRC): } In colorectal malignancies, the distribution of TMB is highly right-skewed. The clinically actionable hypermutated phenotype is captured using a higher threshold of $\ge 20$ mut/Mb, which effectively isolates patients most likely to exhibit durable responses to immunotherapy.
	      The internal cohort, collected from private H4 center, comprised 608 samples across 2,779 slides (83 TMB-High, 525 TMB-Low), partitioned into training (425 samples), validation (61 samples), and testing (122 samples) sets. The external cohort (TCGA-CRC, combining TCGA-COAD and TCGA-READ) included 494 samples across 501 slides (72 TMB-High, 422 TMB-Low).

	\item \textbf{Mismatch Repair (MMR) Status in Colorectal Cancer: } Deficient MMR (dMMR) or High Microsatellite Instability (MSI-H) fundamentally drives the hypermutated phenotype in CRC. It is a robust prognostic indicator and a definitive predictive biomarker for exceptional sensitivity to immunotherapy.
	      The internal cohort, collected from private H4 center, consisted of 595 samples across 2,706 slides (59 MSI/dMMR, 536 MSS/pMMR). This data was randomly stratified into training (416 samples), validation (59 samples), and testing (120 samples) sets. The independent external cohort, collected from private H7 center, comprised 855 samples across 856 slides (37 MSI/dMMR, 818 MSS/pMMR).

	\item \textbf{Mismatch Repair (MMR) Status in Gastric Cancer: } In gastric adenocarcinoma, dMMR delineates a distinct molecular subtype associated with a favorable prognosis and pronounced efficacy of first-line immunotherapy combinations.
	      The internal cohort, collected from private H2 center, included 637 samples across 3,790 slides (53 dMMR, 584 pMMR; partitioned into 445 training, 64 validation, and 128 testing samples). The independent external cohort, collected from private H1 center, consisted of 270 samples across 270 slides (19 dMMR, 251 pMMR).
\end{itemize}
\noindent\textbf{Key Prognostic Biomarkers.}

Beyond targeted therapeutics, accurately assessing key prognostic indicators is indispensable for predicting disease trajectory, estimating recurrence risks, and guiding adjuvant patient management. We established a rigorous evaluation framework to assess STAMP's capacity to infer multifaceted prognostic signatures, including the \textit{TP53} tumor suppressor mutation, the Ki-67 proliferation index, and broad molecular subtypes, across breast, lung, and colorectal cancers.
\begin{itemize}
	\item \textbf{\textit{TP53} Mutation Status: } As the ``guardian of the genome,'' \textit{TP53} is the most frequently mutated tumor suppressor gene across human cancers. Its mutation drives profound genomic instability, treatment resistance, and generally correlates with highly aggressive clinical phenotypes.
	      \begin{itemize}
		      \item \textit{Breast Cancer:} The internal cohort (TCGA-BRCA) comprised 1,013 samples across 1,080 slides (338 mutant, 675 wild-type), stratified into 708 training, 102 validation, and 203 testing samples. The external cohort (CPTAC-BRCA) included 116 samples across 362 slides (42 mutant, 74 wild-type).
		      \item \textit{Lung Cancer (NSCLC):} The combined internal cohort (TCGA-LUAD and TCGA-LUSC) consisted of 893 samples across 990 slides (602 mutant, 291 wild-type; partitioned into 624 training, 89 validation, and 180 testing samples). The external cohort (CPTAC-NSCLC, incorporating CPTAC-LUAD and CPTAC-LSCC) comprised 329 samples across 1,513 slides (205 mutant, 124 wild-type).
		      \item \textit{Colorectal Cancer (CRC):} The internal cohort (TCGA-COAD and TCGA-READ) included 494 samples across 501 slides (286 mutant, 208 wild-type), randomly stratified into training (345 samples), validation (49 samples), and testing (100 samples). The external cohort (CPTAC-COAD) consisted of 104 samples across 217 slides (57 mutant, 47 wild-type).
	      \end{itemize}

	\item \textbf{Ki-67 Proliferation Index: } Ki-67 is a strict cellular marker for proliferation. Evaluating the fraction of Ki-67 positive tumor cells provides a highly reproducible estimate of tumor growth fractions, dictating the necessity for aggressive adjuvant chemotherapy.
	      \begin{itemize}
		      \item \textit{Breast Cancer (Biopsy):} Dichotomized utilizing a clinical threshold of $>14\%$ to indicate high proliferation. The internal cohort from private H3 center comprised 1,481 samples across 3,492 slides (1,171 high, 310 low; stratified into 1,036 training, 148 validation, and 297 testing samples). The external cohort from private H9 center included 703 samples across 703 slides (619 high, 84 low).
		      \item \textit{Breast Cancer (Resection):} Evaluated utilizing the equivalent $>14\%$ threshold. The internal cohort (private H1 center) consisted of 309 samples across 3,534 slides (250 high, 59 low; partitioned into 216 training, 31 validation, and 62 testing samples). The external cohort (private H5 center) comprised 221 samples across 304 slides (109 high, 112 low).
		      \item \textit{Lung Cancer:} Stratified into clinically relevant low, intermediate, and high expression tiers based on established nuclear positivity thresholds. The internal cohort (private H5 center) included 1,128 samples across 1,151 slides (554 high, 473 intermediate, 101 low), partitioned into training (788 samples), validation (113 samples), and testing (227 samples). The external cohort (private H1 center) consisted of 239 samples across 630 slides (43 high, 84 intermediate, 112 low).
		      \item \textit{Colorectal Cancer:} Dichotomized into high versus low proliferative indices based on cohort-specific clinical cutoffs. The internal cohort (private H1 center) comprised 1,491 samples across 3,050 slides (1,269 high, 222 low; stratified into 1,043 training, 149 validation, and 299 testing samples). The external cohort (private H10 center) included 328 samples across 328 slides (291 high, 37 low).
	      \end{itemize}

	\item \textbf{Molecular Subtyping: } Molecular subtypes fundamentally reflect massive, overarching transcriptional programs that govern the macro-level behavior of tumors, enabling precision stratification far beyond single-gene alterations.
	      \begin{itemize}
		      \item \textit{Breast Cancer Subtypes (Biopsy):} Classified into Luminal A, Luminal B, HER2-enriched, and Triple-Negative Breast Cancer (TNBC). The internal cohort (private H3 center) consisted of 1,400 samples across 3,318 slides (972 Luminal B, 172 Luminal A, 131 HER2, 125 TNBC), partitioned into training (980 samples), validation (140 samples), and testing (280 samples). The external cohort (private H9 center) included 703 samples across 703 slides (482 Luminal B, 80 HER2, 77 TNBC, 64 Luminal A).
		      \item \textit{Breast Cancer Subtypes (Resection):} Expanded classification reflecting fine-grained Luminal B subdivisions (Luminal B1/B2). The internal cohort (private H2 center) comprised 2,045 samples across 3,418 slides (614 LumB1, 589 TNBC, 307 LumA, 292 HER2, 243 LumB2), split into 1,431 training, 205 validation, and 409 testing samples. The external cohort (private H8 center) included 788 samples across 791 slides (293 TNBC, 149 HER2, 142 LumA, 122 LumB1, 82 LumB2).
		      \item \textit{Consensus Molecular Subtypes (CMS) in CRC:} CRC is universally categorized into four transcriptomic profiles (CMS1 to CMS4), each distinct in immunological infiltration, stromal remodeling, and metabolic dysregulation. The internal cohort (private H4 center) consisted of 470 samples across 2,137 slides (167 CMS4, 144 CMS2, 104 CMS3, 55 CMS1), stratified into training (326 samples), validation (49 samples), and testing (95 samples). The external cohort (TCGA-CRC, combining COAD and READ) comprised 459 samples across 464 slides (166 CMS2, 126 CMS4, 94 CMS3, 73 CMS1).
	      \end{itemize}
\end{itemize}

\noindent\textbf{Prospective Observational Cohorts.}
To validate the translational viability of STAMP beyond controlled retrospective settings, we deployed the framework on fully independent, prospective observational cohorts. Unlike retrospective datasets, which are often selectively curated, these prospective cohorts comprised consecutive clinical cases collected directly from routine diagnostic workflows. They inherently encompass the natural epidemiological class imbalances, uncurated tissue artifacts, and routine staining variations encountered in daily pathology operations. Crucially, these cohorts were utilized exclusively for the final AI-driven triage simulation and remained strictly blinded during all stages of model pre-training and retrospective threshold calibration.

\begin{itemize}
	\item \textbf{Breast Cancer Prospective Cohort: } Consecutive cases were prospectively enrolled from the private H2 center. This cohort was utilized to evaluate the real-world inference of core diagnostic, proliferative, and subtyping markers.
	      \begin{itemize}
		      \item \textit{ER Status:} Comprised 168 samples across 375 slides, yielding 131 ER-positive and 37 ER-negative cases.
		      \item \textit{PR Status:} Comprised 169 samples across 378 slides, yielding 114 PR-positive and 55 PR-negative cases.
		      \item \textit{HER2 Status:} Comprised 107 samples across 239 slides, yielding 21 HER2-positive and 86 HER2-negative cases.
		      \item \textit{Ki-67 Proliferation Index:} Comprised 167 samples across 373 slides, yielding 136 high-proliferation and 31 low-proliferation cases.
		      \item \textit{Molecular Subtyping:} Comprised 55 completely subtyped samples across 119 slides (18 Luminal A, 11 TNBC, 10 Luminal B2, 8 Luminal B1, 8 HER2).
	      \end{itemize}

	\item \textbf{Lung Cancer Prospective Cohort: } Consecutive cases were prospectively enrolled from the private H1 center, focusing on the definitive lineage markers and proliferation indices essential for NSCLC diagnostic triage.
	      \begin{itemize}
		      \item \textit{Napsin A Status:} Comprised 128 samples across 159 slides, yielding 76 Napsin A-positive and 52 Napsin A-negative cases.
		      \item \textit{TTF-1 Status:} Comprised 138 samples across 174 slides, yielding 91 TTF-1-positive and 47 TTF-1-negative cases.
		      \item \textit{p40 Status:} Comprised 152 samples across 188 slides, yielding 58 p40-positive and 94 p40-negative cases.
		      \item \textit{Ki-67 Proliferation Index:} Comprised 309 samples across 390 slides. Consistent with the three-tier clinical stratification, this yielded 86 high-proliferation, 136 intermediate-proliferation, and 87 low-proliferation cases.
	      \end{itemize}
\end{itemize}

\noindent\textbf{Detailed Cohort Information.} The sample sizes, slide counts, and class distributions for each evaluated biomarker, is comprehensively summarized in Extended Data Table~\ref{dataset-details-breast} and \ref{dataset-details-others}. The sample sizes and slide counts for each center are provided in \exttabref{dataset-accessibility}.

\subsection*{Model Training and Evaluation Details}
\noindent\textbf{PFM Pre-training via Cross-modal Alignment.} The STAMP architecture was pre-trained from scratch (excluding the frozen Virchow2 visual backbone) using the HumanST-1k pre-training corpus. LoRA modules were integrated into the frozen Virchow2 backbone with a bottleneck rank of $r=8$, a scaling factor of $\alpha=16$, and a dropout rate of $0.1$. Training was executed for 30 epochs on a distributed compute node equipped with 8 NVIDIA H800 GPUs. By utilizing Automatic Mixed Precision (AMP) to maximize memory efficiency and computational throughput, we maintained a per-device batch size of 128, yielding a massive effective global batch size of 1,024. The network was optimized utilizing the AdamW algorithm with a base learning rate of $5 \times 10^{-4}$ ($\beta_1 = 0.9, \beta_2 = 0.999, \epsilon = 10^{-8}$) and a weight decay of $0.2$. To stabilize early gradient dynamics, learning rate decay was governed by a cosine annealing scheduler featuring a linear warmup phase of 30,000 iterations. Following the warmup, the learning rate smoothly decayed to a terminal cooldown value of $0.0$ (with a cooldown power of $1.0$). Finally, to properly balance the contrastive alignment and the granular pathway reconstruction within the compound objective function ($\mathcal{L}_{\text{Total}} = \mathcal{L}_{\text{con}} + \lambda \mathcal{L}_{\text{rec}}$), the scaling hyperparameter $\lambda$ was empirically set to $0.1$.

\noindent\textbf{Spatial Gene Expression Prediction via Linear Probing.} To isolate the representational quality of the STAMP framework without confounding non-linear downstream transformations, spatial gene expression prediction was formulated as a linear probing task. To comprehensively capture both baseline tissue architecture and acquired transcriptomic awareness, the morphology-aware embeddings extracted from the frozen original visual encoder were concatenated with the molecularly aligned features from the fine-tuned STAMP encoder. To mitigate the curse of dimensionality and suppress collinearity, this fused high-dimensional feature matrix was first standardized (zero mean, unit variance) and subsequently projected into a compact 256-dimensional subspace utilizing a GPU-accelerated Principal Component Analysis (PCA). An L2-regularized Ridge Regression model (configured without an intercept term) was then deployed to map these PCA-reduced representations directly to the log-transformed ground-truth expression profiles of the targeted 50 HVGs.

\noindent\textbf{Spatial Domain Recognition via Unsupervised Clustering.} To assess the intrinsic capacity of the STAMP framework to capture spatially coherent tissue microenvironments without any supervised guidance, spatial domain recognition was formulated as a fully unsupervised clustering task. For each sample, H\&E patch-level embeddings were loaded alongside their corresponding spatial coordinates and pathologist-annotated ground-truth labels. Spots with missing annotations were excluded to ensure that only unambiguously annotated regions contributed to the evaluation.  The retained embeddings were first standardized to zero mean and unit variance via $Z$-score scaling. To reduce computational dimensionality while preserving the dominant structural variance, the normalized feature matrix was subsequently projected into a compact subspace using GPU-accelerated PCA (up to 50 principal components, with CPU fallback to scikit-learn when GPU resources were unavailable). $k$-nearest-neighbor graph was then constructed over the first 30 principal components with $k=15$ neighbors, and UMAP coordinates were derived for downstream visualization. Unsupervised spatial partitioning was performed by applying $K$-Means clustering to the first 30 principal components. To guarantee a fair and directly comparable evaluation against pathologist-delineated annotations, the number of clusters $k$ was set \textit{a priori} to equal the number of distinct ground-truth annotation categories present in each sample.

\noindent\textbf{Biomarker Prediction via Multiple Instance Learning.} To aggregate patch-level visual features into case-level clinical predictions, we employed the standard Attention-Based Multiple Instance Learning (ABMIL) architecture. During this phase, the pre-trained STAMP feature extractor remained frozen. The extracted patch embeddings for a given WSI were fed into a gated attention network, which learns to assign instance-level importance weights. These weighted features were subsequently pooled into a case-level representation and classified via a Multi-Layer Perceptron (MLP) head. The downstream ABMIL networks were trained for a maximum of 50 epochs utilizing the standard Cross-Entropy loss function. Optimization was performed utilizing the Adam algorithm with an initial learning rate of $2 \times 10^{-4}$ and a weight decay of $1 \times 10^{-5}$ applied exclusively to the trainable parameters. To ensure smooth convergence and stable gradient descent, the learning rate was dynamically modulated via a cosine annealing scheduler, smoothly decaying the rate to an absolute minimum of zero ($\eta_{\text{min}}=0$) over the total training duration. Furthermore, to explicitly prevent overfitting on the training distribution, a strict early stopping mechanism was enforced, terminating the training process if the validation Area Under the Receiver Operating Characteristic (AUC) failed to improve for 10 consecutive epochs.

\subsection*{Statistical Analysis and Evaluation Metrics}
For the \textbf{spatial gene expression prediction} task, the predictive fidelity of the linear probing model was evaluated using the Pearson Correlation Coefficient (PCC). This metric quantifies the linear concordance between the model-inferred expression values and the ground-truth transcriptomic profiles across all spatial spots. For the \textbf{spatial domain recognition} task, the structural agreement between the vision-derived spatial domains and pathologist's annotations was assessed utilizing four quantitative metrics: the Adjusted Rand Index (ARI), Normalized Mutual Information (NMI), Homogeneity, and Completeness. Together, these metrics evaluate the mutual information and cluster purity while penalizing both the over-fragmentation and over-merging of spatially coherent microenvironments. For \textbf{clinical biomarker prediction}, AUC served as the primary evaluation metric for all binary classification tasks (e.g., driver mutations, high/low proliferation indices, and binary protein expressions). For multi-class prediction tasks, such as molecular subtyping and consensus molecular subtypes (CMS), the macro-averaged AUC (Macro-AUC) was utilized. To establish the statistical superiority of the STAMP framework over baseline model (Virchow2), we implemented a non-parametric significance testing protocol. For each predictive task, the performance metrics (AUC or Macro-AUC) were recalculated over 1,000 independent bootstrap resamples of the test cohort. Subsequently, a one-sided Wilcoxon signed-rank test was applied to the bootstrapped metric distributions of STAMP versus baseline model.

\subsection*{Computing Hardware and Software}
To ensure absolute experimental reproducibility across our hierarchical evaluation framework, the computational environments were strictly version-controlled, decoupling the heavy pre-training phase from downstream clinical inference. The large-scale foundational pre-training phase was executed on a high-performance distributed computing node equipped with 8$\times$ NVIDIA H800 Tensor Core GPUs. This environment was built on \texttt{Python} (version 3.13.5) and \texttt{PyTorch} (version 2.6.0, CUDA 12.9). The foundational Virchow2 visual backbone (\url{https://huggingface.co/paige-ai/Virchow2}) was initialized utilizing the \texttt{timm} library (version 1.0.19), while the parameter-efficient Low-Rank Adaptation (LoRA) modules were implemented via the Hugging Face \texttt{PEFT} library (version 0.17.1). Initial spatial transcriptomics parsing and normalization were facilitated by \texttt{Scanpy} (version 1.11.4).

All downstream WSI pre-processing, multiple instance learning (MIL) optimizations, and clinical triage simulations were conducted on standard workstation infrastructure equipped with NVIDIA GeForce RTX 3090 GPUs, running \texttt{Python} (version 3.13.2) and \texttt{PyTorch} (version 2.6.0, CUDA 11.8). Whole-slide image tessellation and automated tissue segmentation were executed utilizing the CLAM preprocessing pipeline (\url{https://github.com/mahmoodlab/clam}), powered by the \texttt{openslide-python} (version 1.4.1) backend. To maintain extreme computational scalability during the spatial gene expression prediction and multi-resolution clustering tasks, high-dimensional dimensionality reduction (PCA) and analytical linear probing (Ridge Regression) were strictly accelerated entirely on the GPU utilizing the NVIDIA RAPIDS \texttt{cuML} library (version 25.8.0). Standardized clinical performance metric calculations were managed by \texttt{TorchMetrics} (version 1.8.0), while non-parametric statistical significance testing (including the Wilcoxon signed-rank tests and bootstrapping) was performed using \texttt{SciPy} (version 1.15.2).

\section*{Data Availability}
The HumanST-1k dataset was curated from publicly accessible data repositories. Spatial transcriptomics matrices and their paired whole-slide images were aggregated from the following platforms: 10$\times$ Genomics Datasets (\url{https://www.10xgenomics.com/datasets}), the China National Center for Bioinformation (CNCB; \url{https://www.cncb.ac.cn}), Dryad Digital Repository (\url{https://datadryad.org}), the European Molecular Biology Laboratory (EMBL; \url{https://www.ebi.ac.uk/}), HEST-1k (\url{https://huggingface.co/datasets/MahmoodLab/hest/}), Heart Cell Atlas (HCA; \url{https://www.heartcellatlas.org/}), the Human Cell Atlas (HCA; \url{https://www.heartcellatlas.org/}), Human Tumor Atlas Network (HTAN; \url{https://humantumoratlas.org/}), Mendeley Data (\url{https://data.mendeley.com/}), the National Center for Biotechnology Information (NCBI; \url{https://www.ncbi.nlm.nih.gov}), Zenodo (\url{https://zenodo.org}), Spatial Research (\url{https://www.spatialresearch.org/}), \textit{etc.}. For downstream tasks, the HEST-Benchmark dataset, utilized for spatial gene expression imputation, is accessible via the Hugging Face repository (\url{https://huggingface.co/datasets/MahmoodLab/hest}). The slides of TCGA and CPTAC cohorts were downloaded from the Genomic Data Commons (GDC) Data Portal (\url{https://portal.gdc.cancer.gov}). The corresponding clinical annotations and genomic alteration labels for both TCGA and CPTAC were robustly queried and extracted via the cBioPortal for Cancer Genomics (\url{https://www.cbioportal.org/}). The slide images and clinical annotations for the eBrain cohort were obtained from the Europe's Digital Infrastructure for Brain Research (\url{https://ebrains.eu}). The multi-centric retrospective and prospective cohorts collected from private clinical centers are not publicly available due to patient privacy regulations and institutional data protection policies.

% The HumanST-1k dataset was curated from publicly accessible data repositories. Spatial transcriptomics matrices and their paired whole-slide images were aggregated from the following platforms: 10$\times$ Genomics Datasets (\url{https://www.10xgenomics.com/datasets}), the China National Center for Bioinformation (CNCB; \url{https://www.cncb.ac.cn}), Dryad Digital Repository (\url{https://datadryad.org}), the European Molecular Biology Laboratory (EMBL; \url{https://www.ebi.ac.uk/}), HEST-1k (\url{https://huggingface.co/datasets/MahmoodLab/hest/}), Heart Cell Atlas (HCA; \url{https://www.heartcellatlas.org/}), the Human Cell Atlas (HCA; \url{https://www.heartcellatlas.org/}), Human Tumor Atlas Network (HTAN; \url{https://humantumoratlas.org/}), Mendeley Data (\url{https://data.mendeley.com/}), the National Center for Biotechnology Information (NCBI; \url{https://www.ncbi.nlm.nih.gov}), Zenodo (\url{https://zenodo.org}), Spatial Research (\url{https://www.spatialresearch.org/}), \textit{etc.}. The details of HumanST-1k dataset are summarized in \textbf{Supplementary Information}. For downstream tasks, the HEST-Benchmark dataset, utilized for spatial gene expression imputation, is accessible via the Hugging Face repository (\url{https://huggingface.co/datasets/MahmoodLab/hest}). The slides of TCGA and CPTAC cohorts were downloaded from the Genomic Data Commons (GDC) Data Portal (\url{https://portal.gdc.cancer.gov}). The corresponding clinical annotations and genomic alteration labels for both TCGA and CPTAC were robustly queried and extracted via the cBioPortal for Cancer Genomics (\url{https://www.cbioportal.org/}). The slide images and clinical annotations for the eBrain cohort were obtained from the Europe's Digital Infrastructure for Brain Research (\url{https://ebrains.eu}). The multi-centric retrospective and prospective cohorts collected from private clinical centers are not publicly available due to patient privacy regulations and institutional data protection policies.

\section*{Code Availability}
% The code and model weights for the STAMP framework, along with scripts for data preprocessing, model training, and evaluation, will be made publicly available upon publication.
% The code and model weights for the STAMP framework, along with scripts for data preprocessing, model training, and evaluation, are available at \url{https://github.com/FT-ZHOU-ZZZ/STAMP}.

All codes and model weights for the STAMP framework will be made publicly available upon publication.

\section*{Author Contribution}
H.C. conceived the study, directed the research, and acquired funding. F.Z. curated the spatial transcriptomics datasets, developed the computational framework, performed model training and evaluation, generated all figures and visualizations, and drafted the manuscript. Y.X., Y.W. (Yihui Wang), Z.G., L.L. (Ling Liang), and J.M. led the data collection, quality control, feature extraction, and experimental design for the breast, colorectal, lung, gastric, and brain cancer cohorts, respectively. Z.Z., provided medical expertise and assisted with experimental design and pathological interpretation. C.J., Z.L. (Ziyi Liu), H.Z., H.W., D.C., C.Z., X.W., and C.Y. assisted with data preprocessing and paper polishing. Y.W. (Yu Wang), W.L., F.G., Z.W., Z.L. (Zhenhui Li), X.Z., and L.L. (Li Liang) contributed to multi-center data acquisition, clinical biomarker validation, and domain-specific pathological interpretation. X.Z., L.L. (Li Liang), and H.C. designed the prospective observational study and coordinated the clinical workflow. All authors reviewed, edited, and approved the final manuscript.

\section*{Acknowledgement}
This work was supported by Research Grants Council of the Hong Kong Special Administrative Region, China (Project R6003-22, C4024-22GF and AoE/E-601/24-N), National Key R\&D Program of China (Project No. 2023YFE0204000), Hong Kong Innovation and Technology Commission (Project No. MHP/002/22 and ITCPD/17-9) and Shenzhen Science and Technology Innovation Committee Fund (Project No. KCXFZ20230731094059008).

\section*{Ethics Declaration}
This study adhered to the Declaration of Helsinki and the International Ethical Guidelines for Biomedical Research Involving Human Subjects, with ethical approval granted by the Human and Artifact Research Ethics Committee of The Hong Kong University of Science and Technology (HREP-2024-0423). The prospective study protocol was approved by the Medical Ethics Committee of NanFang Hospital, Southern Medical University (NFEC-2025-403). The corresponding protocol was registered on ClinicalTrials.gov (NCT07157618).

\bibliography{sample.bib}

\begin{thebibliography}{10}
\urlstyle{rm}
\expandafter\ifx\csname url\endcsname\relax
  \def\url#1{\texttt{#1}}\fi
\expandafter\ifx\csname urlprefix\endcsname\relax\def\urlprefix{URL }\fi
\expandafter\ifx\csname doiprefix\endcsname\relax\def\doiprefix{DOI: }\fi
\providecommand{\bibinfo}[2]{#2}
\providecommand{\eprint}[2][]{\url{#2}}

\bibitem{mateo2022delivering}
\bibinfo{author}{Mateo, J.} \emph{et~al.}
\newblock \bibinfo{journal}{\bibinfo{title}{Delivering precision oncology to
  patients with cancer}}.
\newblock {\emph{\JournalTitle{Nature medicine}}}
  \textbf{\bibinfo{volume}{28}}, \bibinfo{pages}{658--665}
  (\bibinfo{year}{2022}).

\bibitem{akhoundova2022clinical}
\bibinfo{author}{Akhoundova, D.} \& \bibinfo{author}{Rubin, M.~A.}
\newblock \bibinfo{journal}{\bibinfo{title}{Clinical application of advanced
  multi-omics tumor profiling: Shaping precision oncology of the future}}.
\newblock {\emph{\JournalTitle{Cancer cell}}} \textbf{\bibinfo{volume}{40}},
  \bibinfo{pages}{920--938} (\bibinfo{year}{2022}).

\bibitem{brlek2025advances}
\bibinfo{author}{Brlek, P.} \emph{et~al.}
\newblock \bibinfo{journal}{\bibinfo{title}{Advances in precision oncology:
  From molecular profiling to regulatory-approved targeted therapies}}.
\newblock {\emph{\JournalTitle{Cancers}}} \textbf{\bibinfo{volume}{17}},
  \bibinfo{pages}{3500} (\bibinfo{year}{2025}).

\bibitem{bera2019artificial}
\bibinfo{author}{Bera, K.}, \bibinfo{author}{Schalper, K.~A.},
  \bibinfo{author}{Rimm, D.~L.}, \bibinfo{author}{Velcheti, V.} \&
  \bibinfo{author}{Madabhushi, A.}
\newblock \bibinfo{journal}{\bibinfo{title}{Artificial intelligence in digital
  pathology—new tools for diagnosis and precision oncology}}.
\newblock {\emph{\JournalTitle{Nature reviews Clinical oncology}}}
  \textbf{\bibinfo{volume}{16}}, \bibinfo{pages}{703--715}
  (\bibinfo{year}{2019}).

\bibitem{niazi2019digital}
\bibinfo{author}{Niazi, M. K.~K.}, \bibinfo{author}{Parwani, A.~V.} \&
  \bibinfo{author}{Gurcan, M.~N.}
\newblock \bibinfo{journal}{\bibinfo{title}{Digital pathology and artificial
  intelligence}}.
\newblock {\emph{\JournalTitle{The lancet oncology}}}
  \textbf{\bibinfo{volume}{20}}, \bibinfo{pages}{e253--e261}
  (\bibinfo{year}{2019}).

\bibitem{kather2019deep}
\bibinfo{author}{Kather, J.~N.} \emph{et~al.}
\newblock \bibinfo{journal}{\bibinfo{title}{Deep learning can predict
  microsatellite instability directly from histology in gastrointestinal
  cancer}}.
\newblock {\emph{\JournalTitle{Nature medicine}}}
  \textbf{\bibinfo{volume}{25}}, \bibinfo{pages}{1054--1056}
  (\bibinfo{year}{2019}).

\bibitem{kather2020pan}
\bibinfo{author}{Kather, J.~N.} \emph{et~al.}
\newblock \bibinfo{journal}{\bibinfo{title}{Pan-cancer image-based detection of
  clinically actionable genetic alterations}}.
\newblock {\emph{\JournalTitle{Nature cancer}}} \textbf{\bibinfo{volume}{1}},
  \bibinfo{pages}{789--799} (\bibinfo{year}{2020}).

\bibitem{schmauch2020deep}
\bibinfo{author}{Schmauch, B.} \emph{et~al.}
\newblock \bibinfo{journal}{\bibinfo{title}{A deep learning model to predict
  rna-seq expression of tumours from whole slide images}}.
\newblock {\emph{\JournalTitle{Nature communications}}}
  \textbf{\bibinfo{volume}{11}}, \bibinfo{pages}{3877} (\bibinfo{year}{2020}).

\bibitem{saltz2018spatial}
\bibinfo{author}{Saltz, J.} \emph{et~al.}
\newblock \bibinfo{journal}{\bibinfo{title}{Spatial organization and molecular
  correlation of tumor-infiltrating lymphocytes using deep learning on
  pathology images}}.
\newblock {\emph{\JournalTitle{Cell reports}}} \textbf{\bibinfo{volume}{23}},
  \bibinfo{pages}{181--193} (\bibinfo{year}{2018}).

\bibitem{barkley2022cancer}
\bibinfo{author}{Barkley, D.} \emph{et~al.}
\newblock \bibinfo{journal}{\bibinfo{title}{Cancer cell states recur across
  tumor types and form specific interactions with the tumor microenvironment}}.
\newblock {\emph{\JournalTitle{Nature genetics}}}
  \textbf{\bibinfo{volume}{54}}, \bibinfo{pages}{1192--1201}
  (\bibinfo{year}{2022}).

\bibitem{jerby2018cancer}
\bibinfo{author}{Jerby-Arnon, L.} \emph{et~al.}
\newblock \bibinfo{journal}{\bibinfo{title}{A cancer cell program promotes t
  cell exclusion and resistance to checkpoint blockade}}.
\newblock {\emph{\JournalTitle{Cell}}} \textbf{\bibinfo{volume}{175}},
  \bibinfo{pages}{984--997} (\bibinfo{year}{2018}).

\bibitem{chen2024towards}
\bibinfo{author}{Chen, R.~J.} \emph{et~al.}
\newblock \bibinfo{journal}{\bibinfo{title}{Towards a general-purpose
  foundation model for computational pathology}}.
\newblock {\emph{\JournalTitle{Nature medicine}}}
  \textbf{\bibinfo{volume}{30}}, \bibinfo{pages}{850--862}
  (\bibinfo{year}{2024}).

\bibitem{xu2024whole}
\bibinfo{author}{Xu, H.} \emph{et~al.}
\newblock \bibinfo{journal}{\bibinfo{title}{A whole-slide foundation model for
  digital pathology from real-world data}}.
\newblock {\emph{\JournalTitle{Nature}}} \textbf{\bibinfo{volume}{630}},
  \bibinfo{pages}{181--188} (\bibinfo{year}{2024}).

\bibitem{vorontsov2024foundation}
\bibinfo{author}{Vorontsov, E.} \emph{et~al.}
\newblock \bibinfo{journal}{\bibinfo{title}{A foundation model for
  clinical-grade computational pathology and rare cancers detection}}.
\newblock {\emph{\JournalTitle{Nature medicine}}}
  \textbf{\bibinfo{volume}{30}}, \bibinfo{pages}{2924--2935}
  (\bibinfo{year}{2024}).

\bibitem{wang2024pathology}
\bibinfo{author}{Wang, X.} \emph{et~al.}
\newblock \bibinfo{journal}{\bibinfo{title}{A pathology foundation model for
  cancer diagnosis and prognosis prediction}}.
\newblock {\emph{\JournalTitle{Nature}}} \textbf{\bibinfo{volume}{634}},
  \bibinfo{pages}{970--978} (\bibinfo{year}{2024}).

\bibitem{ma2025generalizable}
\bibinfo{author}{Ma, J.} \emph{et~al.}
\newblock \bibinfo{journal}{\bibinfo{title}{A generalizable pathology
  foundation model using a unified knowledge distillation pretraining
  framework}}.
\newblock {\emph{\JournalTitle{Nature Biomedical Engineering}}}
  \bibinfo{pages}{1--20} (\bibinfo{year}{2025}).

\bibitem{oquab2023dinov2}
\bibinfo{author}{Oquab, M.} \emph{et~al.}
\newblock \bibinfo{journal}{\bibinfo{title}{Dinov2: Learning robust visual
  features without supervision}}.
\newblock {\emph{\JournalTitle{arXiv preprint arXiv:2304.07193}}}
  (\bibinfo{year}{2023}).

\bibitem{radford2021learning}
\bibinfo{author}{Radford, A.} \emph{et~al.}
\newblock \bibinfo{title}{Learning transferable visual models from natural
  language supervision}.
\newblock In \emph{\bibinfo{booktitle}{International conference on machine
  learning}}, \bibinfo{pages}{8748--8763} (\bibinfo{organization}{PmLR},
  \bibinfo{year}{2021}).

\bibitem{li2022blip}
\bibinfo{author}{Li, J.}, \bibinfo{author}{Li, D.}, \bibinfo{author}{Xiong, C.}
  \& \bibinfo{author}{Hoi, S.}
\newblock \bibinfo{title}{Blip: Bootstrapping language-image pre-training for
  unified vision-language understanding and generation}.
\newblock In \emph{\bibinfo{booktitle}{International conference on machine
  learning}}, \bibinfo{pages}{12888--12900} (\bibinfo{organization}{PMLR},
  \bibinfo{year}{2022}).

\bibitem{yu2022coca}
\bibinfo{author}{Yu, J.} \emph{et~al.}
\newblock \bibinfo{journal}{\bibinfo{title}{Coca: Contrastive captioners are
  image-text foundation models}}.
\newblock {\emph{\JournalTitle{arXiv preprint arXiv:2205.01917}}}
  (\bibinfo{year}{2022}).

\bibitem{lu2024visual}
\bibinfo{author}{Lu, M.~Y.} \emph{et~al.}
\newblock \bibinfo{journal}{\bibinfo{title}{A visual-language foundation model
  for computational pathology}}.
\newblock {\emph{\JournalTitle{Nature medicine}}}
  \textbf{\bibinfo{volume}{30}}, \bibinfo{pages}{863--874}
  (\bibinfo{year}{2024}).

\bibitem{huang2023visual}
\bibinfo{author}{Huang, Z.}, \bibinfo{author}{Bianchi, F.},
  \bibinfo{author}{Yuksekgonul, M.}, \bibinfo{author}{Montine, T.~J.} \&
  \bibinfo{author}{Zou, J.}
\newblock \bibinfo{journal}{\bibinfo{title}{A visual--language foundation model
  for pathology image analysis using medical twitter}}.
\newblock {\emph{\JournalTitle{Nature medicine}}}
  \textbf{\bibinfo{volume}{29}}, \bibinfo{pages}{2307--2316}
  (\bibinfo{year}{2023}).

\bibitem{ikezogwo2023quilt}
\bibinfo{author}{Ikezogwo, W.} \emph{et~al.}
\newblock \bibinfo{journal}{\bibinfo{title}{Quilt-1m: One million image-text
  pairs for histopathology}}.
\newblock {\emph{\JournalTitle{Advances in neural information processing
  systems}}} \textbf{\bibinfo{volume}{36}}, \bibinfo{pages}{37995--38017}
  (\bibinfo{year}{2023}).

\bibitem{xu2025multimodal}
\bibinfo{author}{Xu, Y.} \emph{et~al.}
\newblock \bibinfo{journal}{\bibinfo{title}{A multimodal knowledge-enhanced
  whole-slide pathology foundation model}}.
\newblock {\emph{\JournalTitle{Nature Communications}}}
  (\bibinfo{year}{2025}).

\bibitem{marx2021method}
\bibinfo{author}{Marx, V.}
\newblock \bibinfo{journal}{\bibinfo{title}{Method of the year: spatially
  resolved transcriptomics}}.
\newblock {\emph{\JournalTitle{Nature methods}}} \textbf{\bibinfo{volume}{18}},
  \bibinfo{pages}{9--14} (\bibinfo{year}{2021}).

\bibitem{rao2021exploring}
\bibinfo{author}{Rao, A.}, \bibinfo{author}{Barkley, D.},
  \bibinfo{author}{Fran{\c{c}}a, G.~S.} \& \bibinfo{author}{Yanai, I.}
\newblock \bibinfo{journal}{\bibinfo{title}{Exploring tissue architecture using
  spatial transcriptomics}}.
\newblock {\emph{\JournalTitle{Nature}}} \textbf{\bibinfo{volume}{596}},
  \bibinfo{pages}{211--220} (\bibinfo{year}{2021}).

\bibitem{subramanian2005gene}
\bibinfo{author}{Subramanian, A.} \emph{et~al.}
\newblock \bibinfo{journal}{\bibinfo{title}{Gene set enrichment analysis: a
  knowledge-based approach for interpreting genome-wide expression profiles}}.
\newblock {\emph{\JournalTitle{Proceedings of the national academy of
  sciences}}} \textbf{\bibinfo{volume}{102}}, \bibinfo{pages}{15545--15550}
  (\bibinfo{year}{2005}).

\bibitem{hu2022lora}
\bibinfo{author}{Hu, E.~J.} \emph{et~al.}
\newblock \bibinfo{journal}{\bibinfo{title}{Lora: Low-rank adaptation of large
  language models.}}
\newblock {\emph{\JournalTitle{Iclr}}} \textbf{\bibinfo{volume}{1}},
  \bibinfo{pages}{3} (\bibinfo{year}{2022}).

\bibitem{han2024parameter}
\bibinfo{author}{Han, Z.}, \bibinfo{author}{Gao, C.}, \bibinfo{author}{Liu,
  J.}, \bibinfo{author}{Zhang, J.} \& \bibinfo{author}{Zhang, S.~Q.}
\newblock \bibinfo{journal}{\bibinfo{title}{Parameter-efficient fine-tuning for
  large models: A comprehensive survey}}.
\newblock {\emph{\JournalTitle{arXiv preprint arXiv:2403.14608}}}
  (\bibinfo{year}{2024}).

\bibitem{zimmermann2024virchow2}
\bibinfo{author}{Zimmermann, E.} \emph{et~al.}
\newblock \bibinfo{journal}{\bibinfo{title}{Virchow2: Scaling self-supervised
  mixed magnification models in pathology}}.
\newblock {\emph{\JournalTitle{arXiv preprint arXiv:2408.00738}}}
  (\bibinfo{year}{2024}).

\bibitem{ma2025pathbench}
\bibinfo{author}{Ma, J.} \emph{et~al.}
\newblock \bibinfo{journal}{\bibinfo{title}{Pathbench: A comprehensive
  comparison benchmark for pathology foundation models towards precision
  oncology}}.
\newblock {\emph{\JournalTitle{arXiv preprint arXiv:2505.20202}}}
  (\bibinfo{year}{2025}).

\bibitem{liberzon2015molecular}
\bibinfo{author}{Liberzon, A.} \emph{et~al.}
\newblock \bibinfo{journal}{\bibinfo{title}{The molecular signatures database
  hallmark gene set collection}}.
\newblock {\emph{\JournalTitle{Cell systems}}} \textbf{\bibinfo{volume}{1}},
  \bibinfo{pages}{417--425} (\bibinfo{year}{2015}).

\bibitem{chen2025visual}
\bibinfo{author}{Chen, W.} \emph{et~al.}
\newblock \bibinfo{journal}{\bibinfo{title}{A visual--omics foundation model to
  bridge histopathology with spatial transcriptomics}}.
\newblock {\emph{\JournalTitle{Nature Methods}}} \textbf{\bibinfo{volume}{22}},
  \bibinfo{pages}{1568--1582} (\bibinfo{year}{2025}).

\bibitem{junttila2013influence}
\bibinfo{author}{Junttila, M.~R.} \& \bibinfo{author}{De~Sauvage, F.~J.}
\newblock \bibinfo{journal}{\bibinfo{title}{Influence of tumour
  micro-environment heterogeneity on therapeutic response}}.
\newblock {\emph{\JournalTitle{Nature}}} \textbf{\bibinfo{volume}{501}},
  \bibinfo{pages}{346--354} (\bibinfo{year}{2013}).

\bibitem{binnewies2018understanding}
\bibinfo{author}{Binnewies, M.} \emph{et~al.}
\newblock \bibinfo{journal}{\bibinfo{title}{Understanding the tumor immune
  microenvironment (time) for effective therapy}}.
\newblock {\emph{\JournalTitle{Nature medicine}}}
  \textbf{\bibinfo{volume}{24}}, \bibinfo{pages}{541--550}
  (\bibinfo{year}{2018}).

\bibitem{roma2019targeting}
\bibinfo{author}{Roma-Rodrigues, C.}, \bibinfo{author}{Mendes, R.},
  \bibinfo{author}{Baptista, P.~V.} \& \bibinfo{author}{Fernandes, A.~R.}
\newblock \bibinfo{journal}{\bibinfo{title}{Targeting tumor microenvironment
  for cancer therapy}}.
\newblock {\emph{\JournalTitle{International journal of molecular sciences}}}
  \textbf{\bibinfo{volume}{20}}, \bibinfo{pages}{840} (\bibinfo{year}{2019}).

\bibitem{jaume2024hest}
\bibinfo{author}{Jaume, G.} \emph{et~al.}
\newblock \bibinfo{journal}{\bibinfo{title}{Hest-1k: A dataset for spatial
  transcriptomics and histology image analysis}}.
\newblock {\emph{\JournalTitle{Advances in Neural Information Processing
  Systems}}} \textbf{\bibinfo{volume}{37}}, \bibinfo{pages}{53798--53833}
  (\bibinfo{year}{2024}).

\bibitem{keren2018structured}
\bibinfo{author}{Keren, L.} \emph{et~al.}
\newblock \bibinfo{journal}{\bibinfo{title}{A structured tumor-immune
  microenvironment in triple negative breast cancer revealed by multiplexed ion
  beam imaging}}.
\newblock {\emph{\JournalTitle{Cell}}} \textbf{\bibinfo{volume}{174}},
  \bibinfo{pages}{1373--1387} (\bibinfo{year}{2018}).

\bibitem{andersson2020spatial}
\bibinfo{author}{Andersson, A.} \emph{et~al.}
\newblock \bibinfo{journal}{\bibinfo{title}{Spatial deconvolution of
  her2-positive breast tumors reveals novel intercellular relationships}}.
\newblock {\emph{\JournalTitle{bioRxiv}}} \bibinfo{pages}{2020--07}
  (\bibinfo{year}{2020}).

\bibitem{dawo_10x_2025}
\bibinfo{author}{Dawo, S.}, \bibinfo{author}{Nonchev, K.} \&
  \bibinfo{author}{Silina, K.}
\newblock \bibinfo{title}{{10x Visium Spatial Transcriptomics Dataset: Kidney
  (3) and Lung (5) Cancer with Tertiary Lymphoid Structures}},
  \doiprefix\url{10.5281/zenodo.14620362} (\bibinfo{year}{2025}).

\bibitem{erickson2022spatially}
\bibinfo{author}{Erickson, A.} \emph{et~al.}
\newblock \bibinfo{journal}{\bibinfo{title}{Spatially resolved clonal copy
  number alterations in benign and malignant tissue}}.
\newblock {\emph{\JournalTitle{Nature}}} \textbf{\bibinfo{volume}{608}},
  \bibinfo{pages}{360--367} (\bibinfo{year}{2022}).

\bibitem{couture2018image}
\bibinfo{author}{Couture, H.~D.} \emph{et~al.}
\newblock \bibinfo{journal}{\bibinfo{title}{Image analysis with deep learning
  to predict breast cancer grade, er status, histologic subtype, and intrinsic
  subtype}}.
\newblock {\emph{\JournalTitle{NPJ breast cancer}}}
  \textbf{\bibinfo{volume}{4}}, \bibinfo{pages}{30} (\bibinfo{year}{2018}).

\bibitem{valieris2024weakly}
\bibinfo{author}{Valieris, R.} \emph{et~al.}
\newblock \bibinfo{journal}{\bibinfo{title}{Weakly-supervised deep learning
  models enable her2-low prediction from h \&e stained slides}}.
\newblock {\emph{\JournalTitle{Breast Cancer Research}}}
  \textbf{\bibinfo{volume}{26}}, \bibinfo{pages}{124} (\bibinfo{year}{2024}).

\bibitem{allison2020estrogen}
\bibinfo{author}{Allison, K.~H.} \emph{et~al.}
\newblock \bibinfo{journal}{\bibinfo{title}{Estrogen and progesterone receptor
  testing in breast cancer: Asco/cap guideline update}}.
\newblock {\emph{\JournalTitle{Journal of Clinical Oncology}}}
  \textbf{\bibinfo{volume}{38}}, \bibinfo{pages}{1346--1366}
  (\bibinfo{year}{2020}).

\bibitem{wolff2018human}
\bibinfo{author}{Wolff, A.~C.} \emph{et~al.}
\newblock \bibinfo{journal}{\bibinfo{title}{Human epidermal growth factor
  receptor 2 testing in breast cancer: American society of clinical
  oncology/college of american pathologists clinical practice guideline focused
  update}}.
\newblock {\emph{\JournalTitle{Archives of pathology \& laboratory medicine}}}
  \textbf{\bibinfo{volume}{142}}, \bibinfo{pages}{1364--1382}
  (\bibinfo{year}{2018}).

\bibitem{travis20152015}
\bibinfo{author}{Travis, W.~D.} \emph{et~al.}
\newblock \bibinfo{journal}{\bibinfo{title}{The 2015 world health organization
  classification of lung tumors: impact of genetic, clinical and radiologic
  advances since the 2004 classification}}.
\newblock {\emph{\JournalTitle{Journal of thoracic oncology}}}
  \textbf{\bibinfo{volume}{10}}, \bibinfo{pages}{1243--1260}
  (\bibinfo{year}{2015}).

\bibitem{yatabe2019best}
\bibinfo{author}{Yatabe, Y.} \emph{et~al.}
\newblock \bibinfo{journal}{\bibinfo{title}{Best practices recommendations for
  diagnostic immunohistochemistry in lung cancer}}.
\newblock {\emph{\JournalTitle{Journal of thoracic oncology}}}
  \textbf{\bibinfo{volume}{14}}, \bibinfo{pages}{377--407}
  (\bibinfo{year}{2019}).

\bibitem{bishop2012p40}
\bibinfo{author}{Bishop, J.~A.} \emph{et~al.}
\newblock \bibinfo{journal}{\bibinfo{title}{p40 ($\delta$np63) is superior to
  p63 for the diagnosis of pulmonary squamous cell carcinoma}}.
\newblock {\emph{\JournalTitle{Modern pathology}}}
  \textbf{\bibinfo{volume}{25}}, \bibinfo{pages}{405--415}
  (\bibinfo{year}{2012}).

\bibitem{mosele2020recommendations}
\bibinfo{author}{Mosele, F.} \emph{et~al.}
\newblock \bibinfo{journal}{\bibinfo{title}{Recommendations for the use of
  next-generation sequencing (ngs) for patients with metastatic cancers: a
  report from the esmo precision medicine working group}}.
\newblock {\emph{\JournalTitle{Annals of Oncology}}}
  \textbf{\bibinfo{volume}{31}}, \bibinfo{pages}{1491--1505}
  (\bibinfo{year}{2020}).

\bibitem{coudray2018classification}
\bibinfo{author}{Coudray, N.} \emph{et~al.}
\newblock \bibinfo{journal}{\bibinfo{title}{Classification and mutation
  prediction from non--small cell lung cancer histopathology images using deep
  learning}}.
\newblock {\emph{\JournalTitle{Nature medicine}}}
  \textbf{\bibinfo{volume}{24}}, \bibinfo{pages}{1559--1567}
  (\bibinfo{year}{2018}).

\bibitem{le2017mismatch}
\bibinfo{author}{Le, D.~T.} \emph{et~al.}
\newblock \bibinfo{journal}{\bibinfo{title}{Mismatch repair deficiency predicts
  response of solid tumors to pd-1 blockade}}.
\newblock {\emph{\JournalTitle{Science}}} \textbf{\bibinfo{volume}{357}},
  \bibinfo{pages}{409--413} (\bibinfo{year}{2017}).

\bibitem{marabelle2020association}
\bibinfo{author}{Marabelle, A.} \emph{et~al.}
\newblock \bibinfo{journal}{\bibinfo{title}{Association of tumour mutational
  burden with outcomes in patients with advanced solid tumours treated with
  pembrolizumab: prospective biomarker analysis of the multicohort, open-label,
  phase 2 keynote-158 study}}.
\newblock {\emph{\JournalTitle{The Lancet Oncology}}}
  \textbf{\bibinfo{volume}{21}}, \bibinfo{pages}{1353--1365}
  (\bibinfo{year}{2020}).

\bibitem{chalmers2017analysis}
\bibinfo{author}{Chalmers, Z.~R.} \emph{et~al.}
\newblock \bibinfo{journal}{\bibinfo{title}{Analysis of 100,000 human cancer
  genomes reveals the landscape of tumor mutational burden}}.
\newblock {\emph{\JournalTitle{Genome medicine}}} \textbf{\bibinfo{volume}{9}},
  \bibinfo{pages}{34} (\bibinfo{year}{2017}).

\bibitem{mcgrail2021high}
\bibinfo{author}{McGrail, D.} \emph{et~al.}
\newblock \bibinfo{journal}{\bibinfo{title}{High tumor mutation burden fails to
  predict immune checkpoint blockade response across all cancer types}}.
\newblock {\emph{\JournalTitle{Annals of Oncology}}}
  \textbf{\bibinfo{volume}{32}}, \bibinfo{pages}{661--672}
  (\bibinfo{year}{2021}).

\bibitem{cancer2012comprehensive}
\bibinfo{author}{13, B. . W. H. . H. M. S. C. L. . . P. P. J. . K.~R.},
  \bibinfo{author}{data analysis: Baylor College~of Medicine Creighton Chad J.
  22 23 Donehower Lawrence A. 22 23 24~25, G.}, \bibinfo{author}{for Systems
  Biology Reynolds Sheila 31 Kreisberg Richard B. 31 Bernard Brady 31 Bressler
  Ryan 31 Erkkila Timo 32 Lin Jake 31 Thorsson Vesteinn 31 Zhang Wei 33
  Shmulevich Ilya~31, I.} \emph{et~al.}
\newblock \bibinfo{journal}{\bibinfo{title}{Comprehensive molecular portraits
  of human breast tumours}}.
\newblock {\emph{\JournalTitle{Nature}}} \textbf{\bibinfo{volume}{490}},
  \bibinfo{pages}{61--70} (\bibinfo{year}{2012}).

\bibitem{parker2009supervised}
\bibinfo{author}{Parker, J.~S.} \emph{et~al.}
\newblock \bibinfo{journal}{\bibinfo{title}{Supervised risk predictor of breast
  cancer based on intrinsic subtypes}}.
\newblock {\emph{\JournalTitle{Journal of clinical oncology}}}
  \textbf{\bibinfo{volume}{27}}, \bibinfo{pages}{1160--1167}
  (\bibinfo{year}{2009}).

\bibitem{guinney2015consensus}
\bibinfo{author}{Guinney, J.} \emph{et~al.}
\newblock \bibinfo{journal}{\bibinfo{title}{The consensus molecular subtypes of
  colorectal cancer}}.
\newblock {\emph{\JournalTitle{Nature medicine}}}
  \textbf{\bibinfo{volume}{21}}, \bibinfo{pages}{1350--1356}
  (\bibinfo{year}{2015}).

\bibitem{sanchez2018oncogenic}
\bibinfo{author}{Sanchez-Vega, F.} \emph{et~al.}
\newblock \bibinfo{journal}{\bibinfo{title}{Oncogenic signaling pathways in the
  cancer genome atlas}}.
\newblock {\emph{\JournalTitle{Cell}}} \textbf{\bibinfo{volume}{173}},
  \bibinfo{pages}{321--337} (\bibinfo{year}{2018}).

\bibitem{yarchoan2017tumor}
\bibinfo{author}{Yarchoan, M.}, \bibinfo{author}{Hopkins, A.} \&
  \bibinfo{author}{Jaffee, E.~M.}
\newblock \bibinfo{journal}{\bibinfo{title}{Tumor mutational burden and
  response rate to pd-1 inhibition}}.
\newblock {\emph{\JournalTitle{New England Journal of Medicine}}}
  \textbf{\bibinfo{volume}{377}}, \bibinfo{pages}{2500--2501}
  (\bibinfo{year}{2017}).

\bibitem{litchfield2021meta}
\bibinfo{author}{Litchfield, K.} \emph{et~al.}
\newblock \bibinfo{journal}{\bibinfo{title}{Meta-analysis of tumor-and t
  cell-intrinsic mechanisms of sensitization to checkpoint inhibition}}.
\newblock {\emph{\JournalTitle{Cell}}} \textbf{\bibinfo{volume}{184}},
  \bibinfo{pages}{596--614} (\bibinfo{year}{2021}).

\bibitem{kleppe2021designing}
\bibinfo{author}{Kleppe, A.} \emph{et~al.}
\newblock \bibinfo{journal}{\bibinfo{title}{Designing deep learning studies in
  cancer diagnostics}}.
\newblock {\emph{\JournalTitle{Nature Reviews Cancer}}}
  \textbf{\bibinfo{volume}{21}}, \bibinfo{pages}{199--211}
  (\bibinfo{year}{2021}).

\bibitem{vandereyken2023methods}
\bibinfo{author}{Vandereyken, K.}, \bibinfo{author}{Sifrim, A.},
  \bibinfo{author}{Thienpont, B.} \& \bibinfo{author}{Voet, T.}
\newblock \bibinfo{journal}{\bibinfo{title}{Methods and applications for
  single-cell and spatial multi-omics}}.
\newblock {\emph{\JournalTitle{Nature Reviews Genetics}}}
  \textbf{\bibinfo{volume}{24}}, \bibinfo{pages}{494--515}
  (\bibinfo{year}{2023}).

\bibitem{klambauer2017self}
\bibinfo{author}{Klambauer, G.}, \bibinfo{author}{Unterthiner, T.},
  \bibinfo{author}{Mayr, A.} \& \bibinfo{author}{Hochreiter, S.}
\newblock \bibinfo{journal}{\bibinfo{title}{Self-normalizing neural networks}}.
\newblock {\emph{\JournalTitle{Advances in neural information processing
  systems}}} \textbf{\bibinfo{volume}{30}} (\bibinfo{year}{2017}).

\bibitem{vaswani2017attention}
\bibinfo{author}{Vaswani, A.} \emph{et~al.}
\newblock \bibinfo{journal}{\bibinfo{title}{Attention is all you need}}.
\newblock {\emph{\JournalTitle{Advances in neural information processing
  systems}}} \textbf{\bibinfo{volume}{30}} (\bibinfo{year}{2017}).

\bibitem{oord2018representation}
\bibinfo{author}{Oord, A. v.~d.}, \bibinfo{author}{Li, Y.} \&
  \bibinfo{author}{Vinyals, O.}
\newblock \bibinfo{journal}{\bibinfo{title}{Representation learning with
  contrastive predictive coding}}.
\newblock {\emph{\JournalTitle{arXiv preprint arXiv:1807.03748}}}
  (\bibinfo{year}{2018}).

\end{thebibliography}

\clearpage

\setcounter{figure}{0}
\setcounter{table}{0}

\renewcommand{\figurename}{Extended Data Figure}
\renewcommand{\tablename}{Extended Data Table}

\section*{Extended Data}

\begin{table*}[!htbp]
	\caption{\textbf{Dataset details for spatial gene expression prediction.} There are a total of 8 datasets across 8 distinct cancer types. The table provides the overview of the oncology cohorts, their corresponding platforms, sample sizes, and total spot counts utilized for benchmarking the gene expression prediction of the STAMP framework.}
	\begin{center}
		\begin{tabular}{lllccccc}
			\toprule
			\textbf{Oncology} & \textbf{Cancer Type}             & \textbf{Platform} & \textbf{Samples} & \textbf{Total Spots} \\
			\midrule
			CCRCC             & Clear Cell Renal Cell Carcinoma  & 10x Visium        & 24               & 74,220               \\
			COAD              & Colon Adenocarcinoma             & 10x Xenium        & 4                & 18,523               \\
			IDC               & Invasive Ductal Carcinoma        & 10x Xenium        & 4                & 44,974               \\
			NSCLC             & Non-Small Cell Lung Cancer       & 10x Xenium        & 2                & 7,505                \\
			PAAD              & Pancreatic Ductal Adenocarcinoma & 10x Xenium        & 3                & 9,793                \\
			PRAD              & Prostate Adenocarcinoma          & 10x Visium        & 23               & 62,710               \\
			READ              & Rectal Adenocarcinoma            & 10x Visium        & 4                & 8,407                \\
			SKCM              & Skin Cutaneous Melanoma          & 10x Xenium        & 2                & 5,716                \\
			\bottomrule
		\end{tabular}
	\end{center}
	\label{hest-bench-dataset}
\end{table*}

\begin{table*}[!htbp]
	\caption{\textbf{Comprehensive benchmarking of spatial gene expression imputation across eight pan-cancer cohorts.} The gene expression prediction is quantified by the mean Pearson Correlation Coefficient (PCC), across the top 50 highly variable genes (HVGs) within eight distinct oncology datasets from the HEST-Benchmark. The ``Gain'' row denotes the absolute performance improvement of STAMP relative to its baseline model (Virchow2). The \textbf{best-performing} method for each cohort is highlighted in bold, while the \underline{second-best} is underlined.}
	\begin{center}
\begin{tabular}{l|cccccccc|c}
			\toprule
			Model    & CCRCC                       & COAD   & IDC    & LUNG   & PAAD   & PRAD   & READ   & SKCM   & Overall \\
			\midrule
			OmiCLIP  & \makecell{0.124                                                                                      \\$\pm$0.059}  &\makecell{0.143\\$\pm$0.107}  &\makecell{0.375\\$\pm$0.040}  &\makecell{0.451\\$\pm$0.012}  &\makecell{0.324\\$\pm$0.048}  &\makecell{0.195\\$\pm$0.006}  &\makecell{0.059\\$\pm$0.057}  &\makecell{0.438\\$\pm$0.059}  &0.264  \\[15pt] % 在这里调整间距，例如 4pt, 6pt 或 1ex
			PLIP     & \makecell{0.192                                                                                      \\$\pm$0.038}  &\makecell{0.176\\$\pm$0.068}  &\makecell{0.464\\$\pm$0.038}  &\makecell{0.479\\$\pm$0.009}  &\makecell{0.405\\$\pm$0.070}  &\makecell{0.273\\$\pm$0.020}  &\makecell{0.087\\$\pm$0.080}  &\makecell{0.486\\$\pm$0.072}  &0.320  \\[15pt]
			CHIEF    & \makecell{0.236                                                                                      \\$\pm$0.044}  &\makecell{0.239\\$\pm$0.046}  &\makecell{0.499\\$\pm$0.042}  &\makecell{0.512\\$\pm$0.032}  &\makecell{0.432\\$\pm$0.072}  &\makecell{0.355\\$\pm$0.045}  &\makecell{0.096\\$\pm$0.081}  &\makecell{0.513\\$\pm$0.087}  &0.360  \\[15pt]
			MUSK     & \makecell{0.219                                                                                      \\$\pm$0.032}  &\makecell{0.249\\$\pm$0.059}  &\makecell{0.534\\$\pm$0.071}  &\makecell{0.510\\$\pm$0.030}  &\makecell{0.435\\$\pm$0.079}  &\makecell{0.331\\$\pm$0.016}  &\makecell{0.102\\$\pm$0.090}  &\makecell{0.530\\$\pm$0.069}  &0.364  \\[15pt]
			CONCH    & \makecell{0.219                                                                                      \\$\pm$0.036}  &\makecell{0.252\\$\pm$0.007}  &\makecell{0.536\\$\pm$0.084}  &\makecell{0.531\\$\pm$0.011}  &\makecell{0.446\\$\pm$0.072}  &\makecell{0.355\\$\pm$0.010}  &\makecell{0.162\\$\pm$0.050}  &\makecell{0.581\\$\pm$0.052}  &0.385  \\
\bottomrule
		\end{tabular}
	\end{center}
	\label{hest-bench}
\end{table*}

\begin{table*}[!htbp]
	\caption{\textbf{Detailed imputation performance of the top 50 highly variable genes (HVGs) in Clear Cell Renal Cell Carcinoma (CCRCC).} The \textbf{best} model is highlighted in bold, while the \underline{second-best} is underlined.}
	\vspace{-0.5cm}
	\begin{center}
		\begin{tabular}{l|ccccccccc}
			\toprule
			HVG      & OmiCLIP & PLIP           & CHIEF             & MUSK              & CONCH             & UNI               & GigaPath          & Virchow2          & STAMP             \\
			\midrule
			AKAP8    & 0.120   & 0.180          & \underline{0.204} & 0.186             & 0.181             & 0.199             & 0.183             & \textbf{0.237}    & 0.194             \\
			BLVRB    & 0.145   & 0.241          & 0.231             & 0.219             & \underline{0.251} & 0.228             & 0.234             & \textbf{0.254}    & \underline{0.251} \\
			BTF3     & 0.214   & 0.256          & 0.301             & 0.309             & 0.329             & 0.306             & \underline{0.330} & 0.323             & \textbf{0.344}    \\
			CD109    & 0.156   & 0.300          & 0.368             & 0.348             & 0.364             & 0.363             & 0.344             & \textbf{0.420}    & \underline{0.418} \\
			CD14     & -0.050  & 0.064          & 0.003             & 0.068             & 0.081             & 0.083             & 0.092             & \underline{0.123} & \textbf{0.193}    \\
			CHST15   & 0.126   & 0.222          & 0.249             & 0.252             & 0.199             & 0.238             & 0.248             & \underline{0.289} & \textbf{0.302}    \\
			CIB1     & 0.131   & 0.237          & 0.288             & 0.263             & 0.239             & 0.262             & 0.272             & \underline{0.298} & \textbf{0.318}    \\
\bottomrule
		\end{tabular}
	\end{center}
	\label{hest-bench-ccrcc}
\end{table*}

\begin{table*}[!htbp]
	\caption{\textbf{Detailed imputation performance of the top 50 highly variable genes (HVGs) in Colorectal Adenocarcinoma (COAD).} The \textbf{best} model is highlighted in bold, while the \underline{second-best} is underlined.}
	\vspace{-0.5cm}
	\begin{center}
		\begin{tabular}{l|ccccccccc}
			\toprule
			HVG      & OmiCLIP         & PLIP              & CHIEF             & MUSK              & CONCH              & UNI               & GigaPath          & Virchow2          & STAMP             \\
			\midrule
			AQP8     & 0.064           & 0.203             & 0.189             & 0.210             & 0.206              & \textbf{0.280}    & \underline{0.256} & 0.243             & 0.235             \\
			AREG     & -0.064          & -0.161            & -0.006            & 0.098             & \textbf{0.302}     & 0.173             & 0.194             & 0.245             & \underline{0.281} \\
			BANK1    & 0.323           & 0.421             & 0.455             & 0.509             & 0.393              & 0.525             & \textbf{0.559}    & \underline{0.538} & 0.493             \\
			CA1      & 0.027           & -0.008            & 0.015             & 0.036             & -0.045             & \underline{0.154} & 0.082             & \textbf{0.163}    & 0.153             \\
			CA4      & -0.021          & \underline{0.168} & 0.019             & \textbf{0.175}    & 0.096              & -0.052            & 0.016             & -0.089            & 0.062             \\
			CALB2    & 0.131           & 0.026             & 0.194             & 0.147             & 0.148              & \underline{0.217} & 0.184             & 0.158             & \textbf{0.231}    \\
			CCL20    & 0.179           & 0.368             & 0.371             & 0.391             & 0.512              & 0.480             & \underline{0.530} & 0.490             & \textbf{0.539}    \\
\bottomrule
		\end{tabular}
	\end{center}
	\label{hest-bench-coad}
\end{table*}

\begin{table*}[!htbp]
	\caption{\textbf{Detailed imputation performance of the top 50 highly variable genes (HVGs) in Invasive Ductal Carcinoma (IDC).} The \textbf{best} model is highlighted in bold, while the \underline{second-best} is underlined.}
	\vspace{-0.5cm}
	\begin{center}
		\begin{tabular}{l|ccccccccc}
			\toprule
			HVG      & OmiCLIP & PLIP              & CHIEF             & MUSK              & CONCH             & UNI               & GigaPath          & Virchow2          & STAMP             \\
			\midrule
			ABCC11   & 0.578   & 0.654             & 0.661             & 0.710             & 0.719             & 0.736             & 0.712             & \underline{0.764} & \textbf{0.791}    \\
			ADH1B    & 0.043   & 0.186             & 0.311             & 0.266             & 0.371             & 0.418             & 0.351             & \underline{0.473} & \textbf{0.543}    \\
			ADIPOQ   & 0.307   & 0.477             & 0.502             & 0.515             & 0.497             & \underline{0.536} & 0.510             & 0.528             & \textbf{0.594}    \\
			ANKRD30A & 0.234   & 0.502             & 0.474             & 0.536             & 0.504             & 0.539             & 0.425             & \underline{0.540} & \textbf{0.587}    \\
			AQP1     & 0.328   & 0.379             & 0.498             & 0.555             & 0.596             & 0.648             & 0.615             & \underline{0.677} & \textbf{0.706}    \\
			AQP3     & 0.342   & \underline{0.409} & 0.404             & \textbf{0.435}    & 0.371             & 0.401             & 0.364             & 0.373             & 0.359             \\
			CCR7     & 0.512   & 0.561             & 0.570             & 0.597             & 0.559             & \textbf{0.625}    & \underline{0.611} & 0.604             & 0.605             \\
\bottomrule
		\end{tabular}
	\end{center}
	\label{hest-bench-idc}
\end{table*}

\begin{table*}[!htbp]
	\caption{\textbf{Detailed imputation performance of the top 50 highly variable genes (HVGs) in Lung Cancer (LUNG).} The \textbf{best} model is highlighted in bold, while the \underline{second-best} is underlined.}
	\vspace{-0.5cm}
	\begin{center}
		\begin{tabular}{l|ccccccccc}
			\toprule
			HVG      & OmiCLIP           & PLIP              & CHIEF          & MUSK              & CONCH             & UNI               & GigaPath          & Virchow2          & STAMP             \\
			\midrule
			ACE2     & 0.660             & 0.675             & \textbf{0.712} & \textbf{0.712}    & 0.645             & 0.709             & \underline{0.711} & \underline{0.711} & 0.689             \\
			ALDH1A3  & 0.626             & 0.674             & 0.706          & 0.666             & 0.589             & 0.702             & 0.675             & \underline{0.725} & \textbf{0.749}    \\
			BBOX1    & 0.192             & 0.288             & 0.288          & 0.231             & 0.310             & 0.362             & \underline{0.389} & \underline{0.389} & \textbf{0.408}    \\
			CAPN8    & 0.740             & 0.828             & 0.823          & \textbf{0.847}    & 0.829             & 0.828             & 0.825             & 0.834             & \underline{0.842} \\
			CCL19    & 0.376             & 0.422             & 0.403          & 0.464             & 0.512             & 0.499             & 0.476             & \underline{0.557} & \textbf{0.560}    \\
			CCR7     & 0.596             & 0.614             & 0.593          & 0.634             & 0.652             & \underline{0.656} & 0.654             & 0.618             & \textbf{0.674}    \\
			CD163    & 0.535             & 0.607             & 0.684          & 0.660             & 0.687             & \underline{0.778} & 0.759             & 0.776             & \textbf{0.804}    \\
\bottomrule
		\end{tabular}
	\end{center}
	\label{hest-bench-lung}
\end{table*}

\begin{table*}[!htbp]
	\caption{\textbf{Detailed imputation performance of the top 50 highly variable genes (HVGs) in Pancreatic Adenocarcinoma (PAAD).} The \textbf{best} model is highlighted in bold, while the \underline{second-best} is underlined.}
	\vspace{-0.5cm}
	\begin{center}
		\begin{tabular}{l|ccccccccc}
			\toprule
			HVG      & OmiCLIP & PLIP              & CHIEF             & MUSK              & CONCH             & UNI               & GigaPath          & Virchow2          & STAMP             \\
			\midrule
			ACE2     & 0.179   & 0.435             & 0.528             & 0.491             & 0.382             & 0.554             & 0.527             & \underline{0.587} & \textbf{0.595}    \\
			ACTA2    & 0.206   & 0.305             & 0.382             & 0.425             & \underline{0.467} & 0.430             & 0.372             & 0.466             & \textbf{0.525}    \\
			AIF1     & 0.450   & 0.500             & \underline{0.578} & 0.529             & 0.513             & 0.568             & 0.577             & 0.541             & \textbf{0.605}    \\
			ANPEP    & 0.535   & 0.604             & \underline{0.624} & 0.603             & 0.518             & 0.618             & 0.568             & 0.479             & \textbf{0.627}    \\
			BANK1    & 0.330   & 0.431             & \textbf{0.496}    & 0.466             & 0.440             & 0.460             & \underline{0.489} & 0.389             & 0.487             \\
			CCL19    & 0.291   & 0.320             & 0.392             & 0.394             & 0.382             & \underline{0.447} & 0.394             & 0.431             & \textbf{0.458}    \\
			CCL5     & 0.427   & 0.507             & 0.582             & 0.596             & 0.613             & 0.640             & \underline{0.651} & 0.612             & \textbf{0.690}    \\
\bottomrule
		\end{tabular}
	\end{center}
	\label{hest-bench-paad}
\end{table*}

\begin{table*}[!htbp]
	\caption{\textbf{Detailed imputation performance of the top 50 highly variable genes (HVGs) in Prostate Adenocarcinoma (PRAD).} The \textbf{best} model is highlighted in bold, while the \underline{second-best} is underlined.}
	\vspace{-0.5cm}
	\begin{center}
		\begin{tabular}{l|ccccccccc}
			\toprule
			HVG     & OmiCLIP & PLIP  & CHIEF             & MUSK              & CONCH             & UNI               & GigaPath          & Virchow2       & STAMP             \\
			\midrule
			AMD1    & 0.207   & 0.382 & 0.461             & 0.449             & \underline{0.528} & 0.462             & \textbf{0.543}    & 0.468          & 0.471             \\
			C1orf43 & 0.210   & 0.290 & 0.389             & 0.350             & 0.386             & 0.300             & \underline{0.403} & 0.368          & \textbf{0.411}    \\
			C6orf48 & 0.140   & 0.186 & 0.232             & 0.226             & 0.222             & 0.215             & \textbf{0.270}    & 0.244          & \underline{0.266} \\
			C8orf59 & 0.196   & 0.280 & 0.313             & 0.305             & \underline{0.327} & 0.270             & 0.321             & 0.313          & \textbf{0.333}    \\
			CAT     & 0.182   & 0.293 & 0.349             & 0.350             & \textbf{0.360}    & 0.311             & \underline{0.355} & 0.318          & 0.350             \\
			CCDC47  & 0.140   & 0.244 & \underline{0.345} & 0.318             & 0.331             & 0.296             & \textbf{0.356}    & 0.303          & \underline{0.345} \\
			COX17   & 0.272   & 0.367 & 0.391             & 0.405             & \underline{0.411} & 0.337             & \underline{0.411} & 0.401          & \textbf{0.427}    \\
\bottomrule
		\end{tabular}
	\end{center}
	\label{hest-bench-prad}
\end{table*}

\begin{table*}[!htbp]
	\caption{\textbf{Detailed imputation performance of the top 50 highly variable genes (HVGs) in Rectum Adenocarcinoma (READ).} The \textbf{best} model is highlighted in bold, while the \underline{second-best} is underlined.}
	\vspace{-0.5cm}
	\begin{center}
		\begin{tabular}{l|ccccccccc}
			\toprule
			HVG        & OmiCLIP & PLIP           & CHIEF             & MUSK              & CONCH             & UNI               & GigaPath          & Virchow2          & STAMP             \\
			\midrule
			AC010883.3 & 0.000   & -0.031         & 0.060             & \textbf{0.091}    & \underline{0.061} & 0.019             & 0.024             & 0.033             & 0.042             \\
			ADAMDEC1   & -0.023  & 0.074          & 0.041             & 0.026             & 0.161             & \underline{0.193} & 0.107             & 0.177             & \textbf{0.231}    \\
			ANGPTL2    & -0.016  & 0.063          & 0.027             & 0.026             & 0.161             & 0.105             & 0.154             & \textbf{0.256}    & \underline{0.250} \\
			APOC1      & 0.081   & 0.118          & 0.085             & 0.085             & 0.095             & 0.131             & \underline{0.269} & 0.266             & \textbf{0.278}    \\
			BTNL3      & 0.065   & 0.097          & 0.166             & 0.106             & 0.252             & \underline{0.319} & \underline{0.319} & 0.299             & \textbf{0.404}    \\
			CA1        & 0.200   & 0.130          & 0.155             & 0.108             & 0.280             & 0.331             & \underline{0.398} & 0.324             & \textbf{0.466}    \\
			CA2        & 0.259   & 0.179          & 0.219             & 0.179             & 0.411             & 0.462             & \textbf{0.531}    & 0.388             & \underline{0.523} \\
\bottomrule
		\end{tabular}
	\end{center}
	\label{hest-bench-read}
\end{table*}

\begin{table*}[!htbp]
	\caption{\textbf{Detailed imputation performance of the top 50 highly variable genes (HVGs) in Skin Cutaneous Melanoma (SKCM).} The \textbf{best} model is highlighted in bold, while the \underline{second-best} is underlined.}
	\vspace{-0.5cm}
	\begin{center}
		\begin{tabular}{l|ccccccccc}
			\toprule
			HVG      & OmiCLIP        & PLIP              & CHIEF          & MUSK              & CONCH             & UNI               & GigaPath          & Virchow2          & STAMP             \\
			\midrule
			ACER1    & 0.436          & 0.478             & 0.539          & 0.573             & 0.670             & \underline{0.701} & 0.604             & 0.670             & \textbf{0.742}    \\
			AHNAK2   & 0.505          & 0.463             & 0.519          & 0.568             & 0.642             & \underline{0.698} & 0.550             & 0.696             & \textbf{0.750}    \\
			AQP3     & 0.537          & 0.650             & 0.679          & 0.621             & 0.738             & \underline{0.825} & 0.743             & 0.850             & \textbf{0.871}    \\
			BIRC3    & 0.533          & 0.624             & 0.684          & 0.719             & 0.735             & 0.711             & 0.610             & \underline{0.771} & \textbf{0.782}    \\
			CAPNS2   & 0.538          & 0.547             & 0.613          & 0.656             & 0.677             & \underline{0.796} & 0.649             & 0.775             & \textbf{0.805}    \\
			CCL19    & 0.217          & 0.347             & 0.498          & 0.496             & 0.432             & 0.426             & 0.363             & \underline{0.569} & \textbf{0.579}    \\
			CCL22    & \textbf{0.266} & \underline{0.257} & 0.199          & 0.055             & 0.188             & 0.213             & 0.208             & 0.188             & 0.161             \\
\bottomrule
		\end{tabular}
	\end{center}
	\label{hest-bench-skcm}
\end{table*}

\begin{table*}[!htbp]
	\caption{\textbf{Performance of spatial domain recognition on breast samples.} We report the Adjusted Rand Index (ARI), Normalized Mutual Information (NMI), Homogeneity (HOM), and Completeness (COM) scores.}
	\begin{center}
		\begin{tabular}{llcm{8cm}}
			\toprule
			\textbf{Organ} & \textbf{Platform} & \textbf{Sections} & \textbf{Annotation Classes}                                                                                                  \\
			\midrule
			Breast         & 2D-array ST       & 8                 & Invasive Cancer, Cancer In Situ, Connective Tissue, Immune Infiltrate, Breast Glands, Adipose Tissue                         \\[15pt]
			Kidney         & 10x Visium        & 3                 & Tumor (Tum), Normal (Nor), Tertiary Lymphoid Structures (Tls), Immune Infiltration (Infl)                                    \\[15pt]
			Lung           & 10x Visium        & 5                 & Tumor (Tum), Normal (Nor), Tertiary Lymphoid Structures (Tls), Immune Infiltration (Infl), Lymph Node (Ln)                   \\[15pt]
			Prostate       & 10x Visium        & 7                 & Benign, Stroma, GG1, GG2, GG4, GG4 Cribriform, PIN, Transition State, Chronic Inflammation, Inflammation, Fat, Vessel, Nerve \\
			\bottomrule
		\end{tabular}
	\end{center}
	\label{spatial-domain-dataset}
\end{table*}

\begin{table*}[!htbp]
	\setlength{\tabcolsep}{11pt}
	\caption{\textbf{Performance of spatial domain recognition on breast samples.} We report the Adjusted Rand Index (ARI), Normalized Mutual Information (NMI), Homogeneity (HOM), and Completeness (COM) scores.}
	% \vspace{-0.5cm}
	\fontsize{9}{10}\selectfont
	\begin{center}
		\begin{tabular}{l|lc|cccccccc}
			\toprule

			Model                     & \multicolumn{2}{c|}{Metrics} & A1    & B1    & C1    & D1    & E1    & F1    & G2    & H1            \\
			\midrule
			\multirow{4}{*}{OmiCLIP}  & ARI                          & 0.241 & 0.057 & 0.444 & 0.245 & 0.288 & 0.127 & 0.118 & 0.237 & 0.411 \\
			                          & NMI                          & 0.293 & 0.159 & 0.499 & 0.304 & 0.259 & 0.181 & 0.156 & 0.289 & 0.500 \\
			                          & HOM                          & 0.344 & 0.266 & 0.488 & 0.403 & 0.305 & 0.213 & 0.246 & 0.311 & 0.522 \\
			                          & COM                          & 0.264 & 0.113 & 0.511 & 0.244 & 0.225 & 0.157 & 0.114 & 0.270 & 0.480 \\
			\midrule
			\multirow{4}{*}{PLIP}     & ARI                          & 0.284 & 0.067 & 0.408 & 0.291 & 0.433 & 0.356 & 0.135 & 0.194 & 0.391 \\
			                          & NMI                          & 0.334 & 0.205 & 0.452 & 0.397 & 0.338 & 0.346 & 0.145 & 0.278 & 0.510 \\
			                          & HOM                          & 0.396 & 0.334 & 0.456 & 0.527 & 0.367 & 0.413 & 0.232 & 0.305 & 0.532 \\
\end{tabular}
	\end{center}
	\label{spatial-breast}
\end{table*}

\begin{table*}[!htbp]
	\setlength{\tabcolsep}{12pt}
	\caption{\textbf{Performance of spatial domain recognition on lung samples.} We report the Adjusted Rand Index (ARI), Normalized Mutual Information (NMI), Homogeneity (HOM), and Completeness (COM) scores.}
	\vspace{-0.5cm}
	\fontsize{9}{10}\selectfont
	\begin{center}
		\begin{tabular}{l|lc|ccccc}
			\toprule
			Model                     & \multicolumn{2}{c|}{Metrics} & LC1   & LC2   & LC3   & LC4   & LC5           \\
			\midrule
			\multirow{4}{*}{OmiCLIP}  & ARI                          & 0.574 & 0.324 & 0.725 & 0.590 & 0.620 & 0.613 \\
			                          & NMI                          & 0.621 & 0.392 & 0.732 & 0.591 & 0.704 & 0.684 \\
			                          & HOM                          & 0.695 & 0.483 & 0.824 & 0.658 & 0.739 & 0.772 \\
			                          & COM                          & 0.562 & 0.329 & 0.658 & 0.536 & 0.672 & 0.615 \\
			\midrule
			\multirow{4}{*}{PLIP}     & ARI                          & 0.566 & 0.423 & 0.729 & 0.295 & 0.647 & 0.735 \\
			                          & NMI                          & 0.614 & 0.519 & 0.740 & 0.456 & 0.699 & 0.654 \\
			                          & HOM                          & 0.677 & 0.628 & 0.831 & 0.515 & 0.722 & 0.688 \\
\end{tabular}
	\end{center}
	\label{spatial-lung}
\end{table*}

\begin{table*}[!htbp]
	\setlength{\tabcolsep}{12pt}
	\caption{\textbf{Performance of spatial domain recognition on kidney samples.} We report the Adjusted Rand Index (ARI), Normalized Mutual Information (NMI), Homogeneity (HOM), and Completeness (COM) scores.}
	\vspace{-0.5cm}
	\fontsize{9}{10}\selectfont
	\begin{center}
		\begin{tabular}{l|lc|ccc}
			\toprule
			Model                     & \multicolumn{2}{c|}{Metrics} & KC1   & KC2   & KC3           \\
			\midrule
			\multirow{4}{*}{OmiCLIP}  & ARI                          & 0.369 & 0.486 & 0.367 & 0.253 \\
			                          & NMI                          & 0.482 & 0.615 & 0.495 & 0.335 \\
			                          & HOM                          & 0.604 & 0.797 & 0.641 & 0.373 \\
			                          & COM                          & 0.403 & 0.501 & 0.403 & 0.304 \\
			\midrule
			\multirow{4}{*}{PLIP}     & ARI                          & 0.474 & 0.346 & 0.628 & 0.449 \\
			                          & NMI                          & 0.588 & 0.474 & 0.750 & 0.540 \\
			                          & HOM                          & 0.714 & 0.616 & 0.920 & 0.607 \\
\end{tabular}
	\end{center}
	\label{spatial-kidney}
\end{table*}

\begin{table*}[!htbp]
	\setlength{\tabcolsep}{12pt}
	\caption{\textbf{Performance of spatial domain recognition on prostate samples.} We report the Adjusted Rand Index (ARI), Normalized Mutual Information (NMI), Homogeneity (HOM), and Completeness (COM) scores.}
	\vspace{-0.5cm}
	\fontsize{9}{10}\selectfont
	\begin{center}
		\begin{tabular}{l|lc|ccccccc}
			\toprule
			Model                     & \multicolumn{2}{c|}{Metrics} & H1\_2 & H1\_4 & H1\_5 & H2\_1 & H2\_2 & H2\_5 & V1\_2         \\
			\midrule
			\multirow{4}{*}{OmiCLIP}  & ARI                          & 0.301 & 0.323 & 0.296 & 0.209 & 0.211 & 0.281 & 0.527 & 0.263 \\
			                          & NMI                          & 0.378 & 0.338 & 0.408 & 0.380 & 0.373 & 0.322 & 0.561 & 0.267 \\
			                          & HOM                          & 0.487 & 0.394 & 0.449 & 0.722 & 0.453 & 0.489 & 0.598 & 0.302 \\
			                          & COM                          & 0.322 & 0.296 & 0.374 & 0.258 & 0.317 & 0.240 & 0.529 & 0.239 \\
			\midrule
			\multirow{4}{*}{PLIP}     & ARI                          & 0.317 & 0.348 & 0.426 & 0.155 & 0.239 & 0.262 & 0.553 & 0.235 \\
			                          & NMI                          & 0.386 & 0.399 & 0.480 & 0.326 & 0.377 & 0.282 & 0.537 & 0.301 \\
			                          & HOM                          & 0.494 & 0.470 & 0.534 & 0.645 & 0.468 & 0.426 & 0.566 & 0.350 \\
\end{tabular}
	\end{center}
	\label{spatial-prostate}
\end{table*}

\begin{table*}[!htbp]
	\caption{\textbf{Predictive performance of diagnostic immunohistochemical biomarkers on the internal breast cancer cohort (biopsy).} Performance of inferring Estrogen Receptor (ER), Progesterone Receptor (PR), and Human Epidermal Growth Factor Receptor 2 (HER2) statuses from biopsy whole-slide images is quantified by AUROC (95\% confidence intervals). ``Gain'' indicates the absolute AUROC improvement of STAMP over the baseline (Virchow2). The \textbf{best} model is highlighted in bold, while the \underline{second-best} is underlined. $P$-values are derived from a one-sided Wilcoxon signed-rank test comparing the best and second-best models across bootstrapped distributions.}
	\fontsize{9}{10}\selectfont
	\begin{center}
		\begin{tabular}{l|cccccc|cc}
			\toprule
			Marker & OmiCLIP         & PLIP & CONCH & UNI & Virchow2 & STAMP & Gain & $P$-value \\
			\midrule
			ER     & \makecell{0.703                                                            \\(0.645-0.768)} & \makecell{0.775\\(0.718-0.828)} & \makecell{0.842\\(0.796-0.886)} & \makecell{\underline{0.862}\\(0.818-0.908)} & \makecell{0.859\\(0.812-0.902)} & \makecell{\textbf{0.864}\\(0.819-0.908)} &+0.5\% &$P$=0.004\\
			\midrule
			PR     & \makecell{0.658                                                            \\(0.589-0.728)} & \makecell{0.698\\(0.634-0.759)} & \makecell{0.725\\(0.666-0.782)} & \makecell{0.732\\(0.672-0.788)} & \makecell{\underline{0.746}\\(0.682-0.804)} & \makecell{\textbf{0.767}\\(0.714-0.819)} &+2.1\% & $P$<0.001 \\
			\midrule
			HER2   & \makecell{0.631                                                            \\(0.565-0.700)} & \makecell{0.762\\(0.702-0.817)} & \makecell{0.800\\(0.745-0.849)} & \makecell{0.797\\(0.739-0.847)} & \makecell{\underline{0.803}\\(0.749-0.853)} & \makecell{\textbf{0.832}\\(0.778-0.880)} &+2.9\% & $P$<0.001\\

			\bottomrule
		\end{tabular}
	\end{center}
	\label{breast-diagnostic-internal-biopsy}
\end{table*}

\begin{table*}[!htbp]
	\caption{\textbf{Predictive performance of diagnostic immunohistochemical biomarkers on the external breast cancer cohort (biopsy).} Performance of inferring Estrogen Receptor (ER), Progesterone Receptor (PR), and Human Epidermal Growth Factor Receptor 2 (HER2) statuses from biopsy whole-slide images is quantified by AUROC (95\% confidence intervals). ``Gain'' indicates the absolute AUROC improvement of STAMP over the baseline (Virchow2). The \textbf{best} model is highlighted in bold, while the \underline{second-best} is underlined. $P$-values are derived from a one-sided Wilcoxon signed-rank test comparing the best and second-best models across bootstrapped distributions.}
	\fontsize{9}{10}\selectfont
	\begin{center}
		\begin{tabular}{l|cccccc|cc}
			\toprule
			Marker & OmiCLIP         & PLIP & CONCH & UNI & Virchow2 & STAMP & Gain & $P$-value \\
			\midrule
			ER     & \makecell{0.643                                                            \\(0.595-0.687)} & \makecell{0.738\\(0.696-0.781)} & \makecell{0.797\\(0.757-0.833)} & \makecell{\underline{0.842}\\(0.807-0.873)} & \makecell{0.837\\(0.802-0.867)} & \makecell{\textbf{0.861}\\(0.826-0.890)} & +2.4\%& $P$<0.001 \\
			\midrule
			PR     & \makecell{0.638                                                            \\(0.593-0.680)} & \makecell{0.668\\(0.624-0.713)} & \makecell{0.694\\(0.654-0.736)} & \makecell{0.702\\(0.659-0.741)} & \makecell{\underline{0.720}\\(0.679-0.759)} & \makecell{\textbf{0.765}\\(0.727-0.800)} & +4.5\%& $P$<0.001 \\
			\midrule
			HER2   & \makecell{0.570                                                            \\(0.528-0.616)} & \makecell{0.643\\(0.603-0.684)} & \makecell{0.672\\(0.631-0.712)} & \makecell{0.675\\(0.634-0.716)} & \makecell{\underline{0.706}\\(0.667-0.744)} & \makecell{\textbf{0.720}\\(0.682-0.758)} & +1.4\%& $P$<0.001 \\
			\bottomrule
		\end{tabular}
	\end{center}
	\label{breast-diagnostic-external-biopsy}
\end{table*}

\begin{table*}[!htbp]
	\caption{\textbf{Predictive performance of diagnostic immunohistochemical biomarkers on the internal breast cancer cohort (resection).} Performance of inferring Estrogen Receptor (ER), Progesterone Receptor (PR), and Human Epidermal Growth Factor Receptor 2 (HER2) statuses from resection whole-slide images is quantified by AUROC (95\% confidence intervals). ``Gain'' indicates the absolute AUROC improvement of STAMP over the baseline (Virchow2). The \textbf{best} model is highlighted in bold, while the \underline{second-best} is underlined. $P$-values are derived from a one-sided Wilcoxon signed-rank test comparing the best and second-best models across bootstrapped distributions.}
	\fontsize{9}{10}\selectfont
	\begin{center}
		\begin{tabular}{l|cccccc|cc}
			\toprule
			Marker & OmiCLIP         & PLIP & CONCH & UNI & Virchow2 & STAMP & Gain & $P$-value \\
			\midrule
			ER     & \makecell{0.801                                                            \\ (0.750-0.851)} 	& \makecell{0.882\\ (0.844-0.916)} 	& \makecell{0.892\\ (0.854-0.927)} 	& \makecell{0.912 \\(0.879-0.939)} 	& \makecell{\underline{0.919}\\ (0.888-0.946)} 	& \makecell{\textbf{0.944}\\ (0.919-0.965)}&+2.5\%&$P$<0.001\\
			\midrule
			PR     & \makecell{0.860                                                            \\ (0.818-0.902)} 	& \makecell{0.879\\ (0.839-0.917)} 	& \makecell{0.889\\ (0.849-0.922)} 	& \makecell{\underline{0.911} \\(0.874-0.941)} 	& \makecell{0.897\\ (0.859-0.930)} 	& \makecell{\textbf{0.926}\\ (0.895-0.953)}&+2.9\%&$P$<0.001\\
			\midrule
			HER2   & \makecell{0.762                                                            \\ (0.696-0.829)} 	& \makecell{0.919\\ (0.878-0.956)} 	& \makecell{0.925\\ (0.887-0.958)} 	& \makecell{0.950 \\(0.919-0.975)} 	& \makecell{\underline{0.958}\\ (0.932-0.978)} 	& \makecell{\textbf{0.977}\\ (0.957-0.990)}&+1.9\%&$P$<0.001\\
			\bottomrule
		\end{tabular}
	\end{center}
	\label{breast-diagnostic-internal-resection}
\end{table*}

\begin{table*}[!htbp]
	\caption{\textbf{Predictive performance of diagnostic immunohistochemical biomarkers on the external breast cancer cohort (resection).} Performance of inferring Estrogen Receptor (ER), Progesterone Receptor (PR), and Human Epidermal Growth Factor Receptor 2 (HER2) statuses from resection whole-slide images is quantified by AUROC (95\% confidence intervals). ``Gain'' indicates the absolute AUROC improvement of STAMP over the baseline (Virchow2). The \textbf{best} model is highlighted in bold, while the \underline{second-best} is underlined. $P$-values are derived from a one-sided Wilcoxon signed-rank test comparing the best and second-best models across bootstrapped distributions.}
	\fontsize{9}{10}\selectfont
	\begin{center}
		\begin{tabular}{l|cccccc|cc}
			\toprule
			Marker & OmiCLIP         & PLIP & CONCH & UNI & Virchow2 & STAMP & Gain & $P$-value \\
			\midrule
			ER     & \makecell{0.480                                                            \\ (0.412-0.546)}	& \makecell{0.699\\ (0.632-0.764)}	& \makecell{0.809\\ (0.752-0.862)}	& \makecell{0.819\\ (0.769-0.865)}	& \makecell{\underline{0.823}\\ (0.765-0.875)}	& \makecell{\textbf{0.859}\\ (0.815-0.903)}&+3.6\%&$P$<0.001\\
			\midrule
			PR     & \makecell{0.506                                                            \\ (0.438-0.567)}	& \makecell{0.672\\ (0.608-0.733)}	& \makecell{0.742\\ (0.688-0.794)}	& \makecell{\underline{0.776}\\ (0.725-0.825)}	& \makecell{0.772\\ (0.720-0.822)}	& \makecell{\textbf{0.804}\\ (0.755-0.849)}&+3.2\%&$P$<0.001\\
			\midrule
			HER2   & \makecell{0.515                                                            \\ (0.436-0.598)}	& \makecell{0.692\\ (0.611-0.775)}	& \makecell{0.728\\ (0.649-0.806)}	& \makecell{0.814\\ (0.742-0.879)}	& \makecell{\underline{0.844}\\ (0.773-0.905)}	& \makecell{\textbf{0.855}\\ (0.786-0.912)}&+1.1\%&$P$<0.001\\
			\bottomrule
		\end{tabular}
	\end{center}
	\label{breast-diagnostic-external-resection}
\end{table*}

\begin{table*}[!htbp]
	\caption{\textbf{Predictive performance of diagnostic immunohistochemical biomarkers on the internal lung cancer cohort.} Performance of inferring Napsin A, thyroid transcription factor-1 (TTF-1), p40, and cytokeratin 7 (CK7) statuses from whole-slide images is quantified by AUROC (95\% confidence intervals). ``Gain'' indicates the absolute AUROC improvement of STAMP over the baseline (Virchow2). The \textbf{best} model is highlighted in bold, while the \underline{second-best} is underlined. $P$-values are derived from a one-sided Wilcoxon signed-rank test comparing the best and second-best models across bootstrapped distributions.}
	\fontsize{9}{10}\selectfont
	\begin{center}
		\begin{tabular}{l|cccccc|cc}
			\toprule
			Marker   & OmiCLIP         & PLIP & CONCH & UNI & Virchow2 & STAMP & Gain & $P$-value \\
			\midrule

			Napsin A & \makecell{0.775                                                            \\ (0.695-0.852)}	& \makecell{0.838\\ (0.769-0.897)}	& \makecell{\underline{0.893}\\ (0.839-0.939)}	& \makecell{0.870\\ (0.808-0.928)}	& \makecell{0.871\\ (0.815-0.924)}	& \makecell{\textbf{0.908}\\ (0.860-0.950)}&+3.7\%&$P$<0.001\\
			\midrule
			TTF-1    & \makecell{0.767                                                            \\ (0.700-0.828)}	& \makecell{0.809\\ (0.753-0.861)}	& \makecell{0.861\\ (0.810-0.909)}	& \makecell{\underline{0.923}\\ (0.887-0.956)}	& \makecell{0.894\\ (0.851-0.933)}	& \makecell{\textbf{0.925}\\ (0.891-0.955)}&+3.1\% & $P$=0.002 \\
			\midrule
			P40      & \makecell{0.760                                                            \\ (0.686-0.829)}	& \makecell{0.840\\ (0.776-0.898)}	& \makecell{0.919\\ (0.874-0.958)}	& \makecell{0.920\\ (0.875-0.956)}	& \makecell{\underline{0.939}\\ (0.900-0.973)}	& \makecell{\textbf{0.958}\\ (0.928-0.982)}&+1.9\% &$P$<0.001\\
			\midrule
			CK7      & \makecell{0.661                                                            \\ (0.559-0.749)}	& \makecell{0.752\\ (0.663-0.830)}	& \makecell{\underline{0.813}\\ (0.741-0.877)}	& \makecell{0.799\\ (0.730-0.864)}	& \makecell{0.806\\ (0.734-0.870)}	& \makecell{\textbf{0.821}\\ (0.748-0.885)}&+1.5\%&$P$<0.001\\
			\bottomrule
		\end{tabular}
		\label{lung-diagnostic-internal}
	\end{center}
\end{table*}

\begin{table*}[!htbp]
	\caption{\textbf{Predictive performance of diagnostic immunohistochemical biomarkers on the external lung cancer cohort.} Performance of inferring Napsin A, thyroid transcription factor-1 (TTF-1), p40, and cytokeratin 7 (CK7) statuses from whole-slide images is quantified by AUROC (95\% confidence intervals). ``Gain'' indicates the absolute AUROC improvement of STAMP over the baseline (Virchow2). The \textbf{best} model is highlighted in bold, while the \underline{second-best} is underlined. $P$-values are derived from a one-sided Wilcoxon signed-rank test comparing the best and second-best models across bootstrapped distributions.}
	\fontsize{9}{10}\selectfont
	\begin{center}
		\begin{tabular}{l|cccccc|cc}
			\toprule
			Marker   & OmiCLIP         & PLIP & CONCH & UNI & Virchow2 & STAMP & Gain & $P$-value \\
			\midrule

			Napsin A & \makecell{0.834                                                            \\ (0.745-0.909)}	& \makecell{0.832\\ (0.735-0.916)}	& \makecell{0.860\\ (0.778-0.929)}	& \makecell{\underline{0.871}\\ (0.794-0.932)}	& \makecell{0.839\\ (0.751-0.918)}	& \makecell{\textbf{0.890}\\ (0.821-0.947)}&+5.1\%&$P$<0.001\\
			\midrule
			TTF-1    & \makecell{0.688                                                            \\ (0.646-0.730)}	& \makecell{0.693\\ (0.649-0.738)}	& \makecell{\underline{0.842}\\ (0.809-0.875)}	& \makecell{0.811\\ (0.771-0.844)}	& \makecell{\underline{0.842}\\ (0.804-0.872)}	& \makecell{\textbf{0.869}\\ (0.838-0.898)}&+2.7\%&$P$<0.001\\
			\midrule
			P40      & \makecell{0.682                                                            \\ (0.601-0.766)}	& \makecell{0.818\\ (0.736-0.887)}	& \makecell{0.876\\ (0.790-0.946)}	& \makecell{0.855\\ (0.771-0.929)}	& \makecell{\underline{0.916}\\ (0.858-0.961)}	& \makecell{\textbf{0.935}\\ (0.877-0.980)}&+1.9\%&$P$<0.001\\
			\midrule
			CK7      & \makecell{0.627                                                            \\ (0.576-0.676)}	& \makecell{0.675\\ (0.621-0.724)}	& \makecell{\underline{0.819}\\ (0.781-0.855)}	& \makecell{0.765\\ (0.721-0.805)}	& \makecell{0.809\\ (0.773-0.847)}	& \makecell{\textbf{0.827}\\ (0.788-0.861)}&+1.8\%&$P$<0.001\\
			\bottomrule
		\end{tabular}
	\end{center}
	\label{lung-diagnostic-external}
\end{table*}

\begin{table*}[!htbp]
	\caption{\textbf{Predictive performance of actionable driver mutations on the internal cohort.} Performance of identifying PIK3CA, EGFR, KRAS, BRAF, and IDH mutations from whole-slide images is quantified by AUROC (95\% confidence intervals). ``Gain'' indicates the absolute AUROC improvement of STAMP over the baseline (Virchow2). The \textbf{best} model is highlighted in bold, while the \underline{second-best} is underlined. $P$-values are derived from a one-sided Wilcoxon signed-rank test comparing the best and second-best models across bootstrapped distributions.}
	\fontsize{9}{10}\selectfont
	\begin{center}
		\begin{tabular}{l|cccccc|cc}
			\toprule
			Marker & OmiCLIP & PLIP & CONCH & UNI & Virchow2 & STAMP & Gain & $P$-value \\
			\midrule
			\makecell{Breast                                                            \\PIK3CA} & \makecell{0.583                                                            \\(0.500-0.665)} & \makecell{0.615\\(0.532-0.698)} & \makecell{\underline{0.669}\\(0.587-0.746)} & \makecell{0.667\\(0.583-0.741)} & \makecell{0.652\\(0.570-0.731)} & \makecell{\textbf{0.686}\\(0.605-0.757)}  & +3.4\%& $P$<0.001\\
			\midrule
			\makecell{Lung                                                              \\EGFR}      & \makecell{0.699                                                            \\(0.629-0.764)} & \makecell{0.729\\(0.668-0.789)} & \makecell{0.741\\(0.676-0.802)} & \makecell{0.783\\(0.724-0.841)} & \makecell{\underline{0.802}\\(0.744-0.853)} & \makecell{\textbf{0.828}\\(0.777-0.876)}  & +2.6\%& $P$<0.001\\
			\midrule
			\makecell{Lung                                                              \\KRAS}      & \makecell{0.698                                                            \\(0.595-0.797)} & \makecell{0.686\\(0.582-0.785)} & \makecell{0.728\\(0.624-0.830)} & \makecell{\underline{0.760}\\(0.659-0.845)} & \makecell{0.747\\(0.645-0.838)} & \makecell{\textbf{0.806}\\(0.733-0.874)} & +5.9\%& $P$<0.001 \\
			\midrule
			\makecell{CRC                                                               \\BRAF}      & \makecell{0.527                                                            \\(0.349-0.699)} & \makecell{0.610\\(0.359-0.842)} & \makecell{0.730\\(0.536-0.903)} & \makecell{\underline{0.758}\\(0.588-0.907)} & \makecell{\underline{0.758}\\(0.592-0.901)} & \makecell{\textbf{0.760}\\(0.570-0.927)}  & +0.2\%& $P$=0.028\\
			\midrule
			\makecell{Brain                                                             \\IDH}         & \makecell{0.722                                                            \\(0.602-0.839)} & \makecell{0.865\\(0.781-0.932)} & \makecell{0.886\\(0.807-0.953)} & \makecell{0.922\\(0.841-0.979)} & \makecell{\underline{0.924}\\(0.850-0.975)} & \makecell{\textbf{0.945}\\(0.873-0.992)} & +2.1\%& $P$<0.001 \\
			\bottomrule
		\end{tabular}
	\end{center}
	\label{targeted-internal}
\end{table*}

\begin{table*}[!htbp]
	\caption{\textbf{Predictive performance of actionable driver mutations on the external cohort.} Performance of identifying PIK3CA, EGFR, KRAS, BRAF, and IDH mutations from whole-slide images is quantified by AUROC (95\% confidence intervals). ``Gain'' indicates the absolute AUROC improvement of STAMP over the baseline (Virchow2). The \textbf{best} model is highlighted in bold, while the \underline{second-best} is underlined. $P$-values are derived from a one-sided Wilcoxon signed-rank test comparing the best and second-best models across bootstrapped distributions.}
	\fontsize{9}{10}\selectfont
	\begin{center}
		\begin{tabular}{l|cccccc|cc}
			\toprule
			Marker & OmiCLIP & PLIP & CONCH & UNI & Virchow2 & STAMP & Gain & $P$-value \\
			\midrule
			\makecell{Breast                                                            \\PIK3CA} & \makecell{0.540										 \\(0.414-0.662)} & \makecell{0.611\\(0.496-0.714)} & \makecell{0.633\\(0.526-0.745)} & \makecell{0.635\\(0.522-0.747)} & \makecell{\underline{0.679}\\(0.561-0.789)} & \makecell{\textbf{0.703}\\(0.597-0.804)} & +2.4\%& $P$<0.001\\
			\midrule
			\makecell{Lung                                                              \\EGFR}       & \makecell{0.623										 \\(0.541-0.705)} & \makecell{0.652\\(0.564-0.730)} & \makecell{0.708\\(0.637-0.774)} & \makecell{0.772\\(0.702-0.838)} & \makecell{\underline{0.801}\\(0.737-0.859)} & \makecell{\textbf{0.819}\\(0.759-0.872)} & +1.8\%& $P$<0.001\\
			\midrule
			\makecell{Lung                                                              \\KRAS}      & \makecell	{0.547										 \\(0.461-0.626)} & \makecell{0.618\\(0.537-0.695)} & \makecell{0.654\\(0.579-0.727)} & \makecell{\underline{0.725}\\(0.653-0.793)} & \makecell{0.716\\(0.647-0.786)} & \makecell{\textbf{0.737}\\(0.669-0.801)} & +2.1\%& $P$<0.001\\
			\midrule
			\makecell{CRC                                                               \\BRAF}      & \makecell{0.511										 \\(0.352-0.664)} & \makecell{0.602\\(0.437-0.757)} & \makecell{0.714\\(0.509-0.902)} & \makecell{0.733\\(0.614-0.848)} & \makecell{\underline{0.756}\\(0.613-0.888)} & \makecell{\textbf{0.763}\\(0.623-0.892)} & +0.7\%& $P$<0.001\\
			\midrule
			\makecell{Brain                                                             \\IDH}            & \makecell{0.750										 \\(0.715-0.785)} & \makecell{0.905\\(0.882-0.925)} & \makecell{0.915\\(0.893-0.935)} & \makecell{0.938\\(0.921-0.954)} & \makecell{\underline{0.948}\\(0.933-0.962)} & \makecell{\textbf{0.956}\\(0.942-0.969)} & +0.8\%& $P$<0.001\\
			\bottomrule
		\end{tabular}
	\end{center}
	\label{targeted-external}
\end{table*}

\begin{table*}[!htbp]
	\caption{\textbf{Predictive performance of immunotherapy response indicators on the internal cohort.} Performance of predicting Tumor Mutational Burden (TMB) and Mismatch Repair (MMR) statuses from whole-slide images is quantified by AUROC (95\% confidence intervals). ``Gain'' indicates the absolute AUROC improvement of STAMP over the baseline (Virchow2). The \textbf{best} model is highlighted in bold, while the \underline{second-best} is underlined. $P$-values are derived from a one-sided Wilcoxon signed-rank test comparing the best and second-best models across bootstrapped distributions.}
	\fontsize{9}{10}\selectfont
	\begin{center}
		\begin{tabular}{l|cccccc|cc}
			\toprule
			Marker & OmiCLIP & PLIP & CONCH & UNI & Virchow2 & STAMP & Gain & $P$-value \\
			\midrule
			\makecell{Lung                                                              \\TMB}  & \makecell{0.725                                         \\(0.657-0.797)} & \makecell{0.757\\(0.682-0.827)} & \makecell{0.766\\(0.704-0.832)} & \makecell{\underline{0.778}\\(0.711-0.841)} & \makecell{0.774\\(0.703-0.842)} & \makecell{\textbf{0.784}\\(0.721-0.846)}& +1.0\%& $P$<0.001\\
			\midrule
			\makecell{CRC                                                               \\TMB}      & \makecell{0.705                                         \\(0.570-0.826)} & \makecell{0.629\\(0.487-0.760)} & \makecell{0.735\\(0.580-0.870)} & \makecell{0.828\\(0.686-0.946)} & \makecell{\underline{0.847}\\(0.744-0.932)} & \makecell{\textbf{0.885}\\(0.785-0.970)} &+3.8\%& $P$<0.001\\
			\midrule
			\makecell{CRC                                                               \\MMR}      & \makecell{0.496                                         \\(0.298-0.688)} & \makecell{0.590\\(0.427-0.749)} & \makecell{0.718\\(0.531-0.877)} & \makecell{0.745\\(0.599-0.873)} & \makecell{\underline{0.828}\\(0.714-0.918)} & \makecell{\textbf{0.859}\\(0.764-0.934)}& +3.1\%& $P$<0.001 \\
			\midrule
			\makecell{Gastric                                                           \\MMR}   & \makecell{0.690										 \\(0.507-0.868)} & \makecell{0.748\\(0.611-0.879)} & \makecell{0.916\\(0.835-0.980)} & \makecell{\underline{0.935}\\(0.856-0.988)} & \makecell{0.918\\(0.857-0.968)} & \makecell{\textbf{0.939}\\(0.843-0.997)} &+2.1\%& $P$<0.001\\
			\bottomrule
		\end{tabular}
	\end{center}
	\label{immune-internal}
\end{table*}

\begin{table*}[!htbp]
	\caption{\textbf{Predictive performance of immunotherapy response indicators on the external cohort.} Performance of predicting Tumor Mutational Burden (TMB) and Mismatch Repair (MMR) statuses from whole-slide images is quantified by AUROC (95\% confidence intervals). ``Gain'' indicates the absolute AUROC improvement of STAMP over the baseline (Virchow2). The \textbf{best} model is highlighted in bold, while the \underline{second-best} is underlined. $P$-values are derived from a one-sided Wilcoxon signed-rank test comparing the best and second-best models across bootstrapped distributions.}
	\fontsize{9}{10}\selectfont
	\begin{center}
		\begin{tabular}{l|cccccc|cc}
			\toprule
			Marker & OmiCLIP & PLIP & CONCH & UNI & Virchow2 & STAMP & Gain & $P$-value \\
			\midrule
			\makecell{Lung                                                              \\TMB}   & \makecell{0.548                                         \\(0.478-0.623)} & \makecell{0.609\\(0.542-0.676)} & \makecell{0.624\\(0.560-0.682)} & \makecell{\underline{0.659}\\(0.593-0.716)} & \makecell{0.647\\(0.582-0.709)} & \makecell{\textbf{0.686}\\(0.623-0.742)}  &+3.9\%& $P$<0.001\\
			\midrule
			\makecell{CRC                                                               \\TMB}       & \makecell{0.564                                         \\(0.491-0.638)} & \makecell{0.582\\(0.512-0.653)} & \makecell{0.725\\(0.662-0.784)} & \makecell{0.762\\(0.698-0.819)} & \makecell{\underline{0.796}\\(0.734-0.848)} & \makecell{\textbf{0.802}\\(0.744-0.853)}  &+0.6\%& $P$<0.001\\
			\midrule
			\makecell{CRC                                                               \\MMR}   & \makecell{0.451                                         \\(0.359-0.542)} & \makecell{0.620\\(0.513-0.718)} & \makecell{0.709\\(0.611-0.799)} & \makecell{0.713\\(0.620-0.801)} & \makecell{\underline{0.824}\\(0.761-0.882)} & \makecell{\textbf{0.845}\\(0.780-0.897)} &+2.1\%& $P$<0.001 \\
			\midrule
			\makecell{Gastric                                                           \\MMR}    & \makecell{0.736                                         \\(0.615-0.843)} & \makecell{0.825\\(0.692-0.938)} & \makecell{0.862\\(0.761-0.946)} & \makecell{\underline{0.931}\\(0.857-0.986)} & \makecell{0.913\\(0.822-0.980)} & \makecell{\textbf{0.935}\\(0.852-0.994)}  &+2.2\%& $P$<0.001\\
			\bottomrule
		\end{tabular}
	\end{center}
	\label{immune-external}
\end{table*}

\begin{table*}[!htbp]
	\caption{\textbf{Predictive performance of molecular prognostic signatures across internal multi-cancer cohorts.} Accuracy of inferring TP53 mutation, Ki-67 proliferation index, and molecular subtyping (Breast MolSub and Colorectal CMS) from whole-slide images is quantified by AUROC (95\% confidence intervals). Macro-AUC is utilized for multi-class subtyping tasks. ``Breast*'' denotes biopsy specimens, while other cohorts utilize surgical resections. ``Gain'' indicates the absolute AUROC improvement of STAMP over the baseline (Virchow2). The \textbf{best} model is highlighted in bold, while the \underline{second-best} is underlined. $P$-values are derived from a one-sided Wilcoxon signed-rank test comparing the best and second-best models across bootstrapped distributions.}
	\fontsize{9}{10}\selectfont
	\begin{center}
		\begin{tabular}{l|cccccc|cc}
			\toprule
			Marker & OmiCLIP & PLIP & CONCH & UNI & Virchow2 & STAMP & Gain & $P$-value \\
			\midrule
			\makecell{Breast                                                            \\TP53}  & \makecell{0.675                                         \\(0.595-0.760)} & \makecell{0.825\\(0.761-0.886)} & \makecell{0.857\\(0.794-0.916)} & \makecell{0.861\\(0.807-0.912)} & \makecell{\underline{0.874}\\(0.817-0.925)} & \makecell{\textbf{0.880}\\(0.823-0.931)}  &+0.6\%& $P$<0.001\\
			\midrule
			\makecell{Lung                                                              \\TP53}     & \makecell{0.658                                         \\(0.566-0.751)} & \makecell{0.758\\(0.683-0.830)} & \makecell{\textbf{0.802}\\(0.730-0.871)} & \makecell{0.736\\(0.650-0.814)} & \makecell{0.785\\(0.706-0.851)} & \makecell{\underline{0.793}\\(0.720-0.859)} &+0.7\%& $P$<0.001\\
			\midrule
			\makecell{CRC                                                               \\TP53}     & \makecell{0.538                                         \\(0.407-0.655)} & \makecell{0.656\\(0.543-0.764)} & \makecell{0.692\\(0.578-0.799)} & \makecell{0.698\\(0.579-0.809)} & \makecell{\underline{0.704}\\(0.586-0.811)} & \makecell{\textbf{0.724}\\(0.612-0.827)} &+2.0\%& $P$<0.001\\
			\midrule
			\makecell{Breast*                                                           \\Ki67}   & \makecell{0.773                                         \\(0.705-0.835)} & \makecell{0.771\\(0.708-0.828)} & \makecell{0.796\\(0.737-0.852)} & \makecell{\underline{0.822}\\(0.762-0.877)} & \makecell{0.817\\(0.756-0.871)} & \makecell{\textbf{0.837}\\(0.777-0.888)} &+2.0\%& $P$<0.001\\
			\midrule
			\makecell{Breast                                                            \\Ki67}  & \makecell{0.814                                         \\(0.684-0.919)} & \makecell{0.770\\(0.602-0.916)} & \makecell{\underline{0.813}\\(0.692-0.915)} & \makecell{0.789\\(0.664-0.900)} & \makecell{0.808\\(0.661-0.931)} & \makecell{\textbf{0.820}\\(0.681-0.929)} &+1.2\%& $P$=0.018\\
			\midrule
			\makecell{Lung                                                              \\Ki67}         & \makecell{0.767										 \\(0.717-0.812)} & \makecell{0.781\\(0.735-0.823)} & \makecell{0.837\\(0.794-0.878)} & \makecell{\textbf{0.848}\\(0.808-0.884)} & \makecell{0.803\\(0.755-0.849)} & \makecell{\underline{0.838}\\(0.796-0.876)} &+3.5\%& $P$<0.001\\
			\midrule
			\makecell{CRC                                                               \\Ki67}         & \makecell{0.689										 \\(0.587-0.790)} & \makecell{0.705\\(0.616-0.797)} & \makecell{0.716\\(0.632-0.801)} & \makecell{0.758\\(0.670-0.837)} & \makecell{\underline{0.775}\\(0.685-0.855)} & \makecell{\textbf{0.791}\\(0.719-0.862)} &+1.6\%& $P$<0.001\\
			\midrule
			\makecell{Breast*                                                           \\MolSub}   & \makecell{0.581                                         \\(0.526-0.637)} & \makecell{0.733\\(0.679-0.781)} & \makecell{0.759\\(0.715-0.803)} & \makecell{0.805\\(0.764-0.843)} & \makecell{\underline{0.813}\\(0.771-0.850)} & \makecell{\textbf{0.833}\\(0.795-0.866)} &+2.0\%& $P$<0.001\\
			\midrule
			\makecell{Breast                                                            \\MolSub}  & \makecell{0.875                                         \\(0.851-0.895)} & \makecell{0.910\\(0.890-0.927)} & \makecell{0.904\\(0.884-0.921)} & \makecell{0.920\\(0.902-0.938)} & \makecell{\underline{0.925}\\(0.908-0.941)} & \makecell{\textbf{0.928}\\(0.912-0.945)} &+0.3\%& $P$<0.001\\
			\midrule
			\makecell{CRC                                                               \\CMS}          & \makecell{0.683                                         \\(0.616-0.746)} & \makecell{0.712\\(0.639-0.781)} & \makecell{0.745\\(0.674-0.813)} & \makecell{0.751\\(0.673-0.822)} & \makecell{\underline{0.753}\\(0.668-0.824)} & \makecell{\textbf{0.763}\\(0.683-0.832)} &+1.0\%& $P$<0.001\\
			\bottomrule
		\end{tabular}
	\end{center}
	\label{prognostic-internal}
\end{table*}

\begin{table*}[!htbp]
	\caption{\textbf{Predictive performance of molecular prognostic signatures across external multi-cancer cohorts.} Accuracy of inferring TP53 mutation, Ki-67 proliferation index, and molecular subtyping (Breast MolSub) from core needle biopsy whole-slide images is quantified by AUROC (95\% confidence intervals). ``Breast*'' denotes biopsy specimens. ``Gain'' indicates the absolute AUROC improvement of STAMP over the baseline (Virchow2). The \textbf{best} model is highlighted in bold, while the \underline{second-best} is underlined. $P$-values are derived from a one-sided Wilcoxon signed-rank test comparing the best and second-best models across bootstrapped distributions.}
	\fontsize{9}{10}\selectfont
	\begin{center}
		\begin{tabular}{l|cccccc|cc}
			\toprule
			Marker & OmiCLIP & PLIP & CONCH & UNI & Virchow2 & STAMP & Gain & $P$-value \\
			\midrule
			\makecell{Breast                                                            \\TP53}     & \makecell{0.674                                         \\(0.579-0.766)} & \makecell{0.765\\(0.676-0.849)} & \makecell{0.786\\(0.700-0.866)} & \makecell{0.790\\(0.704-0.868)} & \makecell{\underline{0.840}\\(0.767-0.905)} & \makecell{\textbf{0.862}\\(0.797-0.922)} &+2.2\%& $P$<0.001\\
			\midrule
			\makecell{Lung                                                              \\TP53}          & \makecell{0.715                                         \\(0.660-0.770)} & \makecell{0.807\\(0.757-0.852)} & \makecell{\textbf{0.861}\\(0.820-0.903)} & \makecell{0.793\\(0.739-0.843)} & \makecell{0.856\\(0.810-0.896)} & \makecell{\underline{0.859}\\(0.813-0.900)} &+0.3\%& $P$<0.001\\
			\midrule
			\makecell{CRC                                                               \\TP53}          & \makecell{0.415                                         \\(0.308-0.528)} & \makecell{0.649\\(0.535-0.760)} & \makecell{\underline{0.689}\\(0.588-0.795)} & \makecell{0.684\\(0.581-0.783)} & \makecell{0.686\\(0.583-0.786)} & \makecell{\textbf{0.711}\\(0.614-0.807)} &+2.5\%& $P$<0.001\\
			\midrule
			\makecell{Breast*                                                           \\Ki67}          & \makecell{0.686                                         \\(0.636-0.735)} & \makecell{0.709\\(0.651-0.762)} & \makecell{0.711\\(0.658-0.764)} & \makecell{\underline{0.722}\\(0.664-0.779)} & \makecell{0.712\\(0.653-0.766)} & \makecell{\textbf{0.737}\\(0.684-0.786)} &+2.5\%& $P$<0.001\\
			\midrule
			\makecell{Breast                                                            \\Ki67}  & \makecell{0.769                                         \\(0.710-0.831)} & \makecell{0.744\\(0.673-0.805)} & \makecell{0.769\\(0.705-0.832)} & \makecell{0.775\\(0.714-0.834)} & \makecell{\underline{0.799}\\(0.735-0.854)} & \makecell{\textbf{0.816}\\(0.752-0.867)} &+1.7\%& $P$<0.001\\
			\midrule
			\makecell{Lung                                                              \\Ki67}         & \makecell{0.560                                         \\(0.511-0.612)} & \makecell{0.689\\(0.633-0.742)} & \makecell{0.756\\(0.701-0.805)} & \makecell{0.755\\(0.701-0.802)} & \makecell{\underline{0.781}\\(0.735-0.823)} & \makecell{\textbf{0.816}\\(0.769-0.856)} &+3.5\%& $P$<0.001\\
			\midrule
			\makecell{CRC                                                               \\Ki67}               & \makecell{0.508                                         \\(0.404-0.617)} & \makecell{0.651\\(0.561-0.742)} & \makecell{\underline{0.738}\\(0.657-0.818)} & \makecell{0.713\\(0.620-0.800)} & \makecell{0.717\\(0.635-0.795)} & \makecell{\textbf{0.744}\\(0.651-0.824)} &+2.7\%& $P$<0.001\\
			\midrule
			\makecell{Breast*                                                           \\MolSub}      & \makecell{0.569                                         \\(0.530-0.607)} & \makecell{0.630\\(0.599-0.661)} & \makecell{0.713\\(0.683-0.745)} & \makecell{0.738\\(0.709-0.768)} & \makecell{\underline{0.744}\\(0.718-0.769)} & \makecell{\textbf{0.756}\\(0.730-0.783)} &+1.2\%& $P$<0.001\\
			\midrule
			\makecell{Breast                                                            \\MolSub}     & \makecell{0.670                                         \\(0.646-0.693)} & \makecell{0.787\\(0.768-0.807)} & \makecell{0.814\\(0.794-0.834)} & \makecell{\underline{0.819}\\(0.801-0.837)} & \makecell{0.818\\(0.801-0.836)} & \makecell{\textbf{0.841}\\(0.823-0.858)} &+2.3\%& $P$<0.001\\
			\midrule
			\makecell{CRC                                                               \\CMS}                 & \makecell{0.563                                         \\(0.526-0.599)} & \makecell{0.655\\(0.616-0.690)} & \makecell{0.756\\(0.724-0.785)} & \makecell{\underline{0.808}\\(0.777-0.838)} & \makecell{\underline{0.808}\\(0.780-0.838)} & \makecell{\textbf{0.823}\\(0.794-0.851)}&+1.5\%& $P$<0.001 \\
			\bottomrule
		\end{tabular}
		\label{prognostic-external}
	\end{center}
\end{table*}

\begin{table*}[!htbp]
	\caption{\textbf{Predictive performance of clinical biomarkers across prospective observational cohorts.} Accuracy of inferring core diagnostic and prognostic markers (including ER, PR, HER2, Ki-67, and molecular subtypes for breast cancer; Napsin A, p40, TTF-1, and Ki-67 for lung cancer) from whole-slide images is quantified by AUROC (95\% confidence intervals). Macro-AUC is utilized for multi-class subtyping tasks. ``Gain'' indicates the absolute AUROC improvement of STAMP over the baseline (Virchow2). The \textbf{best} model is highlighted in bold, while the \underline{second-best} is underlined. $P$-values are derived from a one-sided Wilcoxon signed-rank test comparing the best and second-best models across bootstrapped distributions.}
	\fontsize{9}{10}\selectfont
	\begin{center}
		\begin{tabular}{l|cccccc|cc}
			\toprule
			Marker & OmiCLIP & PLIP & CONCH & UNI & Virchow2 & STAMP & Gain & $P$-value \\
			\midrule
			\makecell{Breast                                                            \\ER}  & \makecell{0.621\\ (0.510-0.731)}	& \makecell{0.840\\ (0.761-0.912)}	& \makecell{\underline{0.911}\\ (0.860-0.957)}	& \makecell{0.868\\ (0.793-0.933)}	& \makecell{0.899\\ (0.843-0.944)}	& \makecell{\textbf{0.922}\\ (0.874-0.961)} &+2.3\%&$P$<0.001\\
			\midrule
			\makecell{Breast                                                            \\PR}& \makecell{0.643\\ (0.550-0.734)}	& \makecell{0.778\\ (0.698-0.845)}	& \makecell{\underline{0.853}\\ (0.781-0.914)}	& \makecell{0.813\\ (0.738-0.880)}	& \makecell{0.843\\ (0.773-0.903)}	& \makecell{\textbf{0.862}\\ (0.797-0.918)} &+1.9\%&$P$<0.001\\
			\midrule
			\makecell{Breast                                                            \\HER2} & \makecell{0.559\\ (0.419-0.693)}	& \makecell{0.833\\ (0.742-0.919)}	& \makecell{0.880\\ (0.811-0.939)}	& \makecell{0.884\\ (0.818-0.939)}	& \makecell{\underline{0.941}\\ (0.887-0.979)}	& \makecell{\textbf{0.957}\\ (0.913-0.989)}&+1.6\%&$P$<0.001 \\
			\midrule
			\makecell{Breast                                                            \\Ki67} & \makecell{0.748\\ (0.645-0.844)}	& \makecell{0.743\\ (0.646-0.834)}	& \makecell{\underline{0.763}\\ (0.671-0.842)}	& \makecell{0.737\\ (0.643-0.821)}	& \makecell{0.758\\ (0.672-0.832)}	& \makecell{\textbf{0.773}\\ (0.687-0.847)} &+1.5\%&$P$<0.001\\
			\midrule
			\makecell{Breast                                                            \\MolSub} & \makecell{0.713\\ (0.625-0.801)}	& \makecell{0.701\\ (0.613-0.781)}	& \makecell{0.805\\ (0.737-0.865)}	& \makecell{0.762\\ (0.688-0.829)}	& \makecell{\underline{0.817}\\ (0.739-0.882)}	& \makecell{\textbf{0.840}\\ (0.770-0.897)}&+2.3\%&$P$<0.001 \\
			\midrule
			\makecell{Lung                                                              \\NapsinA}& \makecell{0.687\\ (0.586-0.772)}	& \makecell{0.789\\ (0.704-0.866)}	& \makecell{0.874\\ (0.810-0.929)}	& \makecell{\underline{0.910}\\ (0.858-0.952)}	& \makecell{0.873\\ (0.811-0.928)}	& \makecell{\textbf{0.924}\\ (0.875-0.961)}&+5.1\%&$P$<0.001\\
			\midrule
			\makecell{Lung                                                              \\TTF1}& \makecell{0.682\\ (0.581-0.772)}	& \makecell{0.707\\ (0.612-0.802)}	& \makecell{0.823\\ (0.743-0.889)}	& \makecell{\underline{0.866}\\ (0.807-0.918)}	& \makecell{0.850\\ (0.782-0.912)}	& \makecell{\textbf{0.878}\\ (0.820-0.928)}&+2.8\%&$P$<0.001\\
			\midrule
			\makecell{Lung                                                              \\Ki67} & \makecell{0.660\\ (0.616-0.700)}	& \makecell{0.737\\ (0.695-0.776)}	& \makecell{0.794\\ (0.752-0.830)}	& \makecell{\underline{0.804}\\ (0.767-0.838)}	& \makecell{0.774\\ (0.739-0.807)}	& \makecell{\textbf{0.828}\\ (0.787-0.858)}&+5.4\%&$P$<0.001\\
			\midrule
			\makecell{Lung                                                              \\P40} & \makecell{0.648\\ (0.551-0.733)}	& \makecell{0.816\\ (0.739-0.879)}	& \makecell{0.925\\ (0.879-0.960)}	& \makecell{0.940\\ (0.896-0.974)}	& \makecell{\underline{0.941}\\ (0.899-0.975)}	& \makecell{\textbf{0.948}\\ (0.911-0.977)}&+0.7\%&$P$<0.001\\
			\bottomrule
		\end{tabular}
	\end{center}
	\label{prospective}
\end{table*}

\begin{table*}[!htbp]
	\caption{\textbf{Summary of cohorts and label distributions for breast cancer.} The table details the number of cases, slides, and label distributions for key breast cancer markers across internal, external, and prospective cohorts. ``*'' denotes biopsy specimens.}
	\begin{center}
		\begin{tabular}{m{1.5cm}m{3.0cm}m{1.7cm}m{1cm}m{1cm}m{6cm}}
			\toprule
			\textbf{Organ} & \textbf{Task}                       & \textbf{Cohort} & \textbf{Cases} & \textbf{Slides} & \textbf{Label Distribution}                                                 \\
			\midrule
			\multirow{39}{*}{Breast}
			               & \multirow{2}{*}{ER*}                & Internal        & 1,485          & 3,502           & \{Negative: 474, Positive: 1011\}                                           \\
			               &                                     & External        & 703            & 703             & \{Negative: 174, Positive: 529\}                                            \\
			\cmidrule{2-6}
			               & \multirow{3}{*}{ER}                 & Internal        & 1,542          & 2,024           & \{Negative: 762, Positive: 780\}                                            \\
			               &                                     & External        & 333            & 3,811           & \{Negative: 102, Positive: 231\}                                            \\
			               &                                     & Prospective     & 168            & 375             & \{Negative: 37, Positive: 131\}                                             \\
			\cmidrule{2-6}
			               & \multirow{2}{*}{PR*}                & Internal        & 1,483          & 3,495           & \{Negative: 364, Positive: 1119\}                                           \\
			               &                                     & External        & 703            & 703             & \{Negative: 238, Positive: 465\}                                            \\
			\cmidrule{2-6}
			               & \multirow{3}{*}{PR}                 & Internal        & 1,552          & 2,029           & \{Negative: 620, Positive: 932\}                                            \\
			               &                                     & External        & 310            & 3,541           & \{Negative: 146, Positive: 164\}                                            \\
			               &                                     & Prospective     & 169            & 378             & \{Negative: 55, Positive: 114\}                                             \\
			\cmidrule{2-6}
			               & \multirow{2}{*}{HER2*}              & Internal        & 1,331          & 3,189           & \{Negative: 857, Positive: 474\}                                            \\
			               &                                     & External        & 703            & 703             & \{Negative: 409, Positive: 294\}                                            \\
			\cmidrule{2-6}
			               & \multirow{3}{*}{HER2}               & Internal        & 918            & 1,164           & \{Negative: 510, Positive: 408\}                                            \\
			               &                                     & External        & 181            & 2,095           & \{Negative: 122, Positive: 59\}                                             \\
			               &                                     & Prospective     & 107            & 239             & \{Negative: 86, Positive: 21\}                                              \\
			\cmidrule{2-6}
			               & \multirow{2}{*}{Ki67*}              & Internal        & 1,481          & 3,492           & \{Low: 310, High: 1171\}                                                    \\
			               &                                     & External        & 703            & 703             & \{Low: 84, High: 619\}                                                      \\
			\cmidrule{2-6}
			               & \multirow{3}{*}{Ki67}               & Internal        & 309            & 3,534           & \{Low: 59, High: 250\}                                                      \\
			               &                                     & External        & 221            & 304             & \{Low: 112, High: 109\}                                                     \\
			               &                                     & Prospective     & 167            & 373             & \{Low: 31, High: 136\}                                                      \\
			\cmidrule{2-6}
			               & \multirow{3}{*}{Molecular Subtype*} & Internal        & 1,400          & 3,318           & \{Luminal A: 172, Luminal B: 972, HER2+: 131, TNBC: 125\}                   \\
			               &                                     & External        & 703            & 703             & \{Luminal A: 64, Luminal B: 482, HER2+: 80, TNBC: 77\}                      \\
			\cmidrule{2-6}
			               & \multirow{5}{*}{Molecular Subtype}  & Internal        & 2,045          & 3,418           & \{Luminal A: 307, Luminal B1: 614, Luminal B2: 243, HER2+: 292, TNBC: 589\} \\
			               &                                     & External        & 788            & 791             & \{Luminal A: 142, Luminal B1: 122, Luminal B2: 82, HER2+: 149, TNBC: 293\}  \\
			               &                                     & Prospective     & 55             & 119             & \{Luminal A: 18, Luminal B1: 8, Luminal B2: 10, HER2+: 8, TNBC: 11\}        \\
			\cmidrule{2-6}
			               & \multirow{2}{*}{PIK3CA}             & Internal        & 1,013          & 1,080           & \{Wildtype: 684, Mutant: 329\}                                              \\
			               &                                     & External        & 116            & 362             & \{Wildtype: 78, Mutant: 38\}                                                \\
			\cmidrule{2-6}
			               & \multirow{2}{*}{TP53}               & Internal        & 1,013          & 1,080           & \{Wildtype: 675, Mutant: 338\}                                              \\
			               &                                     & External        & 116            & 362             & \{Wildtype: 74, Mutant: 42\}                                                \\
			\bottomrule
		\end{tabular}
	\end{center}
	\label{dataset-details-breast}
\end{table*}

\begin{table*}[!htbp]
	\caption{\textbf{Summary of cohorts and label distributions for lung, colorectal, gastric, and brain cancers.} The table details the number of cases, slides, and label distributions for key markers across internal, external, and prospective cohorts for each organ.}
	\begin{center}
		\begin{tabular}{m{1.5cm}m{3.0cm}m{1.7cm}m{1cm}m{1cm}m{6cm}}
			\toprule
			\textbf{Organ}               & \textbf{Task}              & \textbf{Cohort} & \textbf{Cases} & \textbf{Slides} & \textbf{Label Distribution}                   \\
			\midrule
			\multirow{23}{*}{Lung}       & \multirow{3}{*}{Napsin A } & Internal        & 695            & 707             & \{Negative: 391, Positive: 304\}              \\
			                             &                            & External        & 92             & 333             & \{Negative: 38, Positive: 54\}                \\
			                             &                            & Prospective     & 128            & 159             & \{Negative: 52, Positive: 76\}                \\
			\cmidrule{2-6}
			                             & \multirow{3}{*}{TTF-1}     & Internal        & 1,163          & 1,188           & \{Negative: 486, Positive: 677\}              \\
			                             &                            & External        & 558            & 774             & \{Negative: 326, Positive: 232\}              \\
			                             &                            & Prospective     & 138            & 174             & \{Negative: 47, Positive: 91\}                \\
			\cmidrule{2-6}
			                             & \multirow{3}{*}{p40}       & Internal        & 898            & 917             & \{Negative: 541, Positive: 357\}              \\
			                             &                            & External        & 166            & 557             & \{Negative: 109, Positive: 57\}               \\
			                             &                            & Prospective     & 152            & 188             & \{Negative: 94, Positive: 58\}                \\
			\cmidrule{2-6}
			                             & \multirow{3}{*}{Ki67}      & Internal        & 1,128          & 1151            & \{Low: 101, Medium: 473, High: 554\}          \\
			                             &                            & External        & 239            & 630             & \{Low: 112, Medium: 84, High: 43\}            \\
			                             &                            & Prospective     & 309            & 390             & \{Low: 87, Medium: 136, High: 86\}            \\
			\cmidrule{2-6}
			                             & \multirow{2}{*}{CK7}       & Internal        & 845            & 863             & \{Negative: 226, Positive: 619\}              \\
			                             &                            & External        & 460            & 635             & \{Negative: 218, Positive: 242\}              \\
			\cmidrule{2-6}
			                             & \multirow{2}{*}{TP53}      & Internal        & 893            & 990             & \{Wildtype: 291, Mutant: 602\}                \\
			                             &                            & External        & 329            & 1,513           & \{Wildtype: 124, Mutant: 205\}                \\
			\cmidrule{2-6}
			                             & \multirow{2}{*}{EGFR}      & Internal        & 1,265          & 1,265           & \{Wildtype: 588, Mutant: 677\}                \\
			                             &                            & External        & 452            & 515             & \{Wildtype: 396, Mutant: 56\}                 \\
			\cmidrule{2-6}
			                             & \multirow{2}{*}{KRAS}      & Internal        & 1,265          & 1,265           & \{Wildtype: 1,111, Mutant: 154\}              \\
			                             &                            & External        & 223            & 1,050           & \{Wildtype: 156, Mutant: 67\}                 \\
			\cmidrule{2-6}
			                             & \multirow{2}{*}{TMB}       & Internal        & 1,523          & 1,523           & \{Low: 1,274, High: 249\}                     \\
			                             &                            & External        & 330            & 1,515           & \{Low: 254, High: 76\}                        \\
			\midrule
			\multirow{16}{*}{Colorectal} & \multirow{2}{*}{Ki67}      & Internal        & 1,491          & 3,050           & \{Low: 222, High: 1,269\}                     \\
			                             &                            & External        & 328            & 328             & \{Low: 37, High: 291\}                        \\
			\cmidrule{2-6}
			                             & \multirow{3}{*}{CMS}       & Internal        & 470            & 2,137           & \{CMS1: 55, CMS2: 144, CMS3: 104, CMS4: 167\} \\
			                             &                            & External        & 459            & 464             & \{CMS1: 73, CMS2: 166, CMS3: 94, CMS4: 126\}  \\
			\cmidrule{2-6}
			                             & \multirow{2}{*}{MMR}       & Internal        & 595            & 2,706           & \{MSS: 536, MSI: 59\}                         \\
			                             &                            & External        & 855            & 856             & \{MSS: 818, MSI: 37\}                         \\
			\cmidrule{2-6}
			                             & \multirow{2}{*}{TP53}      & Internal        & 494            & 501             & \{Wildtype: 208, Mutant: 286\}                \\
			                             &                            & External        & 104            & 217             & \{Wildtype: 47, Mutant: 57\}                  \\
			\cmidrule{2-6}
			                             & \multirow{2}{*}{TMB}       & Internal        & 608            & 2,779           & \{Low: 525, High: 83\}                        \\
			                             &                            & External        & 494            & 501             & \{Low: 422, High: 72\}                        \\
			\cmidrule{2-6}
			                             & \multirow{2}{*}{BRAF}      & Internal        & 494            & 501             & \{Wildtype: 435, Mutant: 59\}                 \\
			                             &                            & External        & 104            & 217             & \{Wildtype: 89, Mutant: 15\}                  \\
			\midrule
			\multirow{2}{*}{Gastric}     & \multirow{2}{*}{MMR}       & Internal        & 637            & 3,790           & \{dMMR: 53, pMMR: 584\}                       \\
			                             &                            & External        & 270            & 270             & \{dMMR: 19, pMMR: 251\}                       \\
			\midrule
			\multirow{2}{*}{Brain}       & \multirow{2}{*}{IDH}       & Internal        & 543            & 968             & \{IDHwt: 149, IDHmut: 394\}                   \\
			                             &                            & External        & 775            & 852             & \{IDHwt: 530, IDHmut: 322\}                   \\
			\bottomrule
		\end{tabular}
	\end{center}
	\label{dataset-details-others}
\end{table*}

\begin{table*}[!htbp]
	\caption{\textbf{Summary of data resources used in model evaluation.} The table details the data sources, their accessibility status, and the number of cases and slides available for each source. Private datasets (H1-H10) are not publicly accessible, while TCGA, CPTAC, and eBrain are open-access resources.}
	\setlength{\tabcolsep}{15pt}
	\begin{center}
		\begin{tabular}{llcc}
			\toprule
			\textbf{Data Source} & \textbf{Accessible}                 & \textbf{Case} & \textbf{Slide} \\
			\midrule
			H1                   & Private                             & 3,414         & 9,131          \\
			H2                   & Private                             & 3,074         & 8,329          \\
			H3                   & Private                             & 1,516         & 3,585          \\
			H4                   & Private                             & 608           & 2,779          \\
			H5                   & Private                             & 1,483         & 1,593          \\
			H6                   & Private                             & 1,523         & 1,523          \\
			H7                   & Private                             & 855           & 856            \\
			H8                   & Private                             & 788           & 791            \\
			H9                   & Private                             & 703           & 703            \\
			H10                  & Private                             & 328           & 328            \\
			\midrule
			TCGA                 & \url{https://portal.gdc.cancer.gov} & 3,018         & 3,615          \\
			CPTAC                & \url{https://portal.gdc.cancer.gov} & 551           & 3,144          \\
			eBrain               & \url{https://ebrains.eu}            & 775           & 852            \\
			\midrule
			Total                & -                                   & 18,636        & 37,229         \\
			\bottomrule
		\end{tabular}
	\end{center}
	\label{dataset-accessibility}
\end{table*}

\end{document}